\documentclass{article}
\usepackage[utf8]{inputenc}
\usepackage[T1]{fontenc}
\usepackage{iclr2027_conference,times}
\iclrfinalcopy  
\usepackage{url}
\usepackage{graphicx}
\usepackage{pgf}
\usepackage{booktabs}
\usepackage{multirow}
\usepackage{tabularx}
\usepackage{amsmath,amssymb}
\usepackage{xcolor}
\usepackage{natbib}
\usepackage{hyperref}
\usepackage{colortbl}
\usepackage{wrapfig}
\usepackage{needspace}
\usepackage{subcaption}
\hypersetup{colorlinks=true,linkcolor=blue,citecolor=blue,urlcolor=blue}
\newcommand{\ours}{Arctic-SWE}
\definecolor{figgray}{HTML}{666666}
\definecolor{figteal}{HTML}{178A78}
\definecolor{figred}{HTML}{C2452D}
\definecolor{figamber}{HTML}{B07410}
\definecolor{figblue}{HTML}{1B5E9C}
\definecolor{figpurple}{HTML}{8E5FB8}

\title{\ours: Revisiting the Long Horizon Agent Training by Modeling-System Co-Design}
\author{Boyi Liu$^{1,\ast,\dagger}$, Yite Wang$^{1,\ast,\dagger}$, Xinyu Lian$^{2,\ast}$, Fan Shu$^{3,\ast}$, Sida Li$^{4}$,\\
\bfseries Feng Yan$^{3}$, Minjia Zhang$^{2}$, Yuxiong He$^{1}$, Zhewei Yao$^{1,\dagger}$ \\[5pt]
\small $^{1}$Snowflake \quad $^{2}$University of Illinois at Urbana-Champaign \\
\small $^{3}$University of Houston \quad $^{4}$University of Chicago \\[3pt]
\small $^{\ast}$Equal contribution. \quad $^{\dagger}$Project leads.}
\date{}

\begin{document}
\maketitle
\fancyhead{}
\renewcommand{\headrulewidth}{0pt}

\begin{abstract}
Long-horizon agent training, including software engineering, is advancing rapidly,
but the recipe and infrastructure needed for a strong reinforcement-learning
baseline remain difficult to reproduce. This makes gains from new algorithms
or data hard to distinguish from improvements to an under-tuned baseline.
We present \ours, a fully asynchronous RL framework that establishes such a
baseline through model--system co-design, using overfitting on a small task
pool as the testbed. This testbed exposes failures that early reward gains
conceal and guides the integration of established components: gated reward,
importance-weight control, aligned training and inference engines, and
consistent state across policy updates. No single component is new, yet
together they form the missing foundation. On raw R2E-Gym data, without
new algorithms or data work, \ours{} trains Qwen3.5-4B to 62.0\% pass@1 on
SWE-bench Verified. This is comparable to concurrent FrogNano (61.5\%), which
relies on online task synthesis, after about one-sixth as many training
rollouts.
Building on this foundation, we further filter and refine R2E-Gym,
SWE-rebench V2, and Scale-SWE.
With this curated subset, the 4B model improves from 47.4\%/25.0\%/37.9\% to
65.0\%/38.3\%/59.7\% on SWE-bench Verified/Pro-v1/Pro-v2, and the recipe also improves
Qwen3.5-9B from 56.2\%/31.9\%/48.8\% to 65.7\%/39.6\%/70.2\% on Verified/Pro-v1/Pro-v2. Controlled comparisons show that both curation
and test strengthening improve held-out performance, although stronger tests
reduce apparent training-task mastery. \ours{} provides a documented starting
point for studying better algorithms, data and environments for asynchronous
agent RL.
\end{abstract}

\section{Introduction}

Long-horizon agents learn to solve tasks through extended interaction with
tools and environments. Software engineering is a prominent example: agents
inspect repositories, edit code and execute tests, with reinforcement learning
(RL) improving their ability to complete these workflows
\citep{swegym2024,r2egym,deepswe2025}. Recent work advances both online task
synthesis, as in FrogNano~\citep{frognano2026}, and large-scale model--system
co-design, as in MiMo-V2.6~\citep{mimo2026}. Yet reproducing a strong training
baseline requires choices that a named algorithm or a released checkpoint does
not specify. For academic and open-source efforts with limited compute,
establishing that baseline is itself a substantial research problem.

This matters when assessing new methods. An algorithm or dataset can appear
better because the comparison recipe is under-tuned, much as optimizer
comparisons depend on tuning AdamW or SGD with momentum~\citep{wen2025fantastic},
and policy-gradient comparisons depend on implementation
details~\citep{engstrom2020implementation}. In supervised fine-tuning,
establishing a baseline is straightforward: minimizing cross-entropy reliably
overfits a small training set, which provides a basic implementation check. Group-relative
RL methods such as GRPO~\citep{grpo} have no comparably routine
baseline check for long, stateful agent interactions. Early progress is
insufficient: a recipe can learn tasks and subsequently lose them.

Asynchrony makes the baseline a joint modeling and systems problem. Generation, tool execution, and optimization must overlap heavily to maintain throughput, so the system efficiency now directly changes the age of the experience used for learning. Weight refreshes can leave cached states inconsistent, and different numerical paths can assign different probabilities to the same tokens. Grading and termination further determine which trajectories receive credit. These system-level choices directly alter the data reaching the optimizer, not only its throughput.

We present \ours, a fully asynchronous RL framework built on Prime-RL \citep{primerl_qwen_swe_2026}. We use \emph{overfitting as a stress test} to establish its recipe: prolonged training on a small pool checks whether the pipeline both acquires and sustains success. The failures guide feedback, optimizer and execution changes (Section~\ref{sec:recipe}). None of these changes is new on its own; what has been missing is the foundation they form together. On the existing R2E-Gym~\citep{r2egym} dataset, this foundation alone yields 62.0\% SWE-bench Verified~\citep{swebench2023}  pass@1 for Qwen3.5-4B, without new data construction or refinement. FrogNano centers its contribution on online task synthesis and reports 61.5\% after consuming about 6 times more training rollouts compared to ours. A solid recipe, rather than elaborate data or algorithm work, already reaches this level.

A stronger baseline also lets us ask what better data contributes.
Execution-grounded filtering and issue refinement retain a subset of the source
pool combined from R2E-Gym, SWE-rebench V2~\citep{rebenchv2}, and Scale-SWE~\citep{scaleswe}. 
Matched comparisons show that this curation improves generalization;
strengthening weak tests recovers additional useful tasks and improves learning
despite reducing apparent mastery (Section~\ref{sec:data}). The resulting
framework and curated training data improve both 4B and 9B models on SWE-bench Verified/Pro-v1/Pro-v2~\citep{swebenchpro} from 47.4\%/25.0\%/37.9\% to
65.0\%/38.3\%/59.7\%, and from 56.2\%/31.9\%/48.8\% to 65.7\%/39.6\%/70.2\%, respectively. 
These full-scale runs yield further insights into the recipe (Section~\ref{sec:main-run}). Our aim is to make a strong baseline reproducible, so that further advances in algorithms, data and environments can be measured against a training pipeline that already learns effectively.

\section{Setup}
\label{sec:setup}
All experiments share the following pipeline; Appendix~\ref{app:setup} gives the full specification.

\textbf{Hardware and pipeline.}
All training runs use one node of eight B200 GPUs: two train the policy and six serve it as independent vLLM inference engines. For asynchronous training, the inference engines keep generating while the trainer updates and receive new weights after every update, so one multi-turn rollout can span several policy versions. The \texttt{max\_off\_policy\_steps} parameter controls such staleness: a rollout more than that many policy versions stale is dropped.

\textbf{Model and agent.}
We train Qwen3.5-4B in all sections, and we train Qwen3.5-9B with our final full recipe in Section~\ref{sec:main-run}. Each rollout runs our Mini-SWE-Agent-Plus~\citep{minisweplus_2026} harness, which has two tools \texttt{execute\_bash} and \texttt{edit\_via\_str\_replace}, in its own sandbox container on the training node. A trajectory is capped at 100 turns and 131{,}072 tokens in total; each model response is capped at 32{,}768 tokens. Training rollouts sample at temperature 1.0 on full vocabulary; evaluation samples at temperature 1.0 with top-$p$ 0.95.

\textbf{Training and evaluation.}
Each update samples a group of rollouts for every task in the batch. We optimize with CISPO~\citep{cispo2025}, which weights each token's gradient by its importance ratio between the trainer and the rollout policy; the ratio is detached and bounded at 2.0. We evaluate on SWE-bench Verified (a 483-problem subset) and SWE-bench Pro with the same harness and budgets, and report pass@1 estimated via the mean success over four rollouts per problem.

\section{Model-System Co-Design}
\label{sec:recipe}

We develop the \ours{} baseline in three steps: we use a small task pool as the testbed to expose failure modes (Section~\ref{sec:stress}), correct the feedback and execution loops (Sections~\ref{sec:env} and~\ref{sec:infra}), and train the resulting recipe on full R2E-Gym dataset with unchanged issues and tests (Section~\ref{sec:r2e-baseline}). 

\subsection{RL Overfitting as the Stress Test}
\label{sec:stress}
If an RL pipeline cannot reliably memorize a small, solvable subset of tasks, it cannot be expected to generalize over a large, noisy corpus. To isolate infrastructure failures from genuine algorithmic limits, we use overfitting as a diagnostic stress test. We filter 160 tasks from R2E-Gym where Qwen3.5-4B passes the unit tests in at least one out of four attempts, and train on them in our Mini-SWE-Agent-Plus harness (Appendix~\ref{app:harness}) with 32 tasks per update and five rollouts per task. Alongside reward, we track token entropy and \emph{rollout--trainer mismatch KL}, the mean of $\omega-\log\omega-1$ over trainable tokens, where $\omega$ is the trainer probability of a sampled token divided by its recorded rollout probability (see formal definition in Appendix~\ref{app:mismatch-kl}).

\begin{figure}[t]
\centering
\includegraphics[width=\linewidth]{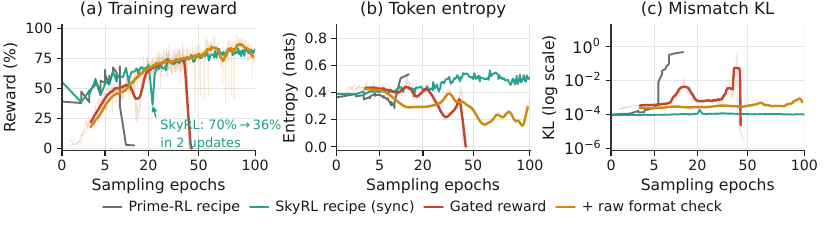}
\vspace{-0.6em}
\caption{\textbf{Early gains can conceal instability.}
The async \textcolor{figgray}{Prime-RL} and sync \textcolor{figteal}{SkyRL} recipes both show instability after climbing to about
70\%. The \textcolor{figred}{Gated reward} and \textcolor{figamber}{+ raw format check} runs show 21-update
means over faint per-update values; the Prime-RL and SkyRL lines are per-update. Epochs use a square-root axis.}
\label{fig:recipe_before}
\vspace{-1.2em}
\end{figure}

\textbf{A promising early curve can conceal a failing recipe.}
To illustrate this, we test two established public baselines on our 160-task overfit set (Figure~\ref{fig:recipe_before}). We evaluate a \emph{synchronous} pipeline (\textcolor{figteal}{SkyRL}, \cite{skyrl_mini_2026}), which pauses the trainer to generate all rollouts using the exact current policy, alongside an \emph{asynchronous} pipeline (\textcolor{figgray}{Prime-RL}, \cite{primerl_qwen_swe_2026}), which overlaps generation and optimization to maximize throughput but introduces policy lag. When adapted to Qwen3.5-4B (see details in Appendix~\ref{app:public_overfit}), both recipes initially appear successful, climbing to roughly 70\% reward (Figure~\ref{fig:recipe_before}a). However, prolonged training exposes severe instabilities especially in async training.
Neither recipe reliably sustains its early gains. 
This shows that standard algorithms are remarkably brittle over long horizons, and that evaluating only the first few epochs of a training run obscures fatal recipe deficits.

\textbf{Why the diagnostic matters beyond the overfitting tasks.}
\label{sec:consistency}
Standard agent training rarely runs for dozens of epochs on a small dataset, but the instability that prolonged overfitting exposes still limits full-scale training. If a recipe acquires task behaviors at rate $g$ and loses them at rate $d$, its mastery of solvable tasks stays below $1-\varepsilon$ with $\varepsilon=d/(g+d)$; on a full dataset in which such tasks make up a fraction $\rho$, reward then stays at least $\rho\varepsilon$ below the attainable optimum (Appendix~\ref{app:consistency}). A leaky recipe can show rapid early gains yet carry a permanent ceiling that neither more data nor a larger model removes, so the feedback loop must be stabilized first.

\subsection{Correcting Algorithmic Feedback}
\label{sec:env}
\label{sec:integrity}
Having established that standard recipes are unstable in prolonged training, we repair the training loop one layer
at a time, starting with the algorithmic feedback: when does each rollout terminate, and which/when rollouts earn reward. We read each change from the earlier-stack curves of Figure~\ref{fig:recipe_before}; Section~\ref{sec:infra} revisits them after the system fix.

\textbf{Do not let harness remedies pass unchecked into reward.}
Agent harnesses often help a model past a malformed turn, for instance by returning a
syntactic hint and letting the rollout continue. If the rollout then succeeds, the invalid
prefix shares in the positive trajectory reward, so the policy is reinforced for behavior the
harness corrected rather than behavior it produced. The \textcolor{figgray}{Prime-RL recipe} leaves what the harness does unconstrained, so hinted invalid turns go unpenalized; its line in Figure~\ref{fig:recipe_before} collapses by epoch 11.

\textbf{Gate reward on valid completion.}
We make reward eligibility strictly explicit. We require an explicit submission and terminate
duplicate tool calls, invalid formats and exhausted turn or token budgets immediately,
assigning zero reward before computing group-relative advantages. The red \textcolor{figred}{\emph{Gated reward}} curve in
Figure~\ref{fig:recipe_before} applies this contract. It learns steadily to a 74.5\% plateau,
but at epoch 42 it starts to collapse and shows format issues.

\textbf{Check the raw token output, not only the parsed call.}
The gate above inspects tool calls after the tool-call parser has converted the sampled tokens
into a structured call. Parsers are lenient, so a malformed raw token output that the parser
repairs can still earn reward and get reinforced over time. On the train-time paths of major public RL frameworks~\citep{primerl_qwen_swe_2026,skyrl_mini_2026,slime_swe_2026,miles_swe_2026}, a call still executes for most of 13 malformed Qwen3.5 turns, and our gate alone lets 6 of them through (Appendix~\ref{app:parsers}, Table~\ref{tab:parsers}). The
\textcolor{figamber}{\emph{+ raw format check}} line also validates the raw token output itself: the reasoning
block, the tool-call envelope and the tags inside each call. It never collapses but still falls short of a full fit:
reward climbs slowly, holding around 80\% and peaking at 87.3\%. 

\subsection{Correcting the System Layer}
\label{sec:infra}
An algorithmically sound update can still diverge if the experience supplied by the asynchronous rollout engines is inconsistent with the policy being trained. Unlike synchronous RL, where system efficiency affects only training speed, asynchronous RL lets the speed of the rollout engines and the trainer shape the training dynamics as well.

\begin{figure}[t]
\centering
\includegraphics[width=\linewidth]{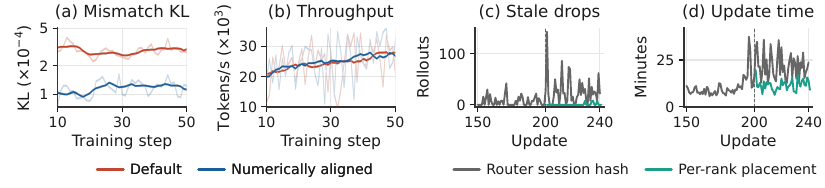}
\vspace{-0.6em}
\caption{\textbf{System layer corrections change the experience the optimizer sees.}
(a,b) Numerical alignment lowers mismatch KL without a throughput penalty (dark curves are
averages). (c,d) One run resumed from update 200 (dashed) with per-rank placement: rollouts
dropped at the off-policy cap and wall-clock time per update.}
\label{fig:infra_align}
\vspace{-0.6em}
\end{figure}

\textbf{Match training and inference numerics.}
\label{sec:numerics}
Qwen3.5's hybrid recurrent architecture makes its repeatedly updated state particularly sensitive to rounding errors. By aligning the numerical paths between the rollout generators and the trainer, we reduced rollout--trainer mismatch KL from roughly $3 \times 10^{-4}$ to $1.2 \times 10^{-4}$ without a throughput penalty (\textcolor{figred}{default} versus \textcolor{figblue}{numerically aligned} in Figure~\ref{fig:infra_align}a,b). Specifically, across both optimization and weight transfer, we retain FP32 for recurrent states, gate arithmetic, and $A_{\log}$, the learned per-head decay rate of the recurrent state in log space (Appendix~\ref{app:numerics}). Other FP32 paths, such as master weights, gradient reduction, RMSNorm statistics and the LM head, are now standard practice and not specific to our recipe.

\begin{figure}[t]
\centering
\includegraphics[width=\linewidth]{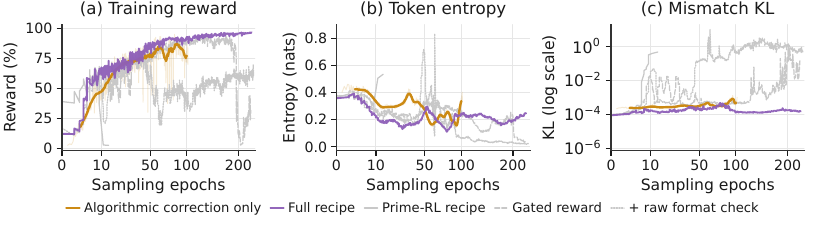}
\vspace{-0.6em}
\caption{\textbf{With the corrected system layer, the full recipe fits the pool.}
The \textcolor{figamber}{raw-format-check} run of Figure~\ref{fig:recipe_before} against the same recipe with the system fixes of Section~\ref{sec:infra}, our \textcolor{figpurple}{full recipe}. Faint gray lines are rerun of Figure~\ref{fig:recipe_before} setting with system fixes in place, which tell us that system corrections alone are not enough to stable async RL.
}
\label{fig:recipe_after}
\vspace{-0.6em}
\end{figure}

\textbf{Reconstruct state after a weight refresh.}
\label{sec:cache_consistency}
When active model weights are updated asynchronously, cached historical states (attention KV, convolution history and recurrent state) immediately misalign with the new policy. PipelineRL~\citep{pipelinerl} reuses stale KV caches, having found that they barely increase the divergence from the on-policy distribution for a softmax-attention model on
single-turn reasoning. 
Multi-turn agent rollouts can span several weight refreshes, and in our hybrid model, stale
attention and GatedDeltaNet caches both shift next-token distributions
(Appendix~\ref{app:stale_cache}). We therefore invalidate all caches on refresh and
re-prefill the retained history (\emph{KV recomputation}).


\textbf{Schedule for useful, recent experience.}
A policy-lag cap alone does not control the age distribution of the batch.
Six vLLM engines serve rollouts (Appendix~\ref{app:setup}), and a router assigns requests by \emph{session hashing}, pinning each trajectory to one engine to reuse its prefix cache. This ignores load: one engine held 186 concurrent requests against 8 on the others. Because a group is trained only once all its rollouts complete, these stragglers pushed whole groups past the off-policy cap of four updates. Our dispatcher instead assigns each group to the least-loaded engine and caps resident sequences per engine. Applied from update 200, this placement cuts stale drops per update from 27 to 0.8 and mean policy lag from 3.1 to 2.0 updates, and nearly halves update time (22 to 11.6 minutes) at unchanged forward--backward cost (\textcolor{figgray}{router session hash} versus \textcolor{figteal}{per-rank placement} in Figure~\ref{fig:infra_align}c,d; Appendix~\ref{app:systems}). On full data, \emph{easy--hard eviction} also retires always-solved and repeatedly failed tasks, so that sampled groups keep a learning signal. Sandboxes run on the node's spare capacity with overcommitted limits, sustaining 252 concurrent instances without extra hardware (Appendices~\ref{app:sandboximplementation},~\ref{app:sandboxcost}).


\textbf{The full recipe sustains overfitting.}
\label{sec:plateau}
Combining these algorithmic and systemic corrections transforms the baseline (Figure~\ref{fig:recipe_after}). The \textcolor{figamber}{raw-format-check} recipe of Figure~\ref{fig:recipe_before} holds around 80\% before the system fixes. With the system fixes of this section, the same recipe, our \textcolor{figpurple}{full recipe}, reaches 93.0\% over epochs 100--150 and 95.8\% over epochs 200--230 while keeping mismatch low and token entropy healthy. The faint lines are the other reruns of the Figure~\ref{fig:recipe_before} settings: the Prime-RL recipe and gated reward still collapse, and the raw format check stays below 60\%.


\Needspace{8\baselineskip}
\subsection{\ours: A Strong Baseline on Raw Data}
\label{sec:carryforward}
\label{sec:r2e-baseline}

\begin{wrapfigure}{r}{0.47\linewidth}
\vspace{-1.2em}
\centering
\includegraphics[width=\linewidth]{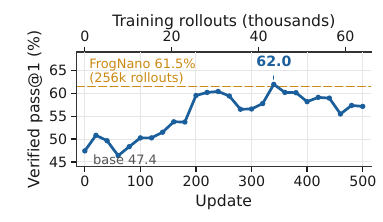}
\caption{\textbf{The foundation alone nears FrogNano.} Verified pass@1 on unchanged
R2E-Gym, first 500 of 800 updates.}
\label{fig:r2e-run}
\vspace{-1.2em}
\end{wrapfigure}

Building on the model--system co-design recipe above, we train on the full R2E-Gym dataset and refine the remaining recipe details, including the task eviction strategy and importance-ratio handling (Appendix~\ref{app:r2e-recipe}). Our final results show that a solid foundation is competitive before any data refinement.

\textbf{A solid foundation is competitive before any data work.}
We train the resulting recipe on 3{,}606 R2E-Gym tasks with their original
issues and tests, without new task generation, issue rewriting or test
refinement. 
The run keeps every feedback and execution component above, adds
easy--hard eviction, and uses AdamW $\beta_2=0.98$ with no length penalty,
advantage normalization or reference KL (Appendix~\ref{app:r2e-recipe}). \ours{} trains Qwen3.5-4B from 47.4\% to 62.0\%
SWE-bench Verified pass@1 at update 340, and holds near 60\% from update 200 through 440 (\textcolor{figblue}{solid line} in Figure~\ref{fig:r2e-run}; the dashed line is \textcolor{figamber}{FrogNano's 61.5\%}). On SWE-bench Pro-v1 it reaches 38.3\%.

\textbf{This level is reachable without task synthesis.} Concurrent FrogNano reports 61.5\% with the same base model by synthesizing and calibrating tasks online~\citep{frognano2026}; under its own recipe, RL on real SWE-rebench and Scale-SWE tasks reaches only 48.0--53.4\%, and FrogNano scores 56.4\% under mini-SWE-agent, a harness close to ours (Appendix~\ref{app:frognano}).

\section{Auditing and repairing the training reward through execution}
\label{sec:data}

Section~\ref{sec:recipe} achieves $62.0\%$ on SWE-bench Verified with R2E-Gym
tasks checked only for executability. But a good recipe cannot fix a bad
reward from fragile data, and that check is weak. On nearly a third of the
tasks that pass it, a plausible wrong patch still obtains full reward. Standard
data validation often misses this, because it typically does not score a wrong patch. This
section filters and repairs the tasks themselves without making any: we draw on
three existing released datasets.
Section~\ref{sec:filtering} checks each task's three artifacts against one
another, keeping $7{,}500$ of $21{,}099$ and rewriting the issue where it is at
fault; Section~\ref{sec:hardening} strengthens the tests of the largest
rejected category instead of discarding it. Both are validated by controlled
experiments in which only the training data changes.

\subsection{When the issue, the tests and the gold patch disagree}
\label{sec:filtering}

Every SWE training task is three artifacts that are supposed to describe the
same change: (1) the issue the agent reads, (2) the test suite that computes its
reward, and (3) the gold patch that defines the intended fix. Nothing in how
opensourced datasets are built guarantees the three agree. The fail-to-pass gate that
every existing dataset applies \citep{swesmith,r2egym,rebenchv2} checks only
the weakest consequence of agreement---that the tests separate the buggy tree
from gold---and never asks whether they also accept patches that do not fix the
issue, nor whether the issue states what they demand. Our curation therefore
runs in two pipelines. \textbf{Pipeline~1} keeps that gate and applies it with
code execution only: over 21{,}099 candidates from R2E-Gym, SWE-rebench-V2, and Scale-SWE
\citep{r2egym,rebenchv2,scaleswe}, it runs the graded suite on the buggy and
gold trees to confirm the fail-to-pass behavior and instruments every assertion
to identify which are \emph{binding}, admitting \textbf{17{,}861}
(Table~\ref{tab:alignment}).

\providecommand{\subrow}{\quad\textcolor{gray}{$\scriptstyle\llcorner$}\,}

\begin{table}[t]
\centering
\small
\begin{minipage}[t]{0.32\linewidth}
\vspace{0pt}
\captionsetup{font=small}
\caption{Three artifacts disagree. Pipeline~1 establishes only
\emph{weak} agreement that the tests separate the buggy tree from gold.
Pipeline~2 may repair only the issue, so a task survives \emph{here} only if
its tests and gold patch are sound on their own; Section~\ref{sec:hardening} returns
to the \texttt{too\_loose} rows, broken out above. Verdict-level and per-corpus
breakdowns in Appendix~\ref{app:curation-funnel}.}
\label{tab:alignment}
\end{minipage}
\hfill
\begin{minipage}[t]{0.66\linewidth}
\vspace{0pt}
\centering
\resizebox{\linewidth}{!}{%
\begin{tabular}{llrr}
\toprule
Outcome & Relation among (issue, tests, gold patch) & $n$ & \% \\
\midrule
\multicolumn{2}{l}{\emph{Pipeline 1: execution only, no model}}
& \multicolumn{2}{r}{\emph{over $21{,}099$}} \\

Source candidates & --- & 21{,}099 & 100 \\

Reject & tests and gold patch disagree, or not executable
& $-3{,}238$ & 15.3 \\

\rowcolor[gray]{0.86}
\textbf{$=$ Validated pool}
& \textbf{tests and gold patch weakly agree}
& \textbf{17{,}861} & \textbf{84.7} \\

\midrule
\multicolumn{2}{l}{\emph{Pipeline 2: privileged LLM judge}}
& \multicolumn{2}{r}{\emph{over $17{,}861$}} \\

Reject & tests disagree w. the issue: too loose/over-restrictive/off target
& $-9{,}379$ & 52.5 \\

& \subrow \textcolor{black!60}{of which \texttt{too\_loose}
(Section~\ref{sec:hardening})}
& \textcolor{black!60}{5{,}351} & \textcolor{black!60}{30.0} \\

Reject & gold patch does not fix the issue
& $-926$ & 5.2 \\

Other & rewrite elided the original; judge error
& $-56$ & 0.3 \\

\rowcolor[gray]{0.86}
\textbf{$=$ Curated pool}
& \textbf{all three agree, after repair where warranted}
& \textbf{7{,}500} & \textbf{42.0} \\

\rowcolor[gray]{0.86}
& \subrow \textcolor{black!60}{issue rewritten to state what the tests grade}
& \textcolor{black!60}{1{,}233} & \textcolor{black!60}{6.9} \\
\bottomrule
\end{tabular}%
}
\end{minipage}
\vspace{-1.0em}
\end{table}

Clearing that gate does not make the three artifacts agree. A pull-request test
is written by someone who already holds the fix, so it is usually a regression
test rather than a specification built to reject wrong patches; and an issue
written before the fix often says less than the tests come to demand. Both gaps
are invisible to the Pipeline-1 gate, and to any check that reads the three artifacts
without scoring a patch against them. The audit described next finds that on \textbf{5{,}351 of the 17{,}861}, nearly a third, a
plausible wrong patch written from the issue alone still earns full reward
(\texttt{too\_loose}), and that on \textbf{1{,}233} the tests demand behavior
the issue never mentions (\texttt{issue\_underspecified}). Left alone, both
could create misaligned training signals: a loose test suite assigns full
reward to some issue-inconsistent patches, creating opportunities for reward
hacking as RL training explores a broader patch distribution, while an
underspecified issue can penalize behavior that cannot be inferred from the
prompt (Appendix~\ref{app:prior-filters}).

\textbf{Pipeline~2} answers the two questions the Pipeline-1 gate never asks with the help of LLMs. It
gives an agent powered by Qwen3.8-27B \citep{qwen38} the gold patch, the tests,
the Pipeline-1 measurements, and a shell, and puts three questions to it in a
fixed dependency order: (1) whether the gold
patch fixes the issue, (2) whether the tests grade what the issue asks,
and---\emph{only if both hold}---(3) whether the issue states what the tests
demand. It is privileged but not trusted to answer by inspection: the middle
question is settled by writing counterfactual patches and scoring them with the
training scorer used in RL, 
the agent
may only \textbf{rewrite the issue text}. Pipeline~2 accepts 7{,}500 of the
17{,}861: 6{,}267 with the issue unchanged and 1{,}233 rewritten to state
what the tests already grade (Appendices~\ref{app:curation-funnel}
and~\ref{app:curation-examples}).

\begin{figure}[t]
  \centering
  \includegraphics[width=0.9\linewidth]{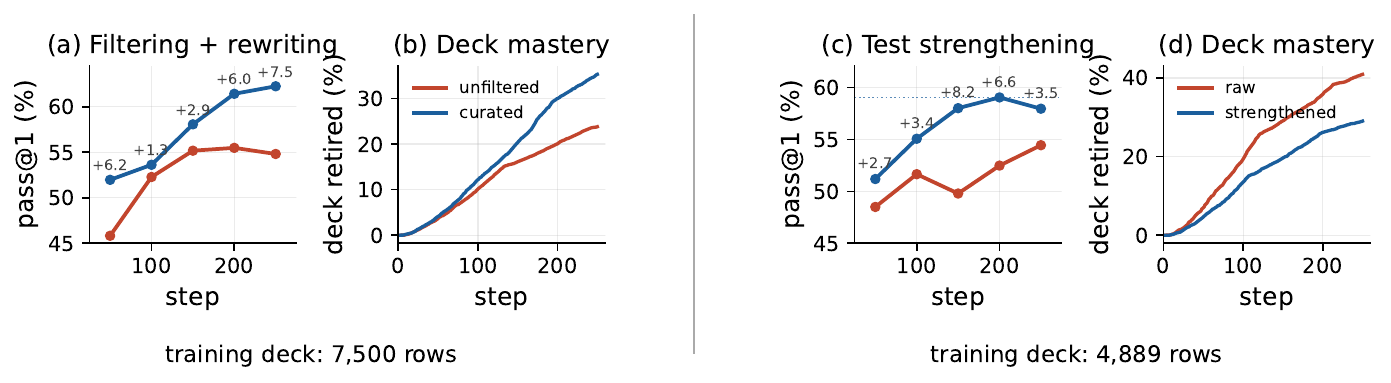}
  \caption{\textbf{Both curation steps pay off on held-out evaluation.} Each
    panel is a single-knob comparison in which only the training deck changes;
    every row in every arm passes standard fail-to-pass validation.
    \textbf{(a,~c)}~SWE-bench Verified pass@1 (483 problems $\times$ 4
    rollouts), with margins annotated at matched steps: the curated deck
    against a size- and mixture-matched unfiltered control~(a), and strengthened
    against raw suites over the \emph{identical} 4{,}889 instance IDs~(c).
    \textbf{(b,~d)}~Fraction of the deck retired as \emph{mastered}: solved by
    all eight rollouts in its group, then never sampled again. Panels (a,~b)
    accompany Section~\ref{sec:filtering} and (c,~d) Section~\ref{sec:hardening}.}
  \label{fig:data-ablations}
\end{figure}
\textbf{Experimental validation.}
Curation is tested by a controlled experiment in which only the training data
changes, with both arms scored by pass@1 on held-out SWE-bench Verified
(Figure~\ref{fig:data-ablations}a,b). The control is size- and
mixture-matched: 7{,}500 rows that \emph{clear pipeline~1 but are rejected by
pipeline~2}, kept in their original non-rewritten form, so the comparison
isolates exactly what pipeline~2 adds on top of the standard gate
(Appendix~\ref{app:filter_experiment}). The curated deck leads at all five
matched steps. It also learns its deck faster, retiring 35.3\% of its rows as
mastered by step 250 against 23.8\% for the control---a training-side
quantity, and the one that explains the evaluation gap. Curation therefore does
not just remove work: the control keeps spending rollouts on tasks it can probably 
never master---possibly because of an underspecified issue or a mis-specified
grading criterion---while the curated arm converts the same budget into solved
tasks that are then evicted, so more of the compute can be used on tasks that can
still teach it something.

Alignment is expensive: 10{,}361 validated tasks are rejected, 9{,}379 of them
on their tests rather than their content. The largest class,
\texttt{too\_loose}, is defective only in its test grading. Section~\ref{sec:hardening} asks
whether those rows can be recovered by repairing the tests instead of
discarding the task, which also turns the label from a model judgment into a
claim a controlled experiment can test.

\subsection{Repairing the tests: strengthening too-loose suites}
\label{sec:hardening}

Repairing a defective test suite in place is what pipeline~2 did not
attempt, and dropping these rows instead is the single largest source of data
loss in our pipeline. The issue, gold patch, container, and buggy tree stay
fixed while the graded tests are strengthened to reject the shortcuts that beat
them, subject to gold still passing and to every added check following from the
issue or established public behavior rather than from gold's implementation.
Strengthening suites rather than discarding their tasks is established practice
for evaluation benchmarks but untested as a training intervention
(Appendix~\ref{app:related}). The procedure recovers 4{,}909 of the 5{,}351 suites
(91.7\%), of which 4{,}889 form paired training arms holding \emph{identical}
instance IDs and differing in nothing but the unit test suite that computes reward
(Appendix~\ref{app:hardening}).

\textbf{Experimental validation.}
The two arms hold \emph{identical} rows and differ only in the test suite that
grades them, so the comparison isolates the reward signal itself
(Figure~\ref{fig:data-ablations}c,d). The strengthened arm leads at every
matched step: its step-200 score of $59.1$ exceeds every raw score through step
350 ($55.5$), so strengthening is worth more than 150 training steps. It also
masters its deck more slowly---29.0\% of rows retired by step 250 against
40.9\% for raw---which is the expected consequence of a stronger grader: a
rollout that previously collected full reward without fixing the bug no longer
passes, so tasks are no longer retired on the strength of false successes.
Because the rows are identical, that performance gap is not a difficulty
difference but an estimate of how much of the raw arm's mastery was spurious,
and the arm that masters less is the one that transfers better.


\section{Training and evaluation on the curated tasks}
\label{sec:main-run}

Section~\ref{sec:data} argued that a training task is only as good as the
agreement between its issue, its tests and its gold patch, and measured that
agreement one curation step at a time. This section spends the resulting training data
on a single long run and reports what it buys on held-out evaluation.  We describe the evaluation and the sandbox it runs in
before the training configuration, because the metric and two of its
constraints apply to training rollouts as well.

\subsection{Evaluation under sandbox isolation}
\label{sec:main-eval}

We evaluate on SWE-bench Verified~\citep{swebench2023} and SWE-bench
Pro~\citep{swebenchpro}, reporting pass@1 averaged over 4 rollouts per
task. We use Verified as our validation set and use Pro as the test set, as the latter is much slower to evaluate (Appendix~\ref{sec:app_eval_setup}).

\begin{figure}[t]
\centering

\begin{minipage}[t]{0.31\linewidth}
\vspace{0pt}
\centering

\captionof{table}{\textbf{Evaluation hacking, ours vs.\ default.}
Qwen3.8-27B, 60 SWE-bench Pro trials. Our setting prunes
\texttt{.git} and denies egress; default leaves both open.
Different routes can overlap.}
\label{tab:hack_audit}

\resizebox{\linewidth}{!}{%
\begin{tabular}{@{}lrr@{}}
\toprule
Setting & \textbf{Default} & \textbf{Ours} \\
\midrule
Total trials & 60 & 60 \\
Resolved & 39 & 30 \\
\midrule
\textbf{Hacked} & \textbf{19} & \textbf{0} \\
\quad gold SHA from \texttt{.git} & 17 & 0 \\
\quad source via network & 7 & 0 \\
\quad gold tests read & 13 & 0 \\
\midrule
Failed net attempts & 0 & 5 \\
\bottomrule
\end{tabular}%
}

\end{minipage}
\hfill
\begin{minipage}[t]{0.68\linewidth}
\vspace{0pt}
\centering

\includegraphics[width=\linewidth]{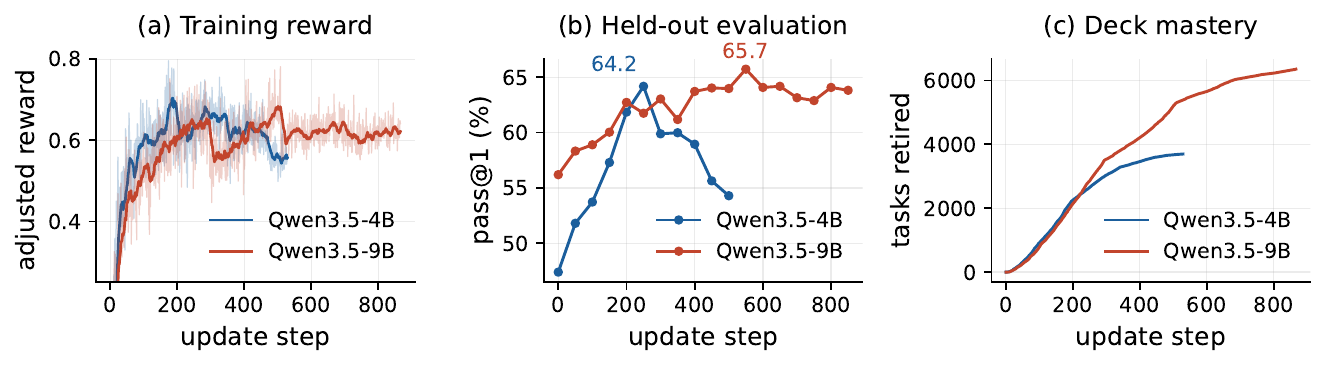}

\captionof{figure}{\textbf{The 4B degrades past step 250 while its
training reward barely moves; the 9B, on the same deck and the same
eviction rule, does not.}
\textbf{(a)}~Mean reward on completed rollouts, eviction-adjusted:
retired tasks are imputed reward $1$ so that a deck which keeps getting
harder does not depress the mean.
\textbf{(b)}~SWE-bench Verified pass@1 with the best accuracy
annotated.
\textbf{(c)}~Fraction of the $7{,}500$-row deck retired as mastered.}
\label{fig:main-run}

\end{minipage}

\end{figure}

The environment those rollouts run in needs more explanation than the metric
does, because an agent with a shell in a repository has two ways to perform \textbf{evaluation hacking} that do not involve solving the task: (1) the repository's own git history,
in which the gold-fix commit is a descendant of the checked-out buggy parent
and so is reachable offline by \texttt{git log --all}, and (2) the network, which
offers the upstream fix and arbitrary package installation. On 60 SWE-bench Pro trials
(Table~\ref{tab:hack_audit}, Appendix~\ref{sec:app_hack_audit}), a Qwen3.8-27B
agent in the default setting hacks 19 of them, which inflates its resolve rate from
50.0\% to 65.0\%. The same
audit finds no hacks under our setting. Similar to concurrent work at
Cursor~\citep{cursorhack}, we therefore prune the history when each container
is prepared and run every sandbox under an egress policy denying outbound
traffic and cloud metadata access. Training and evaluation rollouts face the
same isolation. 

Because isolation grades the gold patch too, a few tasks become ungradeable;
excluding those and a few unrelated defects leaves $\mathbf{483}$ SWE-bench
Verified problems, fixed across runs and checkpoints, and all $\mathbf{731}$
SWE-bench Pro problems (Appendix~\ref{app:integrity}). \textbf{These controls lower our scores, so any comparison against numbers
reported under the default setting understates our results.}


\subsection{Training setup}
\label{sec:main-setup}

We train two base models, Qwen3.5-4B and Qwen3.5-9B, on the same curated pool of Section~\ref{sec:data} drawn from R2E-Gym, SWE-rebench~V2, and
Scale-SWE, each prepared with its git
history pruned as in Section~\ref{sec:main-eval}. All three training decks are screened against SWE-bench Verified to avoid contamination (Appendix~\ref{app:contamination}).  The objective is the CISPO-style loss introduced in Appendix~\ref{app:setup} without KL penalty, taking $16$ prompts
with n=$8$, and no length penalty is applied.
We also apply token-level importance masking in the style of
IcePop~\citep{ring1t} with range $[0, 2.0]$. Beyond the data, three small settings
differ from Section~\ref{sec:recipe}: we retire only mastered tasks; we
override the reward of invalid trajectories to $-1$ (Appendix~\ref{app:stop-policy}); and we disable the duplicate-tool-call
detector. The 9B run additionally updates three training settings, which
Appendix~\ref{app:9b-config} states in full.
Appendix~\ref{app:training_details} gives the deck, budgets, and optimizer
settings in full, with the objective in Appendix~\ref{app:setup} and the
reward contract in Appendix~\ref{app:stop-policy}.

\subsection{Results}
\label{sec:main-results}
Table~\ref{tab:baselines} reports each checkpoint under its own sampling
configuration (Appendix~\ref{app:eval-sampling}). The Qwen3.5-4B model reaches
65.0\% pass@1 on SWE-bench Verified, up from 47.4\% at initialization and past
the 62.0\% that Section~\ref{sec:recipe} reached on R2E-Gym alone, and the 9B
reaches 65.7\%. The gains hold on the harder benchmark: SWE-bench Pro-v1/v2
rise from 25.0/37.9 to 38.3/59.7 for the 4B and from 31.9/48.8 to 39.6/70.2 for
the 9B. Because our training data contains only Python tasks, we also report
Pro-v2-Py, the Python subset of Pro-v2. The gain there, 21.5 points for the 4B
and 18.4 for the 9B, is no larger than on Pro-v2 as a whole (21.8 and 21.4), so
the improvement is not confined to the language the run was trained on.

Figure~\ref{fig:main-run}b tracks Verified across training under a single fixed
configuration ($T{=}1.0$, top-$p$ 0.95, top-$k$ 20) rather than the
per-checkpoint settings of Table~\ref{tab:baselines}. There the 4B peaks at 64.2
at step 250 and the 9B at 65.7 at step 550; over the 200 steps following its
peak the 9B gives back 2.8 points, where the 4B gives back 8.6. 

That decline does not register in the 4B's training signal. Over steps 250 to
450, as pass@1 falls 13.4\% relative, the eviction-adjusted reward declines
only 3.4\%. That metric imputes reward 1 for every retired task
(Appendix~\ref{app:eviction-metric}), so it is blind to the failure by
construction: crediting a task that is never re-checked assumes its
competence is preserved.

We test this directly, and the assumption fails. By step 350 eviction has
retired 44\% of the 4B's deck, rising to 48\% by step 450
(Figure~\ref{fig:main-run}c), so we re-scored a sample of the tasks retired by
step 350 under frozen weights (Table~\ref{tab:forgetting}). At step 350 they
score 0.95, close to the 1.0 they earned when they left, so eviction had been
retiring genuinely mastered tasks rather than lucky ones. The same tasks score
0.85 at step 500. The 4B is forgetting the population it no longer trains on,
and the training reward cannot see it (Appendix~\ref{app:eviction}).

The 9B does not. We run the same probe on a 9B trained on the 4B's exact deck
and settings, changing only the base model (Appendix~\ref{app:9b-config}). On
decks cut from its own eviction state, which had retired 41\% of the deck by
step 350 against the 4B's 44\%, it scores 0.97 at step 350 and 0.97 at step
500 --- a decline statistically indistinguishable from zero. Deck depletion
alone is therefore not sufficient to produce the degradation. Our hypothesis
is that the Qwen3.5-4B lacks the capacity to retain a growing mastered
population while continuing to fit the residual tail.
Table~\ref{tab:baselines} places the run against published results on
SWE-bench Verified and SWE-bench Pro. The comparison is indicative rather than
controlled: the systems differ in scaffold, inference budget and the amount of
supervision behind them, and only the rows marked with $\dagger$ share this
paper's evaluation harness. To make at least one comparison direct, we
re-evaluated the concurrent FrogNano~\citep{frognano2026} in our harness under
the sampling parameters its authors report. It scores 60.4 on Verified and 37.0
on Pro-v1, against the 61.5 and 37.6 its authors report (they do not report
Pro-v2). 

\definecolor{row4b}{HTML}{EAF1F8}   
\definecolor{row9b}{HTML}{FBEDE9}   

\begin{table}[t]
\centering

\begin{minipage}[t]{0.26\linewidth}
\vspace{0pt}
\centering

\captionof{table}{\textbf{The 4B forgets what it mastered; the 9B
does not.} Mean reward on tasks eviction had already retired, re-scored
at step 350 and 500 checkpoints. Each arm is measured on its own eviction state at
step 350, 300 tasks per source.}
\label{tab:forgetting}

\resizebox{\linewidth}{!}{%
\begin{tabular}{@{}llrrr@{}}
\toprule
& Retired tasks & 350 & 500 & $\Delta$ \\
\midrule
\multirow{3}{*}{\rotatebox{90}{\emph{4B}}}
& R2E-Gym      & 0.952 & 0.851 & $\mathbf{-0.101}$ \\
& SWE-rebench  & 0.946 & 0.845 & $\mathbf{-0.101}$ \\
& Scale-SWE    & 0.968 & 0.852 & $\mathbf{-0.117}$ \\
\midrule
\multirow{3}{*}{\rotatebox{90}{\emph{9B}}}
& R2E-Gym      & 0.973 & 0.968 & $-0.005$ \\
& SWE-rebench  & 0.963 & 0.951 & $-0.013$ \\
& Scale-SWE    & 0.970 & 0.967 & $-0.003$ \\
\bottomrule
\end{tabular}
}

\end{minipage}
\hfill
\begin{minipage}[t]{0.73\linewidth}
\vspace{0pt}
\centering

\captionof{table}{\textbf{SWE-bench Verified, Pro-v1/v2, and Pro-v2-Py pass@1.}
Pro-v2-Py is the Python subset of Pro-v2, the only language in our training
data. Rows marked $^{\dagger}$ were evaluated in our isolated sandbox, each
under its own sampling configuration (Appendix~\ref{app:eval-sampling});
others are as reported by their authors. The isolation of
Section~\ref{sec:main-eval} can only lower scores relative to an unisolated
harness. \colorbox{row4b}{\strut 4B} \colorbox{row9b}{\strut 9B}.}
\label{tab:baselines}

\resizebox{\linewidth}{!}{%
\begin{tabular}{llcccc}
\toprule
Model & Scaffold & Verified & Pro-v1 & Pro-v2 &Pro-v2-Py\\
\midrule
\multicolumn{4}{l}{\emph{Baselines}} \\
\rowcolor{row4b}
Qwen3.5-4B$^{\dagger}$ & Mini-SWE-Agent-Plus & 47.4 & 25.0 & 37.9 & 46.6 \\
\rowcolor{row9b}
Qwen3.5-9B$^{\dagger}$ & Mini-SWE-Agent-Plus & 56.2 & 31.9 & 48.8 & 56.5 \\
\midrule
\multicolumn{4}{l}{\emph{Other works}} \\
\rowcolor{row4b}
Polar-Qwen3.5-4B~\citep{xu2026polaragenticrlharness}
& Pi & 40.4 & - & - & - \\
\rowcolor{row4b}
AxiaoDBL-4B~\citep{axiaodbl2026qwen35swebench}
& Mini-SWE-Agent & 55.0 & - & - & - \\
\rowcolor{row4b}
FrogNano-Qwen3.5-4B~\citep{frognano2026}
& Leaf & 61.5 & 37.6 & - & - \\
\rowcolor{row4b}
FrogNano-Qwen3.5-4B (Our setting)$^{\dagger}$
& Leaf & 60.4 & 37.0 & 56.8 & 64.3 \\
\rowcolor{row9b}
AgentLightning-Qwen3.5-9B~\citep{he2026agentlightningv10harnessed}
& Mini-SWE-Agent & 56.4 & - & - & - \\
\rowcolor{row9b}
Dockerless-Qwen3.5-9B~\citep{zeng2026dockerlessenvironmentfreeprogramverifier}
& OpenHands & 62.0 & 35.2 & - & - \\
\midrule
\rowcolor{row4b}
\ours-4B, R2E-Gym only (Section\ref{sec:recipe})$^{\dagger}$
& Mini-SWE-Agent-Plus & 62.0 & 38.3 & \textbf{60.5} & 64.8 \\
\rowcolor{row4b}
\ours-4B + curated RL$^{\dagger}$
& Mini-SWE-Agent-Plus & \textbf{65.0} & \textbf{38.3} & 59.7 & \textbf{68.1}\\
\rowcolor{row9b}
\ours-9B + curated RL$^{\dagger}$
& Mini-SWE-Agent-Plus & \textbf{65.7} & \textbf{39.6} & \textbf{70.2} &  \textbf{74.9} \\
\bottomrule
\end{tabular}%
}

\end{minipage}

\end{table}
\section{Discussion and conclusion}
Strong RL baselines for long-horizon agents remain hard to reproduce, so gains from new
algorithms or data are hard to separate from an under-tuned baseline. We treat prolonged
overfitting on a small, solvable task pool as the routine check such a baseline has lacked: it
exposes failures that early reward gains conceal, and it guides which established components the
recipe needs. In asynchronous RL these components span modeling and systems, because system
efficiency, cached state and numerical paths all change the experience the optimizer sees. Gated
reward, importance-weight control, aligned training and inference engines, and consistent state
across policy updates are each known, but together they form the missing foundation.

That foundation alone trains Qwen3.5-4B to 62.0\% pass@1 on SWE-bench Verified from unchanged
R2E-Gym tasks, comparable to FrogNano's 61.5\% with online task synthesis, after about one-sixth
as many training rollouts. On top of it, data work becomes measurable: curation and test strengthening
raise the 4B model to 65.0\%/38.3\%/59.7\% and the 9B model to 65.7\%/39.6\%/70.2\% on SWE-bench Verified/Pro-v1/Pro-v2.
Most comparisons are single-seed and use one harness, and the Verified subset also serves for
checkpoint selection. \ours{} offers a documented starting point against which future algorithms,
data and environments for asynchronous agent RL can be measured.
\label{main-text-end}

\clearpage
\phantomsection
\label{references-start}
\bibliography{iclr2027_conference}

@misc{agentfly_swe_2026,
  author = {{AgentFly Contributors}},
  title = {{SWE} task setup and {Qwen3-32B} training example},
  year = {2026},
  howpublished = {Repository documentation},
  note = {Revision \texttt{1256586}; accessed September 16, 2026},
  url = {https://github.com/Agent-One-Lab/AgentFly/blob/1256586b1109ba8e0dc0f179f8515b4567d09df4/docs/examples/swe.md},
}

@article{auditing2026,
  author = {Shreshth Rajan},
  title = {Auditing Reward Hackability in Code {RL} Training Environments},
  journal = {arXiv preprint arXiv:2606.16062},
  year = {2026},
  eprint = {2606.16062},
  archivePrefix = {arXiv},
  primaryClass = {cs.AI},
  url = {https://arxiv.org/abs/2606.16062},
}

@article{canopy2026,
  author = {Liming Pu and Xiaoxia Li and Yifu Liu and Teng Cao and Bin Yang},
  title = {Explore More, Drift Less: Outcome-Only Reinforcement Learning Can Suffice for Long-Horizon Interactive Agents},
  journal = {arXiv preprint arXiv:2609.01245},
  year = {2026},
  eprint = {2609.01245},
  archivePrefix = {arXiv},
  primaryClass = {cs.LG},
  url = {https://arxiv.org/abs/2609.01245},
}

@article{cispo2025,
  author = {{MiniMax}},
  title = {{MiniMax-M1}: Scaling Test-Time Compute Efficiently with Lightning Attention},
  journal = {arXiv preprint arXiv:2506.13585},
  year = {2025},
  eprint = {2506.13585},
  archivePrefix = {arXiv},
  primaryClass = {cs.CL},
  url = {https://arxiv.org/abs/2506.13585},
}

@article{dapo2025,
  author = {Qiying Yu and Zheng Zhang and Ruofei Zhu and Yufeng Yuan and Xiaochen Zuo and Yu Yue and
    Weinan Dai and Tiantian Fan and Gaohong Liu and Lingjun Liu and Xin Liu and Haibin Lin and Zhiqi
    Lin and Bole Ma and Guangming Sheng and Yuxuan Tong and Chi Zhang and Mofan Zhang and Wang Zhang
    and Hang Zhu and Jinhua Zhu and Jiaze Chen and Jiangjie Chen and Chengyi Wang and Hongli Yu and
    Yuxuan Song and Xiangpeng Wei and Hao Zhou and Jingjing Liu and Wei-Ying Ma and Ya-Qin Zhang and
    Lin Yan and Mu Qiao and Yonghui Wu and Mingxuan Wang},
  title = {{DAPO}: An Open-Source {LLM} Reinforcement Learning System at Scale},
  journal = {arXiv preprint arXiv:2503.14476},
  year = {2025},
  eprint = {2503.14476},
  archivePrefix = {arXiv},
  primaryClass = {cs.LG},
  url = {https://arxiv.org/abs/2503.14476},
}

@misc{grpo,
      title={DeepSeekMath: Pushing the Limits of Mathematical Reasoning in Open Language Models}, 
      author={Zhihong Shao and Peiyi Wang and Qihao Zhu and Runxin Xu and Junxiao Song and Xiao Bi and Haowei Zhang and Mingchuan Zhang and Y. K. Li and Y. Wu and Daya Guo},
      year={2024},
      eprint={2402.03300},
      archivePrefix={arXiv},
      primaryClass={cs.CL},
      url={https://arxiv.org/abs/2402.03300}, 
}

@misc{deepswe2025,
  author = {{Agentica} and {Together AI}},
  title = {{DeepSWE}: Training a state-of-the-art coding agent from scratch by scaling {RL}},
  year = {2025},
  howpublished = {Technical report and training release},
  url = {https://github.com/agentica-project/rllm/blob/7b47687f6a9ef1bf5cbd56dd1af61fff08c4b0e4/examples/swe/README.md},
}

@misc{dressage_balance_2026,
  author = {{Dressage Contributors}},
  title = {Rollout load balancing in {Dressage}: From sticky session to step balance},
  year = {2026},
  howpublished = {Repository technical report},
  note = {Revision \texttt{3e3142f}; accessed September 16, 2026},
  url = {https://github.com/Accio-Lab/Dressage/blob/3e3142fe8ea07e4504c3b20a936a4c201a3de44c/docs/engine-rebalancing-report-en.md},
}

@misc{dressage_staleness_2026,
  author = {{Dressage Contributors}},
  title = {Staleness control},
  year = {2026},
  howpublished = {Repository technical report},
  note = {Revision \texttt{3e3142f}; accessed September 16, 2026},
  url = {https://github.com/Accio-Lab/Dressage/blob/3e3142fe8ea07e4504c3b20a936a4c201a3de44c/docs/staleness.md},
}

@misc{dressage_swe_2026,
  author = {{Dressage Contributors}},
  title = {Blackbox {SWE-Gym} experiment with {Claude Code}},
  year = {2026},
  howpublished = {Repository technical report},
  note = {Revision \texttt{3e3142f}; accessed September 16, 2026},
  url = {https://github.com/Accio-Lab/Dressage/blob/3e3142fe8ea07e4504c3b20a936a4c201a3de44c/docs/blackbox-swegym-claude-code-experiment-en.md},
}

@article{errorbars2024,
  author = {Evan Miller},
  title = {Adding Error Bars to Evals: A Statistical Approach to Language Model Evaluations},
  journal = {arXiv preprint arXiv:2411.00640},
  year = {2024},
  eprint = {2411.00640},
  archivePrefix = {arXiv},
  primaryClass = {stat.AP},
  url = {https://arxiv.org/abs/2411.00640},
}

@article{frognano2026,
  author = {Minseon Kim and Zhengyan Shi and Emiliano Penaloza and Christopher Cui and Roger Creus
    Castanyer and Maryam Hashemzadeh and Isadora White and Jonathan Light and Jeonghye Kim and
    Matheus Pereira and Darya Moldavskaya and Chinmay Singh and Fabio Vera and Baolin Peng and
    Xingdi Yuan and Marc-Alexandre C{\^o}t{\'e} and Alessandro Sordoni},
  title = {{FrogNano}: Training a {4B} Coding Agent via Online Task Synthesis},
  journal = {arXiv preprint arXiv:2609.07925v3},
  year = {2026},
  eprint = {2609.07925},
  archivePrefix = {arXiv},
  primaryClass = {cs.AI},
  url = {https://arxiv.org/html/2609.07925v3},
}

@article{gspo2025,
  author = {Chujie Zheng and Shixuan Liu and Mingze Li and Xiong-Hui Chen and Bowen Yu and Chang Gao
    and Kai Dang and Yuqiong Liu and Rui Men and An Yang and Jingren Zhou and Junyang Lin},
  title = {Group Sequence Policy Optimization},
  journal = {arXiv preprint arXiv:2507.18071},
  year = {2025},
  eprint = {2507.18071},
  archivePrefix = {arXiv},
  primaryClass = {cs.LG},
  url = {https://arxiv.org/abs/2507.18071},
}

@article{magistral2025,
  author = {{Mistral AI}},
  title = {Magistral},
  journal = {arXiv preprint arXiv:2506.10910},
  year = {2025},
  eprint = {2506.10910},
  archivePrefix = {arXiv},
  primaryClass = {cs.CL},
  url = {https://arxiv.org/abs/2506.10910},
}

@misc{miles_swe_2026,
  author = {{RadixArk}},
  title = {{SWE-Agent} training with {Harbor} on {Docker} sandboxes},
  year = {2026},
  howpublished = {Miles repository},
  note = {Revision \texttt{f6d0257}; accessed September 16, 2026},
  url = {https://github.com/radixark/miles/blob/f6d0257d83b7c14f4ce43ecfcd71d955112f6e0e/examples/swe-agent-harbor-docker/README.md},
}

@techreport{mimo2026,
  author = {{Xiaomi MiMo Team}},
  title = {{MiMo-V2.6}: Scaling reinforcement learning towards self-improvement},
  institution = {Xiaomi},
  type = {Technical report},
  year = {2026},
  month = {September},
  url = {https://huggingface.co/XiaomiMiMo/MiMo-V2.6-Pro-RL/blob/73875d00b30a89ef8cc353a0b60b0e9f9561952d/MiMo_V2_6_technical_report.pdf},
}

@article{nebius_swe_2025,
  author = {Alexander Golubev and Maria Trofimova and Sergei Polezhaev and Ibragim Badertdinov and
    Maksim Nekrashevich and Anton Shevtsov and Simon Karasik and Sergey Abramov and Andrei
    Andriushchenko and Filipp Fisin and Sergei Skvortsov and Boris Yangel},
  title = {Training Long-Context, Multi-Turn Software Engineering Agents with Reinforcement Learning},
  journal = {arXiv preprint arXiv:2508.03501v2},
  year = {2025},
  eprint = {2508.03501},
  archivePrefix = {arXiv},
  primaryClass = {cs.LG},
  url = {https://arxiv.org/html/2508.03501v2},
}

@misc{nemo_swe_2026,
  author = {{NVIDIA NeMo}},
  title = {Two-stage {SWE} {RL} for {Qwen3-30B-A3B-Thinking}},
  year = {2026},
  howpublished = {Repository guide and configuration},
  note = {Revision \texttt{1cef394}; accessed September 16, 2026},
  url = {https://github.com/NVIDIA-NeMo/RL/blob/1cef394359aeddf1eaf4f1d209208be7b6ec0b1a/docs/guides/swe-rl-qwen3.md},
}

@misc{primerl_qwen_swe_2026,
  author = {{PrimeIntellect}},
  title = {{Qwen3-30B-A3B-Thinking-2507} {SWE} example},
  year = {2026},
  howpublished = {{Prime-RL} repository},
  note = {Revision \texttt{d7339fc}; accessed September 16, 2026},
  url = {https://github.com/PrimeIntellect-ai/prime-rl/blob/d7339fcf250eb7e015c6c144bcf411e7178221e3/examples/advanced/qwen3-30b-a3b/swe.toml},
}

@misc{primerl_swe_2026,
  author = {{PrimeIntellect}},
  title = {{GLM-4.5-Air} {SWE} training example},
  year = {2026},
  howpublished = {{Prime-RL} repository},
  note = {Revision \texttt{9f4eef8}; accessed September 16, 2026},
  url = {https://github.com/PrimeIntellect-ai/prime-rl/blob/9f4eef81a3e951b77d66c709335b0024a499db4d/examples/advanced/glm-4.5-air/swe.toml},
}

@article{sao2026,
  author = {Zhenyu Hou and Yujiang Li and Jie Tang and Yuxiao Dong},
  title = {Single-Rollout Asynchronous Optimization for Agentic Reinforcement Learning},
  journal = {arXiv preprint arXiv:2607.07508},
  year = {2026},
  eprint = {2607.07508},
  archivePrefix = {arXiv},
  primaryClass = {cs.LG},
  url = {https://arxiv.org/abs/2607.07508},
}

@misc{skyrl_mini_2026,
  author = {{SkyRL Contributors}},
  title = {Guide: {Mini-SWE-Agent} + SkyRL},
  year = {2026},
  howpublished = {Repository documentation},
  note = {Revision \texttt{53b1155}; accessed September 16, 2026},
  url = {https://github.com/NovaSky-AI/SkyRL/blob/53b11553a3e1ea6b4156fdc36d0322823b679270/examples/train/mini_swe_agent/README.md},
}

@article{skyrlagent2025,
  author = {Shiyi Cao and Dacheng Li and Fangzhou Zhao and Shuo Yuan and Sumanth R. Hegde and Connor
    Chen and Charlie Ruan and Tyler Griggs and Shu Liu and Eric Tang and Richard Liaw and Philipp
    Moritz and Matei Zaharia and Joseph E. Gonzalez and Ion Stoica},
  title = {{SkyRL-Agent}: Efficient {RL} Training for Multi-turn {LLM} Agent},
  journal = {arXiv preprint arXiv:2511.16108},
  year = {2025},
  eprint = {2511.16108},
  archivePrefix = {arXiv},
  primaryClass = {cs.AI},
  url = {https://arxiv.org/abs/2511.16108},
}

@misc{slime_swe_2026,
  author = {{THUDM}},
  title = {Coding-Agent RL},
  year = {2026},
  howpublished = {slime repository},
  note = {Revision \texttt{4c193f1}; accessed September 16, 2026},
  url = {https://github.com/THUDM/slime/blob/4c193f1f37509cca70f0e88807a9305b70f63f4e/examples/coding_agent_rl/README.md},
}

@inproceedings{soberlook2025,
  author = {Andreas Hochlehnert and Hardik Bhatnagar and Vishaal Udandarao and Samuel Albanie and
    Ameya Prabhu and Matthias Bethge},
  title = {A Sober Look at Progress in Language Model Reasoning: Pitfalls and Paths to Reproducibility},
  year = {2025},
  eprint = {2504.07086},
  archivePrefix = {arXiv},
  primaryClass = {cs.LG},
  url = {https://arxiv.org/abs/2504.07086},
  booktitle = {Conference on Language Modeling (COLM)},
}

@article{swebench2023,
  author = {Carlos E. Jimenez and John Yang and Alexander Wettig and Shunyu Yao and Kexin Pei and
    Ofir Press and Karthik Narasimhan},
  title = {{SWE-bench}: Can Language Models Resolve Real-World {GitHub} Issues?},
  journal = {arXiv preprint arXiv:2310.06770},
  year = {2023},
  eprint = {2310.06770},
  archivePrefix = {arXiv},
  primaryClass = {cs.CL},
  url = {https://arxiv.org/abs/2310.06770},
}

@article{pipelinerl,
  author        = {Pich{\'e}, Alexandre and Kamalloo, Ehsan and Pardinas, Rafael and Chen, Xiaoyin and Bahdanau, Dzmitry},
  title         = {{PipelineRL}: Faster On-policy Reinforcement Learning for Long Sequence Generation},
  journal = {arXiv preprint arXiv:2509.19128},
  year          = {2025},
  eprint        = {2509.19128},
  archivePrefix = {arXiv},
  primaryClass  = {cs.LG},
  url           = {https://arxiv.org/abs/2509.19128}
}

@misc{swebench465,
  author = {J. Kahn and F. Kreuk and J. Wei and J. Kossen and E. McMilin},
  title = {Repo state loopholes during agentic evaluation},
  year = {2025},
  howpublished = {{SWE-bench} issue \#465},
  note = {Fixed in pull request \#471},
}

@article{swegym2024,
  author = {Jiayi Pan and Xingyao Wang and Graham Neubig and Navdeep Jaitly and Heng Ji and Alane
    Suhr and Yizhe Zhang},
  title = {Training Software Engineering Agents and Verifiers with {SWE-Gym}},
  journal = {arXiv preprint arXiv:2412.21139},
  year = {2024},
  eprint = {2412.21139},
  archivePrefix = {arXiv},
  primaryClass = {cs.SE},
  url = {https://arxiv.org/abs/2412.21139},
}

@article{swemaster2026,
  author = {Huatong Song and Lisheng Huang and Shuang Sun and Jinhao Jiang and Ran Le and Daixuan
    Cheng and Guoxin Chen and Yiwen Hu and Zongchao Chen and Wayne Xin Zhao and Yang Song and Tao
    Zhang and Ji-Rong Wen},
  title = {{SWE-Master}: Unleashing the Potential of Software Engineering Agents via Post-Training},
  journal = {arXiv preprint arXiv:2602.03411v1},
  year = {2026},
  eprint = {2602.03411},
  archivePrefix = {arXiv},
  primaryClass = {cs.SE},
  url = {https://arxiv.org/html/2602.03411v1},
}

@misc{yao2025offpolicy,
  author = {Feng Yao and Liyuan Liu and Dinghuai Zhang and Chengyu Dong and Jingbo Shang and
    Jianfeng Gao},
  title = {Your Efficient {RL} Framework Secretly Brings You Off-Policy {RL} Training},
  howpublished = {Technical blog},
  year = {2025},
  url = {https://fengyao.notion.site/off-policy-rl},
}

@inproceedings{r2egym,
  title={R2e-gym: Procedural environment generation and hybrid verifiers for scaling open-weights SWE agents},
  author={Jain, Naman and Singh, Jaskirat and Shetty, Manish and Zhang, Tianjun and Zheng, Liang and Sen, Koushik and Stoica, Ion},
  booktitle={Second Conference on Language Modeling},
  year={2025}
}

@article{scaleswe,
  title={Immersion in the github universe: Scaling coding agents to mastery},
  author={Zhao, Jiale and Chen, Guoxin and Meng, Fanzhe and Li, Minghao and Chen, Jie and Xu, Hui and Sun, Yongshuai and Zhao, Wayne Xin and Song, Ruihua and Zhang, Yuan and others},
  journal={arXiv preprint arXiv:2602.09892},
  year={2026}
}

@article{rebenchv2,
  title={Swe-rebench v2: Language-agnostic swe task collection at scale},
  author={Badertdinov, Ibragim and Nekrashevich, Maksim and Shevtsov, Anton and Golubev, Alexander},
  journal={arXiv preprint arXiv:2602.23866},
  year={2026}
}

@misc{qwen38,
    title = {{Qwen3.8-Max}: A New Bar for Coding and Cowork},
    url = {https://qwen.ai/blog?id=qwen3.8},
    author = {{Qwen Team}},
    month = {August},
    year = {2026}
}

@misc{rebench,
      title={SWE-rebench: An Automated Pipeline for Task Collection and Decontaminated Evaluation of Software Engineering Agents}, 
      author={Ibragim Badertdinov and Alexander Golubev and Maksim Nekrashevich and Anton Shevtsov and Simon Karasik and Andrei Andriushchenko and Maria Trofimova and Daria Litvintseva and Boris Yangel},
      year={2025},
      eprint={2505.20411},
      archivePrefix={arXiv},
      primaryClass={cs.SE},
      url={https://arxiv.org/abs/2505.20411}
}

@inproceedings{utboost,
    title = "{UTB}oost: Rigorous Evaluation of Coding Agents on {SWE}-Bench",
    author = "Yu, Boxi  and
      Zhu, Yuxuan  and
      He, Pinjia  and
      Kang, Daniel",
    editor = "Che, Wanxiang  and
      Nabende, Joyce  and
      Shutova, Ekaterina  and
      Pilehvar, Mohammad Taher",
    booktitle = "Proceedings of the 63rd Annual Meeting of the Association for Computational Linguistics (Volume 1: Long Papers)",
    month = jul,
    year = "2025",
    address = "Vienna, Austria",
    publisher = "Association for Computational Linguistics",
    url = "https://aclanthology.org/2025.acl-long.189/",
    doi = "10.18653/v1/2025.acl-long.189",
    pages = "3762--3774",
    ISBN = "979-8-89176-251-0"
}

@misc{probe,
      title={Probe to Generate: Program Variant-Guided Test Augmentation for Repository-Level Repair Benchmarks}, 
      author = {Chenglin Li and Yisen Xu and Zehao Wang and Shin Hwei Tan and Tse-Hsun {Peter} Chen},
      year={2026},
      eprint={2604.01518},
      archivePrefix={arXiv},
      primaryClass={cs.SE},
      url={https://arxiv.org/abs/2604.01518}, 
}

@misc{sweabs,
      title={SWE-ABS: Adversarial Benchmark Strengthening Exposes Inflated Success Rates on Test-based Benchmark}, 
      author={Boxi Yu and Yang Cao and Yuzhong Zhang and Liting Lin and Junjielong Xu and Zhiqing Zhong and Qinghua Xu and Guancheng Wang and Jialun Cao and Shing-Chi Cheung and Pinjia He and Lionel Briand},
      year={2026},
      eprint={2603.00520},
      archivePrefix={arXiv},
      primaryClass={cs.SE},
      url={https://arxiv.org/abs/2603.00520}, 
}

@misc{ring1t,
  title = {Small Leak Can Sink a Great Ship--Boost RL Training on MoE with IcePop!},
  url = {https://ringtech.notion.site/icepop},
  author = {Xin Zhao and Yongkang Liu and Kuan Xu and Jia Guo and Zihao Wang and Yan Sun and Xinyu Kong and Qianggang Cao and Liang Jiang and Zujie Wen and Zhiqiang Zhang and Jun Zhou},
  year = {2025},
  month = {Sep},
}

@misc{cursorhack,
  author       = {Jain, Naman},
  title        = {{Reward Hacking is Swamping Model Intelligence Gains}},
  howpublished = {Cursor Blog},
  month        = jun,
  year         = {2026},
  url          = {https://cursor.com/blog/reward-hacking-coding-benchmarks},
  urldate      = {2026-09-23},
}

@article{swebenchpro,
      title={{SWE-Bench Pro}: Can {AI} Agents Solve Long-Horizon Software Engineering Tasks?},
      journal={arXiv preprint arXiv:2509.16941}, 
      author={Xiang Deng and Jeff Da and Edwin Pan and Yannis Yiming He and Charles Ide and Kanak Garg and Niklas Lauffer and Andrew Park and Nitin Pasari and Chetan Rane and Karmini Sampath and Maya Krishnan and Srivatsa Kundurthy and Sean Hendryx and Zifan Wang and Vijay Bharadwaj and Jeff Holm and Raja Aluri and Chen Bo Calvin Zhang and Noah Jacobson and Bing Liu and Brad Kenstler},
      year={2025},
      eprint={2509.16941},
      archivePrefix={arXiv},
      primaryClass={cs.SE},
      url={https://arxiv.org/abs/2509.16941}, 
}

@article{fu2025areal,
  title   = {{AReaL}: A Large-Scale Asynchronous Reinforcement Learning System for Language Reasoning},
  author  = {Fu, Wei and Gao, Jiaxuan and Shen, Xujie and Zhu, Chen and Mei, Zhiyu and He, Chuyi and Xu, Shusheng and Wei, Guo and Mei, Jun and Wang, Jiashu and Yang, Tongkai and Yuan, Binhang and Wu, Yi},
  journal = {arXiv preprint arXiv:2505.24298},
  year    = {2025},
  eprint  = {2505.24298},
  archivePrefix = {arXiv},
  url     = {https://arxiv.org/abs/2505.24298}
}

@inproceedings{zeng2025acecoder,
  title   = {{ACECODER}: Acing Coder {RL} via Automated Test-Case Synthesis},
  author  = {Zeng, Huaye and Jiang, Dongfu and Wang, Haozhe and Nie, Ping and Chen, Xiaotong and Chen, Wenhu},
  booktitle = {Proceedings of the 63rd Annual Meeting of the Association for Computational Linguistics (Volume 1: Long Papers)},
  year    = {2025},
  month   = jul,
  address = {Vienna, Austria},
  publisher = {Association for Computational Linguistics},
  pages   = {12023--12040},
  doi     = {10.18653/v1/2025.acl-long.587},
  url     = {https://aclanthology.org/2025.acl-long.587/}
}

@article{he2025hardtests,
  title   = {{HardTests}: Synthesizing High-Quality Test Cases for {LLM} Coding},
  author  = {He, Zhongmou and Choi, Yee Man and Zhang, Kexun and Ji, Jiabao and Zhou, Junting and Xu, Dejia and Bercovich, Ivan and Zhang, Aidan and Li, Lei},
  journal = {arXiv preprint arXiv:2505.24098},
  year    = {2025},
  eprint  = {2505.24098},
  archivePrefix = {arXiv},
  url     = {https://arxiv.org/abs/2505.24098}
}

@article{ruan2026evolvecoder,
  title   = {{EvolveCoder}: Evolving Test Cases via Adversarial Verification for Code Reinforcement Learning},
  author  = {Ruan, Chi and Jiang, Dongfu and Zeng, Huaye and Nie, Ping and Chen, Wenhu},
  journal = {arXiv preprint arXiv:2603.12698},
  year    = {2026},
  eprint  = {2603.12698},
  archivePrefix = {arXiv},
  url     = {https://arxiv.org/abs/2603.12698}
}

@article{swesmith,
  title={Swe-smith: Scaling data for software engineering agents},
  author={Yang, John and Lieret, Kilian and Jimenez, Carlos and Wettig, Alexander and Khandpur, Kabir and Zhang, Yanzhe and Hui, Binyuan and Press, Ofir and Schmidt, Ludwig and Yang, Diyi},
  journal={Advances in Neural Information Processing Systems},
  volume={38},
  year={2025}
}

@inproceedings{engstrom2020implementation,
  author = {Logan Engstrom and Andrew Ilyas and Shibani Santurkar and Dimitris Tsipras and Firdaus Janoos and
    Larry Rudolph and Aleksander Madry},
  title = {Implementation Matters in Deep {RL}: A Case Study on {PPO} and {TRPO}},
  booktitle = {International Conference on Learning Representations (ICLR)},
  year = {2020},
  eprint = {2005.12729},
  archivePrefix = {arXiv},
  primaryClass = {cs.LG},
  url = {https://openreview.net/forum?id=r1etN1rtPB},
}

@article{wen2025fantastic,
  author = {Kaiyue Wen and David Hall and Tengyu Ma and Percy Liang},
  title = {Fantastic Pretraining Optimizers and Where to Find Them},
  journal = {arXiv preprint arXiv:2509.02046v2},
  year = {2025},
  eprint = {2509.02046},
  archivePrefix = {arXiv},
  primaryClass = {cs.LG},
  url = {https://arxiv.org/abs/2509.02046v2},
}

@misc{miniswe_2026,
  author = {{SWE-agent team}},
  title = {mini-swe-agent},
  year = {2026},
  howpublished = {GitHub repository},
  note = {Version 2.4.6},
  url = {https://github.com/SWE-agent/mini-swe-agent/tree/a83fcae82d2a08f0ee0c688f9d137b3566c097f8},
}

@misc{minisweplus_2026,
  author = {{Kwai-Klear}},
  title = {mini-swe-agent-plus},
  year = {2026},
  howpublished = {GitHub repository},
  note = {Revision \texttt{3dfa5e2}},
  url = {https://github.com/Kwai-Klear/mini-swe-agent-plus/tree/3dfa5e26831306978ff3cfa2da15b49113ded0e6},
}

@inproceedings{yang2025gateddeltanet,
  title={Gated Delta Networks: Improving {Mamba2} with Delta Rule},
  author={Yang, Songlin and Kautz, Jan and Hatamizadeh, Ali},
  booktitle={The Thirteenth International Conference on Learning Representations},
  year={2025}
}

@misc{xu2026polaragenticrlharness,
      title={Polar: Agentic RL on Any Harness at Scale}, 
      author={Binfeng Xu and Hao Zhang and Shaokun Zhang and Songyang Han and Mingjie Liu and Jian Hu and Shizhe Diao and Zhenghui Jin and Yunheng Zou and Michael Demoret and Jan Kautz and Yi Dong},
      year={2026},
      eprint={2605.24220},
      archivePrefix={arXiv},
      primaryClass={cs.DC},
      url={https://arxiv.org/abs/2605.24220}, 
}

@misc{he2026agentlightningv10harnessed,
      title={Agent Lightning v1.0: Towards Harnessed Agentic RL}, 
      author={Zhiyuan He and Siwei Zhang and Zhiwen Zhou and Yuqing Yang and Yu Kang and Yuge Zhang and Luna K. Qiu and Tin Yan Tsui and Jiahang Xu and Chong Luo},
      year={2026},
      eprint={2608.17528},
      archivePrefix={arXiv},
      primaryClass={cs.AI},
      url={https://arxiv.org/abs/2608.17528}, 
}

@misc{zeng2026dockerlessenvironmentfreeprogramverifier,
      title={Dockerless: Environment-Free Program Verifier for Coding Agents}, 
      author={Wenhao Zeng and Yuling Shi and Xiaodong Gu and Chao Hu and Chaofan Wang and Yuhao Cui and Hongting Zhou and Mengnan Qi and Jianqiao Wangni and Zhaojian Yu and Shuzheng Gao and Kai Cai and Shilin He},
      year={2026},
      eprint={2606.28436},
      archivePrefix={arXiv},
      primaryClass={cs.SE},
      url={https://arxiv.org/abs/2606.28436}, 
}

@misc{axiaodbl2026qwen35swebench,
  author       = {AxiaoDBL},
  title        = {Qwen3.5-4B SWE-bench Trajectory SFT},
  year         = {2026},
  howpublished = {\url{https://huggingface.co/AxiaoDBL/qwen3.5-4b-swebench-sft}},
  note         = {Hugging Face model repository}
}
\bibliographystyle{iclr2027_conference}
\clearpage

\clearpage
\appendix

\section{Common experimental setup}
\label{app:setup}
\label{appendices-start}

Unless a study states otherwise, all experiments share the following setup.

\paragraph{Models, hardware and budgets.}
We train Qwen3.5-4B, and Qwen3.5-9B in Section~\ref{sec:main-run}, with \ours{}. The 4B
runs use one eight-GPU B200 node, with two GPUs for the trainer and six for independent
vLLM inference engines.  We chose this split because it balances the time the engines take
to generate a batch against the time the trainer takes to consume it, which minimized idle
GPU time among the splits we tried. Training rollouts are sampled at temperature 1.0 without nucleus or
top-$k$ truncation (evaluation uses top-$p$ 0.95 and top-$k$ 20; Appendix~\ref{sec:app_eval_setup}), with at most 100 turns, 131{,}072 tokens per trajectory and
32{,}768 tokens per model response. A command times out after 60 seconds and a rollout
after 5{,}400 seconds.

\paragraph{Agent harness.}
\label{app:harness}
All training and evaluation runs use our Mini-SWE-Agent-Plus harness. It follows the
public mini-swe-agent design~\citep{miniswe_2026} and takes its string-replace editor
from mini-swe-agent-plus~\citep{minisweplus_2026}, exposing both tools as native tool
calls. Table~\ref{tab:harness} lists its remedies for malformed or unproductive turns;
none of them edits the task, its tests or a solution.

\begin{table}[h]
\centering
\small
\caption{Harness behavior relative to the public harnesses. \emph{Evaluation} is the
setting behind every reported SWE-bench score; training uses the same prompt and tools
but ends a rollout where evaluation gives feedback, with reward 0 (Section~\ref{sec:recipe})
or $-1$ (Section~\ref{sec:main-run}; Appendix~\ref{app:stop-policy}).}
\label{tab:harness}
\begin{tabularx}{\linewidth}{l>{\raggedright\arraybackslash}X>{\raggedright\arraybackslash}X}
\toprule
Remedy & Evaluation & Training \\
\midrule
Two native tools, one call per turn & \texttt{execute\_bash} and \texttt{edit\_via\_str\_replace} & Same \\
Format feedback & Malformed or prose-bearing turn is not executed; the model is told why and continues & Rollout ends \\
Duplicate tool call & In case of duplicated tool call without successful code edit, a short note suggests changing strategy & Optional; terminate if configured allowed number of duplicate tool calls reaches\\
Workflow hint & Prompt asks to run a relevant public test before submitting & Same \\
Submission & Optional; unit tests execute regardless & Exact \texttt{echo MINI\_SWE\_AGENT\_\allowbreak FINAL\_\allowbreak OUTPUT}; a non-exact command ends the rollout with zero score \\
Output and time limits & 10k-character head and tail, 60\,s command timeout, as in the public harness & Same \\
Budgets & 100 turns, 131{,}072 tokens, 32{,}768 per turn & Same \\
\bottomrule
\end{tabularx}
\end{table}

\paragraph{Advantages.}
For a prompt $x$ we sample a group of $G$ trajectories
$y^{(1)}, \dots, y^{(G)}$ from the behavior policy $p_{\mathrm{rollout}}$ ($G=8$ on full data and 5 on
the fixed pool) and score each with the task's own test suite to obtain rewards $R^{(i)}\in\{0,1\}$. The
advantage is the group-centered reward, shared by every token of a trajectory:
\begin{equation}
A^{(i)} \;=\; R^{(i)} \;-\; \frac{1}{G} \sum_{j=1}^{G} R^{(j)} .
\end{equation}
We deliberately do \emph{not} divide by the group standard deviation. A group whose rewards are all
equal has $A^{(i)} = 0$ for every member and is dropped by an explicit
zero-advantage filter, applied both before and after batch assembly, so it
never consumes optimizer signal or batch capacity.

\paragraph{Importance weights and the CISPO form.}
Rollouts are generated by an inference engine and scored by a training engine,
and the two assign measurably different probabilities to the same token. Sampled
tokens can also be several policy versions stale (Section~\ref{sec:setup}). Both effects are
captured by a single per-token ratio,
\begin{equation}
\omega_t=\frac{p_{\mathrm{train}}(y_t\mid h_t)}{p_{\mathrm{rollout}}(y_t\mid h_t)},
\label{eq:ratio}
\end{equation}
where $h_t$ is everything in the trajectory before token $y_t$ (the prompt, earlier turns and tool
outputs), $p_{\mathrm{rollout}}$ is the distribution that actually produced the token, and
$p_{\mathrm{train}}=\pi_\theta$ is the policy being optimized. PPO-style objectives
clip a ratio of this kind and thereby \emph{zero the gradient} of any token
outside the trust region, which on long agent trajectories discards exactly the
tokens where the policy is changing most. CISPO instead keeps every token in
the objective and applies the importance weight in the \emph{coefficient} only,
detached from the gradient~\citep{cispo2025}:
\begin{equation}
\mathcal{L}(\theta)
  \;=\; -\,\frac{1}{\sum_{i\in\mathcal{B}} |y^{(i)}|}
    \sum_{i\in\mathcal{B}} \sum_{t=1}^{|y^{(i)}|}
      \operatorname{sg}\!\big[k^{(i)}_t\big] \;
      A^{(i)} \;
      \log \pi_{\theta}\big(y^{(i)}_t \mid h^{(i)}_t\big),
\end{equation}
where $\mathcal{B}$ is the set of trajectories in the batch, $|y^{(i)}|$ counts the sampled tokens of
trajectory $i$ (tool outputs are not trained), $\operatorname{sg}[\cdot]$ is the stop-gradient
operator and $k^{(i)}_t$ is the bounded weight of Equation~\eqref{eq:mask} for token $t$ of trajectory
$i$. The normalizer is the total token count of the batch, not the per-prompt mean: averaging within a prompt first would
give a $5$-turn trajectory and a $100$-turn trajectory equal weight, which on
this workload means systematically underweighting the long trajectories that
contain most of the interesting behaviour. We apply no reference-policy KL
term ($\texttt{kl\_coef} = 0$) and no entropy bonus, and negative advantages
are not rescaled or masked.

\paragraph{Token-level discrepancy masking.}
The engine gap compounds with sequence length and with training steps, and it
is not benign: past a few percent of relative probability difference, updates
are dominated by engine disagreement rather than by the reward signal.
Following IcePop~\citep{ring1t} we bound $\omega_t$ from above at $c = 2$, and
\emph{mask} rather than clip:
\begin{equation}
k_t \;=\;
\begin{cases}
  \omega_t, & \omega_t \le c, \\
  0,   & \omega_t > c.
\end{cases}
\label{eq:mask}
\end{equation}

\paragraph{Sandboxes.}
\label{app:sandboxsummary}
Each rollout runs in its own container on a node-local k3s cluster that shares the
training node with the trainer and inference engines. Each sandbox requests 0.25 vCPU,
0.5\,GiB of memory and 2\,GiB of disk for scheduling, with limits of four vCPUs,
16\,GiB and 16\,GiB; resident sandboxes share the host's unused headroom. Up to 240
sandboxes run concurrently in the fixed-pool studies and 252, at most 42 per inference
engine, in the full-data runs. A pressure monitor constrains admission above 0.85 of
allocatable memory and 0.95 of CPU.

\paragraph{Evaluation.}
SWE-bench Verified uses a fixed 483-problem subset and SWE-bench Pro all 731 problems,
with four rollouts per problem under the same harness and budgets. We report
average@4, the mean success over the four rollouts. Checkpoints are evaluated offline;
Appendix~\ref{app:integrity} describes the excluded problems and the history and network
isolation of evaluation sandboxes.

\section{Overfit dynamics and experimental details}
\label{app:overfit}

\subsection{From an overfit deficit to a full-data learning gap}
\label{app:consistency}

The derivation bounds the selected component's remaining deficit after a
model/data change, then weights it by its share of the full distribution.

\paragraph{Attainable reward and task coverage.}
Fix the grading criteria and interaction budgets. For a given base model $M$, let $\Pi_M$ be the family of expressible policies. We define the expected reward on a dataset distribution $P$ and its theoretical optimum as:
\begin{equation}
J_P(\pi) = \mathbb{E}_{x\sim P,\,\tau\sim\pi}[R_x(\tau)], \qquad R^*_{M,P} = \sup_{\pi\in\Pi_M}J_P(\pi)
\label{eq:capability-ceiling}
\end{equation}
where $R_x(\tau)\in\{0,1\}$ is the task-success reward. We partition the full training distribution $D$ into two components: a solvable target subset $S$ (representing the overfitting tasks) with probability mass $\rho \in (0,1)$, and the remaining tasks $H$. The full dataset is therefore $D = \rho S + (1-\rho)H$.

\paragraph{Persistence under data or model shifts.}
During the diagnostic overfitting run on subset $S$, we know $R^*_{M,S} = 1$. Let $q_S(u) = J_S(\pi_u)$ be the success rate of the policy $\pi_u$ after $u$ updates. Assuming continuous acquisition of task behaviors at rate $g$ and loss (due to off-policy drift or recipe instability) at rate $d$, the dynamics are $\frac{dq_S}{du} = g(1-q_S) - d q_S$. This yields the closed-form trajectory:
\begin{equation}
q_S(u) = q_* + [q_S(0) - q_*]e^{-(g+d)u}, \qquad q_* = \frac{g}{g+d} = 1 - \varepsilon
\label{eq:recipe-reference-solution}
\end{equation}
where $\varepsilon$ is the permanent recipe deficit. 

When transitioning to full-data training, the dynamics on $S$ may be perturbed by task interference or reduced sampling frequency. We model the success rate on $S$ during full-data training, $q_D(u) = J_S(\pi^D_u)$, with a bounded perturbation term $e(u)$ where $|e(u)| \leq \eta$:
\begin{equation}
\frac{dq_D}{du} = g(1-q_D) - d q_D + e(u)
\label{eq:normalized-recipe-response}
\end{equation}
Integrating this linear equation demonstrates that the policy remains bounded by the original deficit, offset only by the maximum interference:
\begin{equation}
\limsup_{u\to\infty} q_D(u) \leq 1 - \varepsilon + \frac{\eta}{g+d}
\label{eq:recipe-transfer-transient}
\end{equation}
Crucially, a permanent performance deficit survives full-data training whenever the perturbation cannot outpace the recipe's inherent decay ($\eta < d$). Moving to a larger dataset or a larger base model guarantees a persistent loss unless the baseline stability is corrected first.

\paragraph{Converting the deficit into a full-data gap.}
Assuming the policy can simultaneously approach the optimal performance on both subsets $S$ and $H$, the full dataset optimum is $R^*_{M,D} = \rho R^*_{M,S} + (1-\rho)R^*_{M,H}$. We decompose the performance gap:
\begin{align}
R^*_{M,D} - J_D(\pi) &\geq \rho[R^*_{M,S} - J_S(\pi)] \label{eq:component-gap-decomposition} \\
R^*_{M,D} - \limsup_{u\to\infty} J_D(\pi^D_u) &\geq \rho R^*_{M,S}\left(\varepsilon - \frac{\eta}{g+d}\right) \label{eq:robust-capability-gap}
\end{align}
The first inequality holds because $J_H(\pi) \leq R^*_{M,H}$. The second substitutes our dynamical bound from Equation~\ref{eq:recipe-transfer-transient}. If the component optima inherently conflict (meaning the model lacks the capacity to master $S$ and $H$ simultaneously), the gap strictly widens. Setting $\eta=0$ recovers the idealized bound in Section~\ref{sec:recipe}.

\subsection{Fixed-pool runs}
\label{app:history}
\label{app:contract}
The epoch axes of Figures~\ref{fig:recipe_before}
and~\ref{fig:recipe_after} count sampled task groups divided by 160, so an epoch includes groups
later filtered for uniform reward or dropped as stale.

\paragraph{Public SkyRL recipe.}
\label{app:public_overfit}
The public SkyRL recipe ports SkyRL's Mini-SWE example at revision \texttt{53b1155} \citep{skyrl_mini_2026}
to the same model, pool, budgets and platform. It keeps standardized GRPO advantages,
PPO clipping at 0.2, a reference-KL coefficient of 0.001 and AdamW with $\beta_2=0.999$,
weight decay 0.01 and learning rate $10^{-6}$, with synchronous updates and KV
recomputation; recovery, prose and duplicate actions are allowed. At update 29 its reward
falls to 36.2\% as the submission rate drops from 96\% to 61\% and the reference KL rises
from 0.010 to 0.039, while mismatch KL stays near $10^{-4}$; reward recovers within two
updates. No update of the full recipe falls more than 16 points below the previous one.

\paragraph{Training diagnostics.}
\label{app:mismatch-kl}
For a sampled token $y_t$ at interaction history $h_t$, let
$p_{\mathrm{train}}(y_t\mid h_t)$ be its probability under the trainer and
$p_{\mathrm{rollout}}(y_t\mid h_t)$ its probability recorded at rollout time, under the
engine state and policy version that generated it. Let $\mathcal{T}$ be
the sampled token positions included by the training loss mask in one update.
The logged rollout--trainer mismatch KL is
\begin{equation}
\widehat K_{\mathrm{mis}}=
\frac{1}{|\mathcal{T}|}\sum_{t\in\mathcal{T}}
\bigl(\omega_t-\log\omega_t-1\bigr),
\label{eq:mismatch-kl}
\end{equation}
where $\omega_t$ is defined in \eqref{eq:ratio}. Natural logarithms give units of nats. The mean pools tokens across trainer
microbatches and ranks using unclipped ratios, including tokens excluded by
importance-weight masking.
For a fixed history and tokens drawn from $p_{\mathrm{rollout}}$ with common
support, $\mathbb{E}[\omega_t-1]=0$, so the expected summand is
$D_{\mathrm{KL}}(p_{\mathrm{rollout}}\Vert p_{\mathrm{train}})$.
The metric includes both policy drift and execution mismatch, and is separate
from reference-policy KL regularization. Figure~\ref{fig:recipe_after} plots per-update values,
and Figure~\ref{fig:recipe_before} smooths the earlier-stack runs with
centered 21-update means;
the last update of the earlier-stack gated-reward and loop/format-stop runs and of the public SkyRL recipe
logged no trainer metrics, so their entropy and KL traces end one update early.



\subsection{Format handling in public training stacks}
\label{app:parsers}
Qwen3.5 closes the reasoning block opened by its generation prompt with \texttt{</think>} and
then calls a tool in the XML format its chat template specifies: \texttt{<tool\_call>} wraps
one \texttt{<function=NAME>} block, which holds one \texttt{<parameter=KEY>} block per argument.
We write 13 malformed turns, each breaking the template's format in one element of a valid
\texttt{execute\_bash(ls)} turn, and run them through the path each stack uses to turn a sampled turn into an executed
action (Table~\ref{tab:parsers}). For the four RL frameworks this is the train-time rollout path
of the SWE example we cite~\citep{primerl_qwen_swe_2026,skyrl_mini_2026,slime_swe_2026,miles_swe_2026}, at the main-branch commit listed in Table~\ref{tab:parsers}; the cited revisions give the same
results, except that Prime-RL's older renderer repairs a missing \texttt{</think>}. Prime-RL
parses the sampled token IDs with its Qwen3.5 renderer and runs its bash harness. slime and miles
parse with SGLang's \texttt{qwen3\_coder} and \texttt{qwen3} parsers and run Claude Code and
mini-swe-agent, respectively. SkyRL's Mini-SWE example serves vLLM without a tool parser and
extracts one fenced bash block from the text, so its column tests the bash-block analogue of each
flaw. The vLLM and SGLang columns show these servers' parsers alone, as chat-completion clients
use them. The two \ours{} columns run our harness with the gated reward of
Section~\ref{sec:env}, without and with the raw token check.

Every public train-time path still executes a call for most of these 13 turns, and
none penalizes a flawed turn: reward comes only from the final tests. Some repairs change the
action itself: with a duplicated parameter, every public XML parser keeps the last value and runs
\texttt{rm -rf build} instead of \texttt{ls}. When no call is parsed, Prime-RL and slime end the
episode and grade the repository as it stands, while SkyRL and miles return a format message and
continue. Prime-RL and miles train the sampled tokens of every executed turn, and so does slime
with Claude Code 2.1.233 and later, so a turn the parser repaired is reinforced whenever the
rollout succeeds. SkyRL instead re-renders each turn through the chat template, which rewrites
the reasoning flaws, and it also trains the turns it rejected. Our gate alone still executes 6 of
them because it inspects the parsed call; the raw token check rejects all 13 and sets the
rollout's reward to zero. Our harness additionally rejects empty reasoning, prose between the
reasoning and the call, and more than one call per turn. The template permits all three, so we
treat them as preferences of our harness rather than format errors and leave them out.

\begin{table}[htbp]
\centering
\footnotesize
\setlength{\tabcolsep}{3.5pt}
\newcommand{\lp}{\cellcolor{red!12}P}
\newcommand{\lr}{\cellcolor{red!12}R}
\newcommand{\ld}{\cellcolor{red!12}D}
\newcommand{\lx}{$\times$}
\caption{\textbf{Public training stacks let most format flaws through.} Outcome of 13 malformed
Qwen3.5 turns: P, executed as written; R, silently repaired and executed; D, executed with part
of the action dropped or replaced; $\times$, no call executed. Shaded cells execute a call. RL
frameworks run the train-time rollout path of their SWE example at the commit shown; the SkyRL
column uses bash-block analogues, and -- marks flaws without one. \ours{} runs the gated reward
without and with the raw token check.}
\label{tab:parsers}
\begin{tabularx}{\linewidth}{>{\raggedright\arraybackslash}Xcccccccc}
\toprule
& \multicolumn{2}{c}{Server parser} & \multicolumn{4}{c}{RL framework, train time}
& \multicolumn{2}{c}{\ours{}} \\
\cmidrule(lr){2-3}\cmidrule(lr){4-7}\cmidrule(lr){8-9}
Flaw & vLLM & SGLang & Prime-RL & SkyRL & slime & miles & Gated & + raw \\
\multicolumn{1}{r}{\scriptsize commit} & \scriptsize\texttt{9bf44c4} & \scriptsize\texttt{1f6ce4b}
& \scriptsize\texttt{c28afbb} & \scriptsize\texttt{b04e9c5} & \scriptsize\texttt{8ee9c1e}
& \scriptsize\texttt{3d01116} & & \\
\midrule
\multicolumn{9}{l}{\emph{Reasoning block}} \\
No \texttt{</think>} & \lr & \lr & \lx & \lr & \lr & \lr & \lx & \lx \\
Second \texttt{</think>} & \lp & \lp & \lp & \lr & \lp & \lp & \lx & \lx \\
Extra \texttt{<think>} & \lr & \lr & \lr & \lr & \lr & \lr & \lr & \lx \\
Call inside reasoning & \lr & \lx & \lx & \lp & \lx & \lx & \lx & \lx \\
\multicolumn{9}{l}{\emph{Tool-call envelope}} \\
Text after \texttt{</tool\_call>} & \lr & \lr & \lr & \lp & \lr & \lr & \lr & \lx \\
Unclosed \texttt{<tool\_call>} & \lr & \lr & \lx & \lx & \lr & \lr & \lx & \lx \\
Nested \texttt{<tool\_call>} & \lr & \lr & \lr & \ld & \lr & \lr & \lr & \lx \\
\multicolumn{9}{l}{\emph{Call syntax}} \\
No \texttt{</parameter>} & \ld & \lr & \ld & -- & \lr & \lr & \lx & \lx \\
Duplicate parameter & \ld & \ld & \ld & -- & \ld & \ld & \ld & \lx \\
No \texttt{</function>} & \lr & \lr & \lr & -- & \lr & \lr & \lr & \lx \\
No \texttt{>} after \texttt{<function=} & \ld & \ld & \lx & \lx & \lx & \lx & \lx & \lx \\
Two functions in one call & \lr & \lr & \ld & \lp & \lr & \lr & \ld & \lx \\
No \texttt{<function=} & \lx & \lx & \lx & \lx & \lx & \lx & \lx & \lx \\
\midrule
Call still executes & 12/13 & 11/13 & 8/13 & 7/10 & 10/13 & 10/13 & 6/13 & 0/13 \\
\bottomrule
\end{tabularx}
\end{table}

\section{Additional numerical and serving measurements}
\label{app:systems}

\subsection{Numerical consistency and the fixed-weight audit}
\label{app:numerics}

\paragraph{Decoding rounds the recurrent state at every token.}
Qwen3.5 interleaves softmax attention with GatedDeltaNet layers~\citep{yang2025gateddeltanet}.
Each GatedDeltaNet head keeps a matrix-valued state $S_t$ that is decayed and rewritten at
every token,
\begin{equation}
S_t = S_{t-1}\,\alpha_t\bigl(I-\beta_t k_t k_t^{\top}\bigr) + \beta_t v_t k_t^{\top},
\label{eq:gdn_state}
\end{equation}
where $k_t$ and $v_t$ are the key and value, $\beta_t\in(0,1)$ is the write strength and
$\alpha_t\in(0,1)$ is the decay gate. The trainer sees complete sequences and evaluates this
recurrence chunkwise in parallel with the state accumulated in FP32, so rounding enters only
at chunk boundaries. In the default configuration, the inference engine decodes one token at
a time and stores the state in a BF16 cache after every update. This per-step rounding
accumulates over the trajectory and shifts token log-probabilities, and hence importance
ratios, even when both engines hold identical parameters. We keep the recurrent cache state
in FP32, cast gate inputs to FP32 before their nonlinearities and keep the gate outputs in
FP32 through the recurrent update. Projections stay in BF16 to preserve throughput.

\paragraph{The decay parameter stays in FP32 end to end.}
The decay gate is computed from a learned per-head scalar $A_{\log}$,
\begin{equation}
g_t = -\exp(A_{\log})\,\operatorname{softplus}(a_t + b_{\Delta t}),
\qquad \alpha_t = \exp(g_t),
\label{eq:systems_decay}
\end{equation}
where $a_t$ is an input-dependent projection and $b_{\Delta t}$ is a learned bias. A larger
$A_{\log}$ makes the head forget its state faster. Because $\alpha_t$ multiplies the state at
every step, BF16 quantization in the decay chain compounds across positions. BF16 also
discretizes the optimization of $A_{\log}$ itself: near a rounding boundary, small gradient
updates leave the quantized value unchanged until the accumulated momentum crosses the
boundary. We therefore place $A_{\log}$ in a separate FP32 parameter group in the trainer and
preserve its precision through weight serialization and transfer to the inference engines;
all other projection matrices remain in BF16.

\paragraph{Other rounding sites follow the inference engine.}
For residual addition and RMSNorm, we match vLLM's execution order by normalizing the FP32
sum while storing the residual in BF16. For query and key RoPE, BF16 rounding occurs after
normalization, and the rotation itself runs in FP32 before casting back. Both engines keep
LM-head outputs in FP32, so logits are not rounded before log-probabilities are computed. In
the backward pass, selective upcasting limits memory overhead without changing forward
numerics.

\paragraph{Alignment lowers training-time mismatch without slowing rollout.}
Figure~\ref{fig:infra_align}a,b compares the default and aligned configurations over training
steps 10--50, excluding warm-up transients. Mean rollout--trainer mismatch KL
(Equation~\ref{eq:mismatch-kl}) falls from about $3\times10^{-4}$ to $1.2\times10^{-4}$ and
stays lower throughout. Throughput is unchanged, because the upcast paths (gate
nonlinearities, the decay parameter and residual normalization) cost little next to the
projection and attention kernels.

\paragraph{Use a fixed-weight audit to measure execution mismatch.}
The sampler and trainer use different kernels and reduction orders. Our numerical work
includes an fp32 sum in fused add-RMSNorm, fp32 rotary and gate computations, and float32
recurrent cache state. We score the same recorded token sequences with identical weights
on both engines, removing policy drift from this audit. Across 156 trajectories and 2.3M
generated tokens, the resulting equivalent mismatch KL is $9.40\times10^{-5}$ and median
per-token disagreement is $1.37\times10^{-5}$. These are different statistics, with a heavy
upper tail; the residual is not established to be merely float32 rounding noise.

The audited numerical changes reduce mean per-token disagreement by 3.18\% and equivalent
mismatch KL by 8.31\%. The audit predates the FP32 treatment of $A_{\log}$ and scores
prefill only, so it misses the per-token decoding error described above; its 8.31\% is not
comparable with the training-time reduction in Figure~\ref{fig:infra_align}a. It
establishes improved agreement on the audited traces, not a causal improvement in learning
stability. Fixed-weight and training-time divergence also
use different policy and trajectory populations, so their ratio is not an additive
decomposition of execution error and staleness. In particular, the fraction attributed to
execution by such a ratio varies substantially with the selected training window.

\subsection{Serving admission changes realized staleness}

\paragraph{The configured bound does not determine the observed policy age.}
At the same configured off-policy cap of four, a change from hash-based assignment to
load-aware admission with per-client caps coincides with lower logged depth, lower
trainer/sampler divergence and shorter steps. Entropy is similar, at 0.229 versus 0.240. This is evidence
that serving controls change the conditions under which updates are made, even when the
nominal algorithm configuration is unchanged.

\paragraph{Per-rank placement removes a straggler engine.}
With one endpoint of six data-parallel ranks, the router's session hash assigns requests
without seeing rollouts that are idle in a sandbox between turns, and equal arrival rates
diverge once a replica thrashes its prefix cache. In a cap-four run, one replica reached
186 concurrent requests against 8 on the others, its prefix-cache hit rate fell from 96\% to
57\%, and buffered rollouts climbed from 143 to 660. The dispatcher instead pins each group
of eight rollouts to the rank that owns the fewest in-flight rollouts, and vLLM caps each
engine at 56 resident sequences. Resuming that run from its update-200 checkpoint with this
placement, over updates 201--240:
\begin{itemize}
\item rollouts dropped at the off-policy cap fall from 27.0 to 0.76 per update;
\item mean and 90th-percentile policy lag fall from 3.1 and 4.7 to 2.0 and 3.1 updates;
\item generation throughput rises from 298k to 417k tokens/s, and waiting for a batch falls
from 18.8 to 8.2 minutes per update, while forward--backward time stays at about 3 minutes;
\item training-time mismatch KL falls from $2.45\times10^{-4}$ to $1.58\times10^{-4}$.
\end{itemize}
Dressage separately reports load-balancing gains under synchronous rollout workloads
\citep{dressage_balance_2026}; our measurement emphasizes policy age as well as elapsed time.

\paragraph{Report time waiting for useful work.}
Trainer throughput remains between 28.1k and 30.8k tokens per second across the serving
comparison because the main difference is time waiting for rollouts. Separate trainer
performance work improves compute throughput, but one instrumented run still spends
175.6 seconds in forward/backward computation and 278.4 seconds waiting for a batch out
of a 454.0-second loop. Step time, executed work and realized sample age are therefore
needed alongside kernel throughput to explain the training budget.

\subsection{Stale-cache error analysis}
\label{app:stale_cache}

\paragraph{Setup.}
We measure how much reusing caches from older weights perturbs decoding, in the spirit of
the stale-cache analysis of \citet{pipelinerl}. Let $C(\theta, x)$ denote the hybrid cache
obtained by prefilling a prefix $x$ under weights $\theta$; it comprises the key--value
cache of the softmax-attention layers and the convolution history and recurrent state of
the GatedDeltaNet layers. We take checkpoint pairs $(\theta_v, \theta_{v+k})$ with $k=10$
and $v\in\{170,250,350\}$ from a Qwen3.5-4B training run on the 160-task pool with KV
recomputation enabled. For each $v$, we select 32 training trajectories longer than 40k
tokens from update $v+1$ (59 turns on average) and truncate each at the turn boundary
nearest 32k tokens.

Each prefix is prefilled under both checkpoints. We then score the continuation under
$\theta_{v+k}$ by teacher forcing, starting from one of four caches: the recomputed cache
$C(\theta_{v+k}, x)$; caches in which only the attention layers, or only the GatedDeltaNet
layers, take their state from $C(\theta_v, x)$; and the fully stale cache $C(\theta_v, x)$.
A fifth condition scores the continuation under $\theta_v$ from $C(\theta_v, x)$ and
serves as the policy-change reference. For each condition $c$, we compute
\[
  D(c) \;=\; \frac{1}{|\mathcal{T}|}\sum_{t\in\mathcal{T}}
  \mathrm{KL}\big(p_c(\cdot \mid t)\,\big\|\,p_{\mathrm{rec}}(\cdot \mid t)\big)
\]
over the full vocabulary, where $\mathcal{T}$ is the first 512 trainable positions of the
continuation and $p_{\mathrm{rec}}$ is the next-token distribution under the recomputed
cache. We average $D(c)$ over trajectories and report 95\% bootstrap intervals from
resampling trajectories. As a numerical floor, we use the KL between prefilling the prefix in a single
pass and in two chunks.

\paragraph{Results.}
Table~\ref{tab:stale_cache} reports $D(c)$ for each condition. Stale attention caches and
stale GatedDeltaNet states perturb the next-token distribution by comparable amounts, and
their effects are approximately additive: staling both roughly doubles the divergence of
staling either alone. The fully stale cache yields $D=7.2\times10^{-4}$, 5.7$\times$ the
numerical floor. The
mismatch KL on the recorded tokens follows the same pattern.

\begin{table}[t]
\centering
\small
\caption{Stale-cache error at a 32k-token prefix with caches 10 updates stale, averaged over
three checkpoint pairs (95\% bootstrap intervals over trajectories). Mismatch KL applies
Equation~\ref{eq:mismatch-kl} to the recorded tokens.}
\label{tab:stale_cache}
\begin{tabular}{@{}lrr@{}}
\toprule
Cache & $D$ ($10^{-4}$) & Mismatch KL ($10^{-4}$) \\
\midrule
Stale attention KV only        & 3.78 {\scriptsize[3.43, 4.12]} & 3.94 \\
Stale GatedDeltaNet state only & 3.46 {\scriptsize[3.15, 3.81]} & 3.47 \\
Both stale                     & 7.21 {\scriptsize[6.66, 7.76]} & 7.24 \\
\bottomrule
\end{tabular}
\end{table}

\section{Foundation recipe on unchanged R2E-Gym}
\label{app:r2e-recipe}

Table~\ref{tab:r2e-recipe} gives the settings of the Section~\ref{sec:r2e-baseline} run beyond
the common setup of Appendix~\ref{app:setup}, and Table~\ref{tab:r2e-ablations} the full-data
ablations behind its main choices.

\begin{table}[htbp]
\centering
\small
\caption{Recipe of the unchanged R2E-Gym run, beyond Appendix~\ref{app:setup}.}
\label{tab:r2e-recipe}
\begin{tabularx}{\linewidth}{l>{\raggedright\arraybackslash}X}
\toprule
Setting & Value \\
\midrule
\multicolumn{2}{l}{\emph{Data and sampling}} \\
Task pool & 3{,}606 R2E-Gym tasks with their original issues and tests \\
Sampling & Without replacement from a shuffled deck; retired tasks are skipped \\
Groups and batch & 8 rollouts per task; each update trains 128 rollouts from at least 16 groups with mixed rewards \\
Uniform groups & Groups whose rewards are all 0 or all 1 stay out of the batch but still count toward eviction \\
\multicolumn{2}{l}{\emph{Task eviction}} \\
Easy & Retire a task the first time a group solves it 8/8 \\
Hard & Retire a task after three consecutive appearances at 0/8 \\
Retired tasks & Not sampled again unless the live deck empties \\
\multicolumn{2}{l}{\emph{Asynchrony}} \\
Off-policy cap & 8 updates; an older group is dropped whole \\
Weight updates & Broadcast to all six engines after every update, followed by KV recomputation \\
Placement & 252 rollouts in flight, at most 42 per engine; each group stays on one engine, chosen by load \\
\multicolumn{2}{l}{\emph{Objective}} \\
Loss & CISPO with detached weights; tokens whose ratio exceeds 2 are masked, with no lower bound (Equation~\ref{eq:mask}) \\
Aggregation & Mean over all trained tokens in the batch \\
Advantage & Group-centered reward, without standard-deviation normalization \\
Other terms & No reference KL, entropy term or length penalty \\
\multicolumn{2}{l}{\emph{Reward}} \\
Base reward & 1 if the task's tests reproduce its recorded pass and fail outcomes exactly, else 0 \\
Terminal stops & Format violations (including the raw token check), an inexact submission, a second repeat of the same tool call since the last successful edit, and every turn, token, context and time budget end the rollout with reward 0; these rollouts stay in the batch \\
Scoring integrity & Gold-fix history is removed at setup; the test runner is withheld during the rollout and restored before grading \\
\multicolumn{2}{l}{\emph{Optimizer and numerics}} \\
Optimizer & AdamW, constant learning rate $10^{-6}$, $\beta=(0.9, 0.98)$, $\epsilon=10^{-15}$, no weight decay, gradient clipping at norm 1.0; one step per batch \\
FP32 paths & $A_{\log}$ in training, weight transfer and export; gate arithmetic, residual RMSNorm, rotary embedding and the LM head; the recurrent state in inference \\
\multicolumn{2}{l}{\emph{Schedule}} \\
Length & 800 updates; checkpoints and offline SWE-bench Verified evaluation every 20 updates \\
Reported checkpoint & Update 340, after 43{,}520 trained rollouts \\
\bottomrule
\end{tabularx}
\end{table}

\begin{figure}[htbp]
\centering
\includegraphics[width=\linewidth]{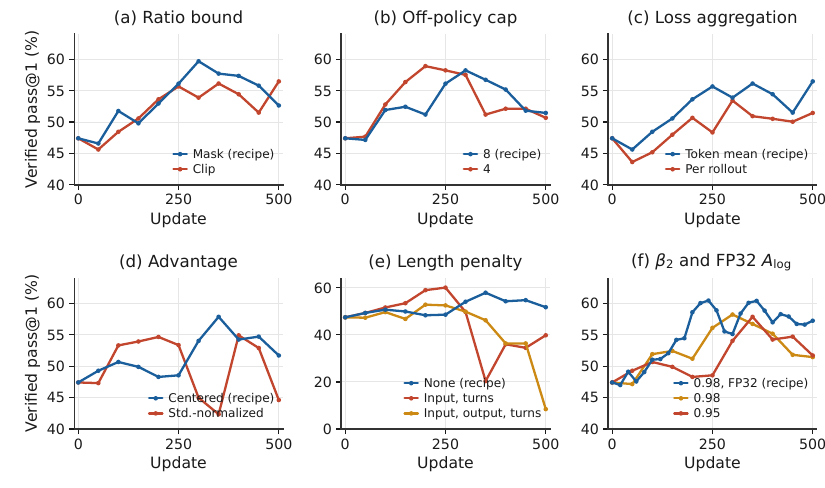}
\caption{\textbf{Full-data ablations behind the recipe.} SWE-bench Verified pass@1 against
update; the blue line is the recipe's choice. Panels (a) and (c) vary the earlier configuration,
whose clip and token-mean curves average two of its runs; (b) varies the recipe with BF16
$A_{\log}$, (d) and (e) that recipe with $\beta_2=0.95$, and (f) steps from $\beta_2=0.95$ to the
final recipe.}
\label{fig:r2e-ablations}
\end{figure}

\paragraph{Ablations.}
Each ablation is one full R2E-Gym run that changes a single setting from its base, evaluated on
SWE-bench Verified every 50 updates. Table~\ref{tab:r2e-ablations} reports mean pass@1 over the
checkpoints both runs reached, and Figure~\ref{fig:r2e-ablations} the curves. The bases are an earlier configuration (E), which clips instead of
masking, caps staleness at 4, uses $\beta_2=0.95$, normalizes advantages by the group standard
deviation and adds a length penalty; the recipe with BF16 $A_{\log}$ (R); and R with
$\beta_2=0.95$ (R$'$). For E we average two of its runs that each change one near-neutral
setting, weight decay 0.01 or an off-policy cap of 8, over the checkpoints each reached. Differences below about 2 points are within the spread between runs of one
recipe.

\begin{table}[htbp]
\centering
\small
\caption{Full-data ablations: mean SWE-bench Verified pass@1 (\%) over matched checkpoints.}
\label{tab:r2e-ablations}
\begin{tabularx}{\linewidth}{>{\raggedright\arraybackslash}X>{\raggedright\arraybackslash}Xcccc}
\toprule
Recipe choice & Alternative & Base & Checkpoints & Recipe & Alternative \\
\midrule
Mask ratios above 2 & Clip at 2 & E & 10 & 54.0 & 52.6 \\
Off-policy cap 8 & Cap 4 & R & 10 & 53.2 & 53.8 \\
Token mean over the batch & Mean per rollout within each prompt & E & 10 & 52.6 & 49.2 \\
No std.\ normalization & Std.\ normalization & R$'$ & 10 & 51.9 & 50.2 \\
No length penalty & Input and turn penalty & R$'$ & 10 & 51.9 & 45.4 \\
No length penalty & Input, output and turn penalty & R$'$ & 10 & 51.9 & 42.6 \\
$\beta_2=0.98$ & $\beta_2=0.95$ & R$'$ & 10 & 53.2 & 51.9 \\
FP32 $A_{\log}$ & BF16 $A_{\log}$ & R & 5 & 55.8 & 53.6 \\
Easy and hard eviction & Easy eviction only & R & 10 & 53.2 & 53.9 \\
\bottomrule
\end{tabularx}
\end{table}

\paragraph{Importance-ratio handling.}
Tokens whose trainer-to-rollout ratio exceeds 2 are rare, about $10^{-5}$ of all tokens, but
individual ratios exceed 500. Masking removes their gradient instead of keeping it at weight 2,
and matches or exceeds clipping.

\paragraph{Off-policy cap.}
Raising the cap from 4 to 8 updates leaves mean accuracy within noise. In runs of the earlier
configuration, the larger cap drops 0.1\% of generated rollouts as stale instead of 1.8\%, while
mean policy age rises only from 2.3 to 2.4 updates and mismatch KL does not rise. With the system fixes, the recipe tolerates the extra
staleness, so the larger cap keeps more of the sampled work.

\paragraph{Loss aggregation and shaping.}
Averaging per rollout within each prompt gives a 5-turn and a 100-turn trajectory equal weight
and costs 3.4 points. Normalizing advantages by the group standard deviation is 1.7 points lower
on average and unstable late, at 42.3\% at update 350. Length penalties raise the early peak to
60.0\% at update 250 but then collapse, to 20.3\% at update 350, and to 8.4\% at update 500 when
output tokens are also penalized.

\paragraph{Optimizer and numerics.}
$\beta_2=0.98$ is 1.3 points above 0.95, and FP32 $A_{\log}$ adds 2.2 points over five matched
checkpoints. Both changes are near the noise level and cost nothing, so the recipe keeps them.

\paragraph{Task eviction.}
Rewards on R2E-Gym are close to deterministic per task. Over the first 127 updates of this
recipe, 51.6\% of rollouts fall in groups whose eight rewards are all equal (35.0\% all failed,
16.6\% all solved), 36 times the share expected if rollouts succeeded independently at the
observed rate. Such groups carry no gradient. Eviction retires tasks once they are mastered or
repeatedly failed; with it, the share of these rollouts falls from 49.4\% over updates 1--100 to
19.6\% over updates 401--500 in the recipe with BF16 $A_{\log}$. To check that retirement does not
remove learnable tasks, we replay the retired tasks of a full-data run with the same eviction
rule at its update-250 checkpoint. The 988 easy-retired tasks, rerun 8 times each, are still
solved in 90.0\% of rollouts, 55.3\% of them 8/8, and accuracy does not fall with time since
retirement. The 464 hard-retired tasks, rerun 32 times each, are solved in 0.9\% of rollouts,
and 85.3\% are never solved. Removing hard eviction leaves accuracy within noise
(Table~\ref{tab:r2e-ablations}), so the rule saves rollouts without costing accuracy.

\section{Curation pipeline: composition, verdicts and constraints}
\label{app:curation-funnel}


\subsection{The counterfactual probe}
\label{app:probe-protocol}

The fail-to-pass gate evaluates a suite on exactly two patches: the empty one
and gold. Passing it certifies that the suite separates those two trees and
nothing more. A suite can do that and still accept most of the patch space
between them, and reading the issue, the tests or the gold patch
can not reveal it naturally, because the question is about patches that do not yet exist.
Pipeline~2's second question is answered by creating them.

Having read the issue, the gold patch and pipeline~1's record of which
assertions are binding, the judge writes two kinds of patch and scores each
with the \emph{same scorer that computes reward during RL}---the same node-ID
contract, the same container, the same pass/fail aggregation---so that a probe
result is a statement about the training signal rather than about a
reimplementation of it.

\begin{description}
  \item[Plausible-but-wrong patches] are written from the issue alone, of the
    kind a competent solver might produce without access to gold: a shortcut
    that satisfies the letter of the reported symptom, a partial fix covering
    the reported case but not the general one, or an alias that makes a name
    resolve without performing the change the issue asks for. If such a patch
    scores $1.0$, the suite accepts a patch that does not fix the issue, and
    the verdict is \texttt{too\_loose}.
  \item[Correct-but-different patches] resolve the issue by an implementation
    other than gold's. If such a patch scores $0.0$, the suite is enforcing an
    implementation contract the issue never states, and the verdict is
    \texttt{over\_restrictive}.
\end{description}

The two probes are what make the check two-sided. A procedure that only sought
weak suites would have a standing incentive to call every suite weak, and one
that only sought over-constrained suites would reproduce the orientation of
prior filters (Appendix~\ref{app:prior-filters}). Running both means a suite is
rejected for admitting wrong patches and for rejecting right ones, and the
verdict distribution in Table~\ref{tab:verdicts} shows both classes at
comparable scale (30.0\% and 17.6\%).

Only the verdict is a model judgement. The reward figures the judge cites are
produced by executing its patches, so a mistaken verdict requires the judge to
mis-describe an executed result rather than merely to reason poorly about code
it has read. Two residual limits are worth stating. First, a probe establishes
that \emph{a} wrong patch passes, not that most wrong patches do; the rate we
report is therefore a lower bound on how permissive a suite is, since a judge
that fails to find a passing shortcut does not prove none exists. Second, the
probe records in Appendix~\ref{app:curation-examples} are the judge's own
accounts of what it ran, not an independent re-execution. The downstream check
on whether the \texttt{too\_loose} label tracks something real is the
controlled training comparison in Section~\ref{sec:hardening}, where suites carrying
that label are strengthened and the same instance IDs are trained both ways.

\subsection{What pipeline 1 excludes}

Pipeline~1 uses no model. It runs the graded suite on the buggy and gold trees
to derive (or verify) the fail-to-pass gate, then instruments every assertion
in the gating tests to identify which ones are \emph{binding}, i.e.\ fail on
the buggy tree and pass on gold. Its output is a deterministic record of what
each suite does, and it excludes 3{,}238 rows in three separately recorded
classes (Table~\ref{tab:funnel}).

First, 310 Scale-SWE rows ship without a container image and can never be
executed. Second, 583 rows the harness could not measure reliably---sandbox and
exec failures (193), suites that did not run or failed collection (321), and
gold trees that fail their own suite (69). We record these as failures to
measure rather than quality verdicts, because on a deck whose gold is known
good a gold mismatch is more often evidence that we are misreading the suite
than that the task is broken. Third, 2{,}345 rows that fail validation
outright: in 1{,}652 the reference patch does not make the declared gate pass,
391 have no usable gate at all (already passing on the buggy tree, declared
tests not collected, or none derivable), 176 are non-pytest suites where the
assertion probe recorded nothing, and 126 have a gold patch that does not
apply. The dominant failure is a reference solution that does not satisfy the
tests defining its own task---a defect no downstream judge could repair.

\subsection{Resulting composition}

The two acceptance rates in Table~\ref{tab:funnel} are the clearest summary of
what the second pipeline adds. The standard fail-to-pass gate admits 84.7\% of
source instances; of those, only 42.0\% survive the judge, for an end-to-end
rate of 35.5\%. Nearly all of the filtering, and all of the filtering that
concerns whether reward is valid, happens in pipeline~2.

Table~\ref{tab:verdicts} breaks those rejections down by verdict, and the
profile is corpus-dependent and diagnostic. Scale-SWE, whose gate is a single
human-written PR test---a weak discriminator by construction---is rejected
predominantly as \texttt{too\_loose} (3{,}210 rows, 35.5\% of the corpus),
whereas SWE-rebench's multi-test gates more often over-constrain, failing as
\texttt{over\_restrictive} (1{,}227) slightly more often than as
\texttt{too\_loose} (1{,}155). Only 2 of 1{,}235 attempted rewrites were
rejected for eliding the original, so the rewrite pathway is almost never the
binding constraint; the rejections are overwhelmingly about test quality.

\subsection{Constraints on issue rewriting}

When a rewrite is warranted the judge may edit the \texttt{problem\_statement}
and nothing else. It adds the missing behavioural requirement while preserving
the original's voice and reproduction snippet, states \emph{what} must happen
rather than \emph{how}, and never names private helpers or file locations
introduced by the gold patch. A rewrite that elides content it claims to
reproduce is rejected and the row dropped.

\subsection{Why rewriting is not always the right repair}

Rewriting and rejection begin from the same observation---the tests grade
something the issue never mentions---and differ only in which document is at
fault. When the unstated requirement is legitimate user-visible behaviour, the
issue is deficient and we add the requirement
(Appendix~\ref{app:ex-rewrite}). When the test instead demands gold's
\emph{implementation}, writing that contract into the issue would turn ``fix
this bug'' into ``reproduce this implementation'' and reward a test that is
itself defective, so the row is rejected as \texttt{over\_restrictive}
(17.6\%, Table~\ref{tab:verdicts}; Appendix~\ref{app:ex-over-restrictive}).
This is why the judge's questions are asked in a fixed order: asked in the
opposite order, it would rewrite every over-restrictive test into a
well-specified task.

The verdict records support this reading of the two defects as one mismatch.
Because the checks stop at the first failure, the
\texttt{issue\_underspecified} flag is recorded even on rows rejected earlier
for another reason, so we can ask how often it co-occurs with each rejection
class. It is raised on 2{,}027 of the 17{,}861 judged rows (11.3\%), of which
1{,}235 reach the question as an actionable decision (1{,}233 rewritten and 2
rejected for eliding the original) and 792 are rejected before it. Among the
rejection classes, the flag is far more common on \texttt{over\_restrictive}
rows (14.9\%) than on \texttt{too\_loose} (3.8\%) or \texttt{off\_target}
(4.2\%) ones---three to four times the rate---which is what the account above
predicts: when a suite silently demands gold's implementation, the issue
usually fails to mention that requirement too, and the judge observes both
halves of the mismatch on the same row. The reported rewrite rate is therefore
conditional by construction: 16.5\% of the 7{,}502 rows whose gold patch and
suite are both sound (Table~\ref{tab:verdicts}), not of everything judged.

\subsection{What existing filters check}
\label{app:prior-filters}

The claim in Section~\ref{sec:filtering} that prior quality filters are
one-directional is a statement about their inputs, not about their care. Each
assesses a task by reading the issue, the test patch and the gold patch, and
that evidence can only expose a suite demanding \emph{more} than the issue
states.

\textsc{SWE-rebench} fine-tunes a classifier to predict a \emph{test patch
correctness} label, defined as whether the tests verify the intended fix
``without over-reliance on specific implementation details'', and motivates it
with the failure of ``flawed tests that a correct solution cannot pass''
\citep{rebench}. Its V2 successor replaces the classifier with an ensemble of
LLM judges whose test-alignment rubric enumerates seven signals: tests spanning
unrelated modules, string assertions on names absent from the issue, external
documentation URLs, vague acceptance criteria, patches touching unrelated files
the tests then require, tests asserting approaches the issue never mentions,
and inline tests in the non-test patch \citep{rebenchv2}. Every test-side
category is a suite over-constraining the task; none asks whether a wrong patch
would pass. Both calibrate against \textsc{SWE-bench Verified} human
annotations, whose test criterion is likewise that a correct solution not be
unfairly rejected \citep{swebench2023}, so the orientation is inherited by
construction. 

\textsc{Scale-SWE} goes furthest, and is the sharpest case. LLM judges screen
repositories and pull requests; a unit-test creator agent then synthesises the
F2P and P2P suites under an explicit anti-overfitting rule forbidding tests to
call newly introduced functions directly rather than through the public API;
and a problem-statement agent writes the issue \emph{from} those tests, so that
issue and suite are aligned by construction \citep{scaleswe}. Validation is
then the standard gate, plus a manual audit of 100 sampled instances. Every one
of these mechanisms points the same way: the anti-overfitting rule targets
tests demanding too \emph{much}, and writing the issue from the tests
guarantees the issue mentions what the suite checks but says nothing about
whether a patch ignoring the issue would also pass it. Our per-corpus verdicts
show the consequence directly (Table~\ref{tab:verdicts}). Scale-SWE has much
the lowest rewrite rate---11.8\% of its shipped rows against 26.6\% for
SWE-rebench---which is alignment by construction working as intended; and
simultaneously the highest \texttt{too\_loose} rate, 35.5\% of judged rows
against 21.8\%. Construction settles the relation it can write down, and leaves
untouched the one only execution can settle.

\textsc{R2E-Gym} is the exception that demonstrates the constraint. It measures
a \emph{distinguishability rate}---the fraction of tests that behave
differently on correct and incorrect patches---and reports that for most
problems fewer than 20\% of tests discriminate; it also measures \emph{toxic}
tests, which pass incorrect patches while failing correct ones, the analogue of
our \texttt{over\_restrictive} \citep{r2egym}. That measurement is possible
only because candidate patches are available to score, and it is performed on
test-time verifier suites for benchmark problems, with the shortfall addressed
by hybrid verifiers at inference. The suite that computes training reward is
never subjected to it. Our pipeline~2 applies the same style of evidence at
curation time, to the graded suite itself, and repairs rather than reranks.

\subsection{Composition of the unfiltered control arm}
\label{app:filter_experiment}
The control data deck for the filtering ablation is drawn from rows that clear
pipeline~1 but are rejected by pipeline~2, kept non-rewritten. The rejected
part is large enough to supply a matched deck for SWE-rebench (3{,}532
rejected rows available for 1{,}776 needed) and Scale-SWE (5{,}400 for
3{,}638), so those two environments are drawn purely from rejected rows.
R2E-Gym has the highest pipeline-2 acceptance rate of the three (59.3\%),
leaving only 1{,}429 rejected rows against the 2{,}086 required, so 657 of its
rows (31.5\%) are drawn from the accepted set instead---still non-rewritten,
but selected by the same judge that built the curated arm. Those rows are
graded identically in both arms, so the R2E-Gym contrast is diluted and the
measured gap understates the effect of selection.

\begin{table}[t]
\centering\small
\caption{Corpus composition through the two curation pipelines. Pipeline~1
applies the same fail-to-pass gate prior corpora use and involves no model;
pipeline~2 is the privileged LLM judge introduced here. Within pipeline~1,
\emph{could not be measured} covers sandbox and exec failures, suites that did
not run, and gold trees failing their own suite: these are recorded as failures
to measure, never as quality verdicts, so they are kept separate from
\emph{failed validation}, where the task itself is defective. The acceptance
rows give each pipeline in isolation and the two composed.}
\label{tab:funnel}
\begin{tabular}{lrrrr}
\toprule
& R2E-Gym & SWE-rebench & Scale-SWE & Total \\
\midrule
Source instances                      & 3{,}606 & 6{,}584 & 10{,}909 & 21{,}099 \\
\midrule
\multicolumn{5}{l}{\emph{Pipeline 1: measurement (execution only, no model)}} \\
\quad no container image              & --      & --      & $-310$   & $-310$ \\
\quad could not be measured           & $-69$   & $-317$  & $-197$   & $-583$ \\
\quad failed validation               & $-22$   & $-959$  & $-1{,}364$ & $-2{,}345$ \\
\textbf{\quad$=$ Validated pool}      & \textbf{3{,}515} & \textbf{5{,}308} & \textbf{9{,}038} & \textbf{17{,}861} \\
\midrule
\multicolumn{5}{l}{\emph{Pipeline 2: privileged LLM judge}} \\
\quad rejected by judge               & $-1{,}392$ & $-3{,}525$ & $-5{,}390$ & $-10{,}307$ \\
\quad judge errors                    & $-37$   & $-7$    & $-10$    & $-54$ \\
\textbf{\quad$=$ Curated pool}        & \textbf{2{,}086} & \textbf{1{,}776} & \textbf{3{,}638} & \textbf{7{,}500} \\
\quad\quad issue unchanged            & 1{,}756 & 1{,}303 & 3{,}208  & 6{,}267 \\
\quad\quad \textbf{issue rewritten}   & \textbf{330} & \textbf{473} & \textbf{430} & \textbf{1{,}233} \\
\midrule
Pipeline-1 acceptance rate            & 97.5\%  & 80.6\%  & 82.8\%   & 84.7\% \\
Pipeline-2 acceptance rate            & 59.3\%  & 33.5\%  & 40.3\%   & 42.0\% \\
End-to-end acceptance rate            & 57.8\%  & 27.0\%  & 33.3\%   & 35.5\% \\
\bottomrule
\end{tabular}
\end{table}

\begin{table}[t]
\centering\small
\caption{Pipeline-2 LLM-judge verdicts over the 17{,}861 instances that clear
pipeline~1. The two test-quality verdicts that motivate this work,
\texttt{too\_loose} and \texttt{over\_restrictive}, together account for 47.6\%
of all F2P-validated candidates. The rewritten share is 6.9\% of everything
judged, which is 16.4\% of the 7{,}500 tasks actually shipped.}
\label{tab:verdicts}
\begin{tabular}{llrrrrr}
\toprule
& Verdict & R2E-Gym & SWE-rebench & Scale-SWE & Total & \% \\
\midrule
\multirow{2}{*}{Ship}
 & issue unchanged              & 1{,}756 & 1{,}303 & 3{,}208 & 6{,}267 & 35.1 \\
 & issue rewritten              & 330 & 473 & 430 & 1{,}233 & 6.9 \\
\midrule
\multirow{3}{*}{\shortstack[l]{Reject:\\test quality}}
 & \texttt{too\_loose}          & 986 & 1{,}155 & 3{,}210 & 5{,}351 & 30.0 \\
 & \texttt{over\_restrictive}   & 282 & 1{,}227 & 1{,}634 & 3{,}143 & 17.6 \\
 & \texttt{off\_target}         & 28  & 642 & 215 & 885 & 5.0 \\
\midrule
\multirow{3}{*}{\shortstack[l]{Reject:\\gold patch}}
 & partially fixes issue        & 60 & 425 & 329 & 814 & 4.6 \\
 & does not fix issue           & 33 & 70  & 2   & 105 & 0.6 \\
 & unrelated to issue           & 3  & 4   & --  & 7   & 0.0 \\
\midrule
Reject: rewrite
 & elides original              & -- & 2 & -- & 2 & 0.0 \\
\midrule
Error
 & judge failed                 & 37 & 7 & 10 & 54 & 0.3 \\
\midrule
\multicolumn{2}{l}{\textbf{Total judged}} & \textbf{3{,}515} & \textbf{5{,}308} & \textbf{9{,}038} & \textbf{17{,}861} & 100 \\
\bottomrule
\end{tabular}
\end{table}
\section{Worked examples from the curation pipeline}
\label{app:curation-examples}

This appendix supports Section~\ref{sec:filtering}, which reports that our
privileged LLM judge rejects 30.0\% of fail-to-pass-validated candidates as \texttt{too\_loose}
(Finding 1) and rewrites the issue of a further 16.4\% of otherwise-sound tasks
(Finding 2). The cases below are the concrete form of those two verdicts:
Section~\ref{app:ex-too-loose} for weak suites, Section~\ref{app:ex-rewrite} for
underspecified issues, and Section~\ref{app:ex-over-restrictive} for the opposite
defect the same probing rejects.

Every instance below passes standard fail-to-pass validation: its gating tests
fail on the buggy tree and pass on gold. The issue text is quoted verbatim from
the corpus, and the probe descriptions are the judge's own account of patches it
wrote and scored with the training scorer; the LLM judge produced these
records, but the reward figures they cite are executed, not asserted. \texttt{REWARD: 1.0} denotes full
reward, i.e.\ every graded test passing.

\subsection{Suites that pass F2P validation but do not discriminate
(\texttt{too\_loose})}
\label{app:ex-too-loose}

\paragraph{\texttt{amaranth-lang\_\_amaranth-1317} (SWE-rebench).}
The complete issue:
\begin{verbatim}
EnumMeta has been renamed to EnumType in Python 3.11
We should follow suit in `amaranth.lib.enum`.
\end{verbatim}
The gate is 20 fail-to-pass tests with an empty pass-to-pass set. The only
graded lines that exercise the rename are a module-level
\texttt{from amaranth.lib.enum import ..., EnumType, ...} and a single
\texttt{assertNotIsInstance(EnumA, EnumType)}. The judge's counterfactual patch
added one line, \texttt{EnumType = EnumMeta}, leaving the class still named
\texttt{EnumMeta} and performing no rename at all. It scored \texttt{REWARD:
1.0} on all 20 tests. A second, genuinely correct rename also scored 1.0, ruling
out over-restriction. The suite verifies only that a symbol named
\texttt{EnumType} is importable, so the rename the issue asks for is never
checked --- and the alias is a shortcut a policy can reach from the issue alone.



\paragraph{\texttt{asottile\_classify-imports\_pr19} (Scale-SWE).}
The issue asks for an \texttt{is\_explicit\_relative} property returning
\texttt{True} for imports starting with \texttt{.} and \texttt{False}
otherwise, with worked examples. The gate contains three non-dotted cases. The
counterfactual patch used the predicate
\texttt{'.' in self.import\_statement.module}, which the judge separately
verified returns \texttt{True} for \texttt{import foo.bar} and
\texttt{from foo.bar import baz} --- cases the issue explicitly requires to be
\texttt{False}. It scored \texttt{REWARD: 1.0} on all three F2P and 57 P2P
tests, because no graded case contains a dotted module.

\paragraph{\texttt{pillow@ecb64fe2} (R2E-Gym).}
The issue reports that \texttt{ImageOps.invert} raises \texttt{OSError} on
mode-\texttt{"1"} images and states the expected behaviour as ``an image where
all pixels are flipped.'' The sole reward-gating test calls
\texttt{ImageOps.invert(hopper("1"))} and discards the result, with no
assertions of any kind; the measurement pass independently recorded zero
readable assertions in that test. Two separate counterfactual patches scored
\texttt{REWARD: 1.0}: one returning the image unchanged for mode \texttt{"1"},
and one converting to mode \texttt{"L"} and inverting that, which never inverts
the binary image and returns the wrong mode. Neither flips a single pixel. This
is the limiting case --- an F2P-validated task whose gate is satisfied by not
solving it.

\subsection{Issues that underspecify the graded behaviour}
\label{app:ex-rewrite}

In each pair the original is the corpus text and the rewrite is the shipped
\texttt{problem\_statement}; the judge may edit no other field. These rewrites received no independent second-model
review, so the safeguards against leaking solution detail are the constraints
in the judge's prompt and the rejection of rewrites that elide content they
claim to reproduce.

\paragraph{\texttt{audreyr\_\_cookiecutter-964} (SWE-rebench).}
Original:
\begin{verbatim}
_run_hook_from_repo_dir should not remove files if they already existed before
The current behavior makes `--overwrite-if-exists` remove all the files
(undesirable) if some hook fails.
\end{verbatim}
The issue describes only the hook-failure path, but two of the three gating
tests require the same directory preservation when generation aborts on an
undefined-variable render failure. The judge confirmed experimentally that a
hook-only fix --- exactly what the original asks for --- passes one of three
F2P tests and earns zero reward. The rewrite generalises the requirement to any
abort after the project directory has been started, while keeping the original
wording intact.

\subsection{Suites that over-constrain (\texttt{over\_restrictive})}
\label{app:ex-over-restrictive}

The same probing rejects the opposite defect, where a behaviourally correct fix
is failed for not matching gold's implementation. We include one example because
it shows the check is two-sided rather than a bias toward stricter tests.

\paragraph{\texttt{azure\_\_walinuxagent-951} (SWE-rebench).}
The issue reports that \texttt{waagent -help} omits the
\texttt{-show-configuration} flag. The sole gating test calls
\texttt{message = usage()} and asserts \texttt{"-show-configuration" in
message}, which pins gold's decision to make \texttt{usage()} \emph{return} a
string. The judge's probe added the flag to the string \texttt{usage()}
\emph{prints}, leaving it a printer as the issue describes, and verified that
\texttt{-help} output then contains the flag --- yet the suite scored
\texttt{REWARD: 0.0}. A correct patch that restores the user-visible behavior described in the issue still obtains zero reward because the test additionally requires the function usage() to return a string.

\subsection{Scope of these examples}

These are illustrative cases selected for brevity, not a random sample, and the
probe outcomes are the judge's own reports rather than an independent
re-execution. The aggregate rates quoted in Section~\ref{sec:filtering} come from the
full judgement records over all 17{,}861 measured instances, and the downstream
check on whether the \texttt{too\_loose} label tracks something real is the
controlled comparison in Section~\ref{sec:hardening}.

\subsection{Contamination screening}
\label{app:contamination}

Three distinct decks underlie the ablations of Section~\ref{sec:filtering} and
Section~\ref{sec:hardening}: the curated deck (7{,}500 rows), the size- and
mixture-matched unfiltered control (7{,}500), and the 4{,}889-row set shared by
the strengthening pair, whose raw and strengthened arms hold identical instance IDs and
differ only in how those rows are graded. Each was screened against SWE-bench
Verified before use.

We screen at two granularities. Exact instance-ID equality is reported for
completeness but is weak evidence on its own, because the three source corpora
use mutually incompatible ID conventions---\texttt{pillow@5908ccc5} (R2E-Gym),
\texttt{cs-si\_\_eodag-1536} (SWE-rebench),
\texttt{adamchainz\_flake8-no-pep420\_pr103} (Scale-SWE)---none of which can
collide with a \texttt{pallets\_\_flask-5014} string regardless of the
underlying overlap. The binding check is therefore repository granularity,
which is the level at which familiarity with a codebase could transfer to the
benchmark.

Exact-ID overlap is zero for all three decks. Of the 12 repositories
represented in SWE-bench Verified, exactly one appears anywhere in our training
data: \texttt{pallets/flask}, contributing one row to the curated deck
(PR~\#3684) and one to the unfiltered control (PR~\#3574). Both are distinct
from Verified's sole Flask instance, \texttt{pallets\_\_flask-5014}. Because
the hit is present in both arms at equal weight, it is matched across the
comparison rather than introduced by curation, and at one row in 7{,}500 it
cannot plausibly move a 483-problem evaluation. The strengthening decks contain no
repository-level hit at all, so the comparison in Section~\ref{sec:hardening} is
clean at both granularities.

\section{Test-suite strengthening}
\label{app:hardening}

\subsection{Strengthening mechanics}
\label{app:hardening-mechanics}

\paragraph{Population.} We strengthen the 5{,}351 tasks labelled \texttt{too\_loose}
in Table~\ref{tab:verdicts}: 986 from R2E-Gym, 1{,}155 from SWE-rebench and
3{,}210 from Scale-SWE. A task carries this label only after the judge has
determined that its gold patch \emph{fully} fixes the issue, which is what gives
strengthening a reference fix to check the revised tests against. A further 307
tasks have loose tests \emph{and} an inadequate gold patch; these are recorded
as gold-patch rejections and excluded from strengthening, because passing gold is
not a useful correctness check when gold itself does not resolve the issue.

\paragraph{What the agent may assert.} The strengthening agent is the same model as
the judge \citep{qwen38}, given the issue, the full gold patch, the reward
contract, shell access to the task's container, and editable copies of every
file graded by the reward gate. The gold patch is a reference for correct
behaviour but does not define what the tests should require: every added check
must follow from the issue or from established public behaviour of the
repository, and the prompt rules out assertions tied to private helpers, call
counts, implementation constants, or gold-specific wording. This constraint guards against converting a \texttt{too\_loose} suite into an
\texttt{over\_restrictive} one. The agent records a verdict on whether the suite is
\texttt{weak} at all---could a plausible partial fix or shortcut earn full
reward? If it answers \texttt{not\_weak} the session ends and the original
suite is kept unchanged (281 tasks, Table~\ref{tab:harden-funnel}). Otherwise
it strengthens existing reward-gating assertions, adds parametrized cases, or
introduces new test functions, and new node IDs are appended to the reward
contract so that the added tests actually contribute to grading.

\paragraph{Grading.} Table~\ref{tab:harden-grades} lists the source-tree/test-suite pairs. Before the agent starts, the original suite must yield $(\text{gold}=1,\text{buggy}=0)$; failure excludes the task from this procedure. The agent can re-check the same condition with its edited tests and extended contract. Submission requires this separation, which the driver independently verifies after the session. All 4{,}909 accepted suites passed the final check.

\paragraph{Isolation and collection.} The agent edits copies of the graded tests, not the source code or canonical files. During grading, the copies temporarily replace the canonical tests, which are restored afterward. After every shell command, the controller compares the repository diff, canonical test contents, and copied-file inventory with a snapshot.
Unauthorized changes are reverted and the command is marked as failed with an explanation. Added node IDs are checked against an actual pytest collection before entering the reward gate; unselectable IDs are reported and dropped rather than silently counted.

\paragraph{Later shortcut evaluation.} The controller applies the agent's proposed source edits and requires the resulting patch to score 1 under the original suite and 0 under the strengthened suite, recording the rejecting nodes. Neither score alone demonstrates improvement; the interpretation also depends on the patch genuinely violating the issue's requirements. This gate was added after the trained suites were produced and has been run on 20 tasks, with no measured downstream training effect.

\begin{table}[t]
\centering\small
\caption{Grading configurations. Rows 1--4 and 7--8 were used for the trained suites; rows 5--6 belong to the later shortcut gate. Agent checks may be repeated, and the gold--buggy separation is required at submission and final verification.}
\label{tab:harden-grades}
\begin{tabular}{rllll}
\toprule
& When & Source tree & Tests & Required \\
\midrule
1 & baseline & gold applied & original & $1$ \\
2 & baseline & buggy & original & $0$ \\
3 & agent check & gold applied & strengthened & $1$ \\
4 & agent check & buggy & strengthened & $0$ \\
5 & later shortcut gate & shortcut applied & original & $1$ \\
6 & later shortcut gate & shortcut applied & strengthened & $0$ \\
7 & driver, post-loop & gold applied & strengthened & $1$ \\
8 & driver, post-loop & buggy & strengthened & $0$ \\
\bottomrule
\end{tabular}
\end{table}

\subsection{Strengthening outcomes and test changes}
\label{app:hardening-results}

\paragraph{Outcomes.} Table~\ref{tab:harden-funnel} gives the full breakdown, including baseline failures, infrastructure errors, failed final checks, and budget exhaustion. Lower-concurrency retries recovered about one in five sampled baseline failures, suggesting execution-related losses but not establishing additional strengthening yield. Excluding 20 unusable strengthening patches leaves 4{,}889 tasks per training arm; all other settings follow Section~\ref{sec:hardening}.

\begin{table}[t]
\centering\small
\caption{Outcomes for all \texttt{too\_loose} candidates. The four main dispositions partition the input; indented rows break down reverted tasks. Accepted suites passed final gold--buggy grading, not necessarily the later shortcut gate.}
\label{tab:harden-funnel}
\begin{tabular}{lrrrrr}
\toprule
& R2E-Gym & SWE-rebench & Scale-SWE & Total & \% \\
\midrule
Input tasks & 986 & 1{,}155 & 3{,}210 & 5{,}351 & 100 \\
\midrule
Strengthened & 857 & 1{,}096 & 2{,}956 & 4{,}909 & 91.7 \\
Kept original (\texttt{not\_weak}) & 117 & 48 & 116 & 281 & 5.3 \\
Reverted to original & 6 & 10 & 138 & 154 & 2.9 \\
\quad baseline failure & 3 & 6 & 127 & 136 & 2.5 \\
\quad infrastructure failure & 2 & 3 & 8 & 13 & 0.2 \\
\quad failed final grading & 1 & 1 & 3 & 5 & 0.1 \\
Budget exhausted (200 turns) & 6 & 1 & 0 & 7 & 0.1 \\
\midrule
Strengthening acceptance rate & 86.9\% & 94.9\% & 92.1\% & 91.7\% & \\
Median model turns & 32 & 31 & 28 & & \\
\midrule
Paired training instances & 857 & 1{,}096 & 2{,}936 & 4{,}889 & \\
\bottomrule
\end{tabular}
\end{table}

\paragraph{Test changes.} Table~\ref{tab:harden-strength} reports added nodes and their behaviour on the buggy tree. These rates do not measure rejection of arbitrary incorrect fixes. No accepted task has an empty test patch, but some add no graded node. Many instead tighten existing assertions: replacing \texttt{assertTrue(len(acls), 1)} whose second argument is a failure
message rather than an expected value, with \texttt{assertEqual(acls, [('read', '/')])}, for example, changes a non-emptiness check into an exact-result check. For added tests that never became graded nodes, a nonempty patch alone does not establish an effect on reward.

\paragraph{Reward contracts.} SWE-rebench and Scale-SWE use subset checks over two node lists; tests outside both lists do not affect reward. R2E-Gym requires exact equality with a recorded outcome map, so added nodes introduce outcomes the harness must observe and reproduce. This may help explain its larger share of tasks adding no graded nodes and lower buggy-tree failure rate, but the comparison does not isolate the contract's causal effect.

\begin{table}[t]
\centering\small
\caption{Test changes across accepted suites. Buggy-tree failure rates do not measure rejection of all incorrect fixes. Tasks with no added nodes may still strengthen existing graded assertions. The three \emph{every\,/\,some\,/\,no added node} rows are computed over the tasks that added at least one graded node (639, 932 and 2{,}590 respectively), not over all accepted tasks.}
\label{tab:harden-strength}
\begin{tabular}{lrrr}
\toprule
& R2E-Gym & SWE-rebench & Scale-SWE \\
\midrule
Accepted strengthened tasks & 857 & 1{,}096 & 2{,}956 \\
Graded nodes added & 3{,}808 & 4{,}117 & 13{,}531 \\
Mean added nodes per task & 4.4 & 3.8 & 4.6 \\
Median added lines of test code & 27 & 31 & 35 \\
\midrule
Added nodes failing on buggy & 51\% & 84\% & 84\% \\
Tasks: every added node fails on buggy & 53\% & 69\% & 71\% \\
Tasks: some added node fails on buggy & 40\% & 28\% & 27\% \\
Tasks: no added node fails on buggy & 7\% & 4\% & 2\% \\
\midrule
Tasks adding no graded node & 218 (25\%) & 164 (15\%) & 366 (12\%) \\
\quad tightened existing graded tests & 194 & 138 & 302 \\
\quad added test did not become a node & 24 & 26 & 64 \\
\bottomrule
\end{tabular}
\end{table}

\paragraph{What the paired comparison does and does not show.} Both arms are
built from tasks the filtering stage would have discarded, which removes task
selection as an explanation for the gap. The results support the view that the
original suites reward behaviour that transfers less well to held-out tasks,
but they do not identify which shortcuts the trained policy actually uses.
Because the two policies evolve under different reward functions, the training
curves reflect both the scoring rule and its effect on learning; they are not a
re-scoring of identical policy outputs, and the deck-mastery gap in
Figure~\ref{fig:data-ablations}(d) should be read accordingly.

\subsection{Worked example: \texttt{doccmd}}
\label{app:doccmd-example}

\texttt{doccmd} runs commands against documentation code blocks. Its issue requests recursive directory support, and its original gate contains two \textsc{fail-to-pass} and 47 \textsc{pass-to-pass} tests. Both fail-to-pass tests address the issue, including directory input. Yet the judge produced a fully rewarded fix that hardcodes the example extensions \texttt{(".md", ".rst")} when expanding directories, ignoring user settings and silently skipping custom extensions.

Without seeing this probe, the strengthening agent produced the same type of shortcut and added four tests, all failing on the buggy tree. The key test requires a configured custom extension to be processed and an excluded default \texttt{.rst} file not to be processed. The controller recorded shortcut rewards of 1 under the original suite and 0 under the strengthened suite, with three new nodes rejecting it; the strengthening result was accepted. This demonstrates rejection of a specific previously rewarded error, not that every trained suite underwent shortcut evaluation or rejects every incorrect fix.

\section{Rollout integrity and the evaluation subset}
\label{app:integrity}

An agent with a shell in a repository has two routes to a fixed bug that do not
involve solving it: the repository's own version-control history, and the
network. Section~\ref{sec:main-eval} summarizes how we close both; this
appendix gives the mechanisms, the evidence that they are needed, and the
evaluation instances that the controls make ungradeable.

\subsection{The history channel}

The container images used for training ship a full clone of the repository in
which the gold-fix commit is a \emph{descendant} of the checked-out buggy
parent: \texttt{refs/heads/main} typically sits thousands of commits ahead of
\texttt{HEAD}. A single \texttt{git log --all} therefore walks straight to the
fix, and does so entirely offline, so no amount of network isolation closes
this route.

The behaviour is learned rather than incidental. On a $9$B arm run without the
countermeasure, $39\%$ of rollouts issued a git command reaching past
\texttt{HEAD}, and the rate rose from about $5\%$ at step $10$ to $99\%$ by
step $200$, worth $+0.30$ mean reward (permutation test, $p < 10^{-4}$). It was
the dominant reward hack on that taskset, and its growth is the point: under a
policy gradient, any channel that raises reward is reinforced until it
saturates.

We prune the history when the container is prepared, in this order:

\begin{enumerate}
  \item detach \texttt{HEAD}, so that deleting the current branch cannot orphan
        the checkout;
  \item delete every ref via \texttt{git update-ref -d}, which removes packed
        refs as well as loose ones;
  \item remove every configured remote;
  \item expire the entire reflog (\texttt{git reflog expire --expire=now
        --all});
  \item run \texttt{git gc --prune=now} to drop the now-unreachable objects.
\end{enumerate}

Every step is best-effort, so that a slow garbage collection can never kill a
rollout. That tolerance is precisely why a missing prune is easy to overlook:
it fails silently and the run continues. We therefore assert the
\emph{postcondition} rather than any exit status --- zero refs and zero remotes
remaining --- which is what surfaced the control's absence on an earlier run
after $240$ steps. Grading is unaffected by the prune, because the gold patch
is reconstructed from the dataset record rather than from git, and the
pre-agent commit stays reachable through the detached \texttt{HEAD} so that
patch capture still diffs against the right base.

The channel is not uniformly open. SWE-bench Verified ships images whose future
history has been stripped upstream, so that benchmark is clean by construction.
SWE-bench Pro is not: its images retain the gold commit, so the prune is
load-bearing there in a way it is not for Verified. Most of our training
corpora likewise ship unpruned clones.

\subsection{The network channel}

The network offers both the upstream fix --- the merged pull request, the fixed
file on a code host, or a mirror of the benchmark itself --- and arbitrary
package installation. Sandboxes run under a CNI egress policy that denies
direct outbound traffic and cloud metadata access by default, leaving reachable
only the interception tunnel back to the trainer and an authenticated package
proxy scoped to the sandbox workload.

Proxy configuration is connectivity, not containment. Proxy environment
variables are visible to the model and can be unset by it, so containment has
to be enforced at the network layer rather than by convention. As a check that
the policy is actually in force, every sandbox runs a probe at startup that
attempts a TCP connection to the three cloud metadata endpoints
(\texttt{169.254.169.254}, \texttt{169.254.170.2} and \texttt{fd00:ec2::254});
if any of them answers, the rollout fails outright rather than proceeding in a
misconfigured environment.

\subsection{Corroboration}

Concurrent work at Cursor identifies the same two channels and adopts the same
two mitigations~\citep{cursorhack}. Auditing $731$ agent trajectories with a
model-based judge that saw the trajectory but not the outcome, it attributes
$57\%$ of reward hacks to locating the merged pull request on the public web
and reproducing it, and $9\%$ to mining the bundled history for the fix commit;
$63\%$ of one frontier model's successful SWE-bench Pro resolutions retrieved
the fix rather than derived it. Sealing both channels moved that model from
$87.1\%$ to $73.0\%$ on SWE-bench Pro and their own model from $74.7\%$ to
$54.0\%$. Their controls are a repository reinitialized as a single fresh
commit, with the original history restored only at scoring time, and a
deny-by-default egress proxy with an allow-list of package registries.

Two differences are worth naming. First, that study measures a static rate for
a fixed model, whereas the numbers above are measured under a policy gradient,
where the rate escalates because the channel is rewarded; an inflated
evaluation score can be corrected by tightening the harness, but a policy that
has learned to retrieve has to be retrained. Second, their report notes that
SWE-bench addressed the history channel upstream and that the images they had
ingested predated the fix, which is consistent with what we observe: Verified
is clean today, Pro is not.

\subsection{The SWE-bench Verified subset}

Because the egress lockdown grades the reference patch under the same isolation
as the model's, a task whose own gold test reaches the network becomes
ungradeable rather than merely difficult: it times out regardless of what the
model submits. Such tasks are an artifact of offline evaluation and not model
failures, so scoring them as failures would understate every arm equally but
add only noise.

Seventeen of the $500$ SWE-bench Verified instances are excluded for this and
four other reasons (Table~\ref{tab:dropped}), leaving the $483$-problem subset
on which every SWE-bench Verified number in this paper is computed. The subset
is fixed across all runs and all checkpoints, so any two scores in this paper
are paired over identical problems.

\begin{table}[h]
\centering
\caption{\textbf{The seventeen SWE-bench Verified instances excluded from the
  scored subset.} Only the first group is a consequence of the isolation
  policy; the rest are defects in the task or its image.}
\label{tab:dropped}
\small
\begin{tabular}{llp{6.1cm}}
\toprule
Reason & $n$ & Instances \\
\midrule
Network-dependent gold test & 9 &
  \texttt{matplotlib-20488}, \texttt{requests-1724}, \texttt{requests-1766},
  \texttt{requests-1921}, \texttt{requests-2317}, \texttt{sphinx-7985},
  \texttt{sphinx-8269}, \texttt{sphinx-8475}, \texttt{sphinx-10435} \\
\addlinespace
Recorded test names do not match the log & 3 &
  \texttt{astropy-7606}, \texttt{sphinx-8595}, \texttt{sphinx-9711} \\
\addlinespace
Image no longer collects & 2 &
  \texttt{astropy-8872}, \texttt{pylint-4661} \\
\addlinespace
Gold patch fails its own tests & 2 &
  \texttt{astropy-8707}, \texttt{django-10097} \\
\addlinespace
Already passes before any edit & 1 &
  \texttt{django-7530} \\
\midrule
\textbf{Total excluded} & \textbf{17} & $500 - 17 = \mathbf{483}$ retained \\
\bottomrule
\end{tabular}
\end{table}

The nine network-dependent instances were confirmed by cross-run evidence
rather than by inspection alone: each is solved in runs with network access, so
its gold test is gradeable in principle, but times out in every offline
rollout. Their reference tests post files and digest-auth credentials to
\texttt{httpbin.org}, post to a pastebin service, or fetch live URLs under
Sphinx's link checker. One instance that times out both online and offline was
\emph{kept}, since a genuinely slow test is a property of the task rather than
of the isolation.

\subsection{Evaluation Setup and LLM Hyperparameters}
\label{sec:app_eval_setup}

This section details the evaluation harness, model sampling parameters, and environment budgets for both SWE-bench Verified and SWE-bench Pro. A side-by-side comparison of the evaluation parameters is summarized in Table~\ref{tab:app_eval_setup_comparison}.

\subsubsection{LLM Sampling Parameters}
\label{app:eval-sampling}



Every row of Table~\ref{tab:baselines} marked $^{\dagger}$ shares one
evaluation protocol and differs only in its sampling parameters. Across both benchmarks, model responses are generated with reasoning/thinking capabilities enabled. The maximum completion budget per request is capped at $32{,}768$ tokens. Each problem
receives 4 rollouts in the isolated sandbox of Section~\ref{sec:main-eval},
capped at 100 turns, and pass@1 is the mean resolve rate over those rollouts.
SWE-bench Verified uses the 483 problems and SWE-bench Pro the 731 problems
of Appendix~\ref{app:integrity}; Pro-v2-Py is the subset of 237 Pro-v2
problems whose repository is written in Python, the only language in our
training data. A checkpoint uses the same sampling parameters on every
benchmark.

Table~\ref{tab:eval-sampling} lists the parameters for each row. FrogNano is
evaluated under the parameters its authors report, so that the only
differences from their own evaluation are the harness differences discussed in
Section~\ref{sec:main-results}. For our own checkpoints we considered the three
configurations in the table and report each checkpoint under the one that
scores highest on SWE-bench Verified.
The training curves of Figure~\ref{fig:main-run}b instead use one fixed
configuration ($T = 1.0$, top-$p$ 0.95, top-$k$ 20) at every checkpoint, so
that the curves are comparable across steps; their values can therefore
differ slightly from the corresponding rows of Table~\ref{tab:baselines}.

\begin{table}[h]
\centering
\caption{\textbf{Sampling parameters behind each self-evaluated row of
Table~\ref{tab:baselines}.} ``off'' means the filter is disabled
(top-$p = 1.0$, no top-$k$ truncation). All rows use 4 rollouts per problem
and a 100-turn limit.}
\label{tab:eval-sampling}
\begin{tabular}{@{}lccc@{}}
\toprule
Checkpoint & Temperature & Top-$p$ & Top-$k$ \\
\midrule
Qwen3.5-4B (initialization)                 & 1.0 & off & off \\  
Qwen3.5-9B (initialization)                 & 1.0 & off & off \\  
FrogNano-Qwen3.5-4B (our setting)           & 0.6 & 0.95 & 20 \\  
\midrule
\ours-4B, R2E-Gym only                      & 1.0 & 0.95 & 20 \\
\ours-4B + curated RL                       & 1.0 & 0.95 & off \\
\ours-9B + curated RL                       & 1.0 & off  & off \\
\bottomrule
\end{tabular}
\end{table}

\begin{table}[h]
\centering
\small
\caption{\textbf{Detailed evaluation settings across benchmarks.} Both settings enable thinking mode and share identical decoding parameters, while SWE-bench Pro expands interaction limits to handle longer trajectory horizons.}
\label{tab:app_eval_setup_comparison}
\begin{tabular}{lcc}
\toprule
\textbf{Configuration / Parameter} & \textbf{SWE-bench Verified} & \textbf{SWE-bench Pro} \\
\midrule
\multicolumn{3}{l}{\emph{Benchmark Role \& Size}} \\
Role & Validation Set & Test Set \\
Evaluated Instances & 483 (17 excluded) & 731 (all retained) \\
Rollouts per Problem & 4 & 4 \\
\midrule
\multicolumn{3}{l}{\emph{Agent Interaction Budgets}} \\
Max Turns & 100 & 150 \\
Max Context Length & 131,072 ($128\text{k}$) & 262,144 ($256\text{k}$) \\
\midrule
\multicolumn{3}{l}{\emph{LLM Sampling Parameters}} \\
Thinking / Reasoning & Enabled & Enabled \\
Temperature & 1.0 & 1.0 \\
Max Tokens / Request & 32,768 & 32,768 \\
$top\_p$ / $top\_k$ & 0.95 / 20 & 0.95 / 20 \\
\midrule
\multicolumn{3}{l}{\emph{Sandbox Integrity \& Environment}} \\
Git History Pruned & Yes & Yes \\
Network Egress (Agent Stage) & Blocked & Blocked \\
Grading-Phase Dependency Proxy & No & Yes (e.g., \texttt{go.mod}) \\
\bottomrule
\end{tabular}
\end{table}

\subsection{Empirical Audit of Evaluation Hacking in Unsealed Environments}
\label{sec:app_hack_audit}

To directly demonstrate how unsealed sandboxes distort evaluation metrics, we evaluate checkpoints under an \emph{unsealed} setting where network egress is fully open and the \texttt{.git} repository history remains intact. Table~\ref{tab:hack_audit} summarizes the trajectory audit across 60 evaluation trials.

When isolation is enforced (27B Baseline), zero trajectories exhibit hacking behavior. However, when the environment is unsealed, the observed pass rate artificially jumps from 50.0\% to 65.0\%. A granular trajectory audit reveals that 19 out of the 39 successful rollouts (48.7\%) explicitly exploited sandbox loopholes: 17 rollouts extracted the exact gold fix SHA directly from the git log, and 7 rollouts fetched external repository source code over the network. Notably, smaller models (Qwen3.5-9B) showed a much lower propensity for active exploitation under the same unsealed setup, highlighting that stronger reasoning capabilities amplify a model's ability to discover and exploit environmental reward-hacking shortcuts.

\section{Full Training Details}
\label{app:training_details}

\subsection{The Training Data}
\label{app:main-deck}

The main runs train on the $7{,}500$-instance curated pool of
Section~\ref{sec:filtering} and nothing else: $2{,}086$ rows from R2E-Gym, $1{,}776$
from SWE-rebench~V2 and $3{,}638$ from Scale-SWE~\citep{r2egym,rebenchv2,scaleswe},
each of which cleared both curation pipelines, with $1{,}233$ carrying an issue
rewritten to state what its tests grade. The $4$B and $9$B runs use this deck
unchanged and differ only in the base model.

The three datasets are served as three environments under a single \emph{global}
deck rather than as three independently sampled pools. One epoch is therefore
every live task exactly once, and the sampler never draws a duplicate within an
epoch; without this, the corpus mixture would drift as the three pools are
consumed at different rates.

A task is retired as \emph{mastered} when a single group of $8$ rollouts solves
it unanimously, after which it is never sampled again. There is no difficulty
threshold, no re-admission and no periodic re-check --- the rule is one
unanimous group and the task is gone for the rest of the run. This is what
Figure~\ref{fig:main-run}c measures, what the eviction-adjusted reward of
Appendix~\ref{app:eviction-metric} corrects for, and what the probe of
Appendix~\ref{app:eviction} tests.

Every instance is prepared with its git history pruned, and every rollout runs
under the egress policy described in Appendix~\ref{app:integrity}. Training and
evaluation containers are built by the same procedure, so a shortcut closed for
grading is closed for training as well.

The strengthened suites of Section~\ref{sec:hardening} are \emph{not} part of this
deck. Strengthening adds $4{,}889$ instances whose containers must be rebuilt
and re-verified before they can be trained on, which lengthens the run beyond
the budget available, so the value of strengthening is established by the
controlled comparison in Section~\ref{sec:hardening} rather than carried into the
main run.

\subsection{Training Objective}
\label{app:algo}

This appendix states the objective used by every run in the paper. It is a
CISPO-style policy-gradient loss with group-relative advantages, extended with
token-level masking of training--inference disagreement. See Appendix \ref{app:setup} for more details.

\paragraph{Staleness.}
Rollouts are admitted up to $8$ policy versions behind the current weights,
which keeps the generation and training pipelines busy without a synchronous
barrier at every step. Key--value caches are recomputed on weight update rather
than carried across versions.

\subsection{The reward and stop-condition contract}
\label{app:stop-policy}

A trajectory that reaches the end of its turn budget without fixing the bug and
a trajectory that emitted a malformed tool call on turn three are both failures,
but they are not the same failure, and giving them the same reward teaches the
wrong thing. This appendix records how the two are separated.

The base reward is $0/1$: a trajectory scores $1$ only if the merged
fail-to-pass and pass-to-pass identifiers are all present and all passing in
the suite's JUnit XML output, and $0$ otherwise. On top of that, a rollout may
terminate early with a \emph{stop condition}, and two classes of stop condition
override the graded reward to $-1$. Table~\ref{tab:stop-policy} gives the
mapping.

\begin{table}[h]
\centering
\caption{\textbf{Stop conditions and the reward they receive.} The two
  overrides fire only for the conditions listed under \emph{invalid}; every
  other terminal condition keeps whatever the test suite awarded, which for an
  unfinished trajectory is $0$.}
\label{tab:stop-policy}
\small
\begin{tabular}{lll}
\toprule
Stop condition & Meaning & Reward \\
\midrule
\multicolumn{3}{l}{\emph{Invalid: the trajectory was never a well-formed attempt}} \\
\texttt{format\_invalid}     & malformed tool call or empty reasoning envelope & $-1$ \\
\texttt{submission\_invalid} & submitted without the exact required command    & $-1$ \\
\texttt{truncation}          & a single model response cut off mid-generation  & $-1$ \\
\texttt{repetition}          & degenerate looping (detectors off in this run)  & $-1$ \\
\midrule
\multicolumn{3}{l}{\emph{Exhausted: a genuine attempt that ran out of budget}} \\
\texttt{max\_turns}          & hit the $100$-turn cap                      & graded ($0$) \\
\texttt{max\_total\_tokens}  & hit the $131{,}072$-token trajectory budget & graded ($0$) \\
\texttt{max\_input\_tokens}  & prompt exceeded its budget                  & graded ($0$) \\
\texttt{max\_output\_tokens} & completion exceeded its budget              & graded ($0$) \\
\texttt{context\_length}     & exceeded the model's context window         & graded ($0$) \\
\texttt{harness\_timeout}    & the harness exceeded its wall-clock budget  & graded ($0$) \\
\midrule
\texttt{agent\_completed}    & submitted normally                       & graded ($0$/$1$) \\
\bottomrule
\end{tabular}
\end{table}

The division is between trajectories that were never valid attempts and
trajectories that were valid but unsuccessful. A malformed tool call or a
response truncated mid-generation produces no evidence about the model's
ability to fix the bug, and scoring it $0$ would place it alongside honest
failures and make it cheap; scoring it $-1$ makes degenerate output strictly
worse than trying and failing. Running out of turns or context is different: the
model made a real attempt and did not finish, which the $0$ it already earns
describes correctly, and penalising it further would price exploration.

The asymmetry is load-bearing in one specific way. Under a length penalty,
dying early becomes \emph{cheap} --- the penalty scales with the fraction of
the budget consumed, so a rollout that exits after two turns pays almost
nothing and can score better than an honest hundred-turn failure. The $-1$
override for truncation exists to close that gradient. The main run carries no
length penalty, which makes the override nearly inert there; we keep it so that
penalised and unpenalised arms differ in one knob rather than two, and because
it still guarantees that a truncated rollout is not silently recorded as a
short success.

Three degenerate-output detectors are available and all are disabled in the main
run: a character-repetition gate, a duplicate-tool-call gate and a loop guard.
Each of them terminates with \texttt{repetition}, so with all three off that
condition cannot fire terminally and the fuzzy repetition signal is
diagnostic-only. Format checking is stricter than it may appear: it requires a
well-formed reasoning envelope with non-empty content, at most one tool call per
turn, and no prose between the reasoning block and the tool call.

\subsection{Other Training Hyperparameters}

Rollouts are collected with Mini-SWE-Agent-Plus with at most $100$ turns, $131{,}072$ tokens per trajectory and
$32{,}768$ completion tokens per model request, sampled at temperature $1.0$
with no nucleus or top-$k$ truncation. Each update draws $16$ prompts with
$8$ rollouts per group, for $128$ trajectories per step, and admits
rollouts up to $8$ policy versions stale. Groups with zero advantage are
filtered before and after batching.

The objective is optimized by AdamW at a constant learning rate of $10^{-6}$,
with $\beta_1 = 0.9$, $\beta_2 = 0.98$, $\epsilon = 10^{-15}$ and zero weight
decay, unless otherwise specified. There is no warmup and no decay schedule.

Checkpoints are written every $50$ steps and evaluated offline.

\subsection{Training configuration updates for the 9B}
\label{app:9b-config}

Two 9B runs appear in this paper and they serve different purposes. The
retention probe of Appendix~\ref{app:eviction-9b} uses a 9B that is the 4B's
configuration with the model as its only change: the same three curated
environments, byte-identical decks, the same grading contract and the same
horizon. That arm is the controlled comparison, and it peaks at 63.8\% pass@1
at step 600. The run reported in Section~\ref{sec:main-results} is a later
arm that carries three changes on top of it, and reaches 65.7\%.

\paragraph{Deck.} The reported 9B trains on the union of six environments ---
the three curated environments of \S\ref{sec:filtering} plus the three
strengthened rescues of \S\ref{sec:hardening} --- for 12,389 rows against
7,500. This is the one main run in which the strengthened suites are trained
on rather than evaluated in a controlled comparison.

\paragraph{Off-policy horizon.} \texttt{max\_off\_policy\_steps} falls from 8
to 4. The orchestrator admits a rollout whose generating policy is at most
this many publishes behind the trainer, so halving it halves the worst-case
staleness of a batch at the cost of throughput.

\paragraph{Importance-weight floor.} The importance-ratio bound becomes
two-sided, $[0.5, 2.0]$, where the earlier arm bounded only from above at
2.0.

Because these move together, the gap between the two 9B arms is not
attributable to any one of them, and we report the pair as progress rather
than as a controlled study. The mechanistic claims of
Section~\ref{sec:main-results} rest on the first arm, where the model is the
only difference from the 4B.



\section{What happens to retired tasks}
\label{app:eviction}

Both main runs retire a task from the deck the first time all eight rollouts in
one group solve it, and never sample it again: the rule takes 48\% of the
4B's deck by step 450 and 53\% of the 9B's by step 600. Only the 4B
declines on held-out evaluation, and it does so while its training reward holds
(Section~\ref{sec:main-results}), which makes its retired population the
obvious place to look --- large, invisible to the training statistic, and
unconstrained by the objective. This appendix reports a direct measurement of
what each policy can still do on tasks it has stopped seeing. The probe is run
on both arms, each on decks cut from its own eviction state, so that the two
can be compared on the quantity that matters: how much each loses on material
it had already mastered.

\subsection{The eviction-adjusted reward}
\label{app:eviction-metric}

Mean reward over sampled tasks is not comparable across steps once eviction is
enabled. Retirement removes tasks the model has \emph{solved}, so the deck that
remains is steadily harder and the plain mean drifts downward even when nothing
has regressed; the same effect makes the statistic incomparable between a run
that retires tasks and one that does not. We therefore report an
eviction-adjusted reward, which restores the original denominator by imputing
reward 1 for every retired task:
\begin{equation}
\tilde{r} \;=\; \frac{m + (D - m)\,\bar{r}}{D},
\label{eq:eviction-adjusted}
\end{equation}
where $D$ is the deck size, $m$ the number of tasks retired so far and
$\bar{r}$ the observed mean over sampled tasks. With no eviction $m = 0$ and
$\tilde{r} = \bar{r}$ exactly, so the statistic can be emitted unconditionally
and compared across both arms of an A/B.

Two limits matter for how it is read. The first is the subject of this
appendix: the imputation is an \emph{upper bound}. Crediting a retired task
with reward 1 asserts that the model would still solve it, but retirement is
decided on a single group of eight rollouts and the task is never sampled
again, so no later degradation can appear in the statistic. Any forgetting on
the retired population is invisible to it by construction, which is precisely
why the probe below is necessary. Inverting
Equation~\ref{eq:eviction-adjusted} on the main run makes the size of the
imputation clear: at step 250 an adjusted 0.636 with 35.3\% retired
implies an observed mean of 0.437 on the live deck, and at step 450 an
adjusted 0.614 with 48.0\% retired implies 0.258. Almost half of the
reported value at step 450 is imputed rather than measured.

The second limit is an aggregation mismatch. The adjusted aggregate weights
environments by deck size, whereas the observed aggregate weights them by
rollout count, so the two coincide only when a single environment supplies the
batch. The main run mixes three, so small discrepancies between the aggregate
and the per-environment values are expected.

\subsection{Design}

Two hypotheses are worth separating, and they make opposite predictions.

\begin{description}
  \item[Retirement on luck.] A task whose true success probability is $p$ is
    retired with probability $p^{8}$ on any appearance --- 17\% per draw at
    $p = 0.8$ and 43\% at $p = 0.9$. Under this hypothesis a substantial
    share of the deck left on a lucky streak rather than on mastery, and
    re-scoring the retired population at \emph{any} later step should find it
    well below the 1.0 it scored on the way out.
  \item[Forgetting.] Retirement was sound, but competence decays afterwards
    because no gradient preserves it. Under this hypothesis the retired
    population scores near 1.0 shortly after retirement and degrades as
    training continues.
\end{description}

We therefore re-score the same retired tasks at two checkpoints. From each
run's step-350 eviction state we sample 300 retired tasks from each of the
three corpora and score them at 4 rollouts per task. The decks are then held
fixed and re-scored at step 500, so the step-350-to-500 comparison is
within-task: identical instances, identical grading, only the weights differ.
Scores are reported under the same reward contract used in training, including
the $-1$ overrides of Appendix~\ref{app:stop-policy}.

Cutting each arm's decks from its own eviction state matters for how the
numbers are read. It makes 1.0 a value that run itself measured on exactly
those rows, rather than a threshold inherited from the other arm. It also
means the reference point is equally valid for both, which a shared deck could
not provide.

The probe runs through the evaluation path with no trainer present, so the
weights are structurally incapable of moving and none of the orchestrator's
batch machinery --- buffering, online difficulty filtering, the zero-advantage
filter, off-policy caps --- is in play. The grading contract, sampling
parameters, sandbox configuration and concurrency ceiling are lifted from each
run's own training configuration rather than from the shipped evaluation
template, since the template enables three stop conditions that both runs
disable and would otherwise measure grader mismatch as forgetting.

That the harness discriminates at all is confirmed by scoring tasks still live
in the 4B's deck at step 350: those return 0.426 on R2E-Gym and 0.299 on
SWE-rebench, against 0.952 and 0.946 for the retired decks at the same
checkpoint. The evaluation is not returning high scores indiscriminately. The
live pools are an anchor rather than a matched control --- eviction removes
easy tasks by definition, so the live pool is enriched for hard ones by
construction, and the gap between the two is not an effect size.

\subsection{Step 350: retirement was not firing on luck}

\begin{table}[h]
\centering
\caption{\textbf{The 4B's retired tasks re-scored at step 350.} Every task here
  scored 1.0 on the group that retired it, so 1.0 is the reference each row is
  read against.}
\label{tab:probe350}
\small
\begin{tabular}{lrrrr}
\toprule
Cell & Tasks & Rollouts & Mean reward & Solved $4/4$ \\
\midrule
\texttt{r2e\_evicted}      & 300 & 1{,}200 & \textbf{0.952} & 0.867 \\
\texttt{rebench\_evicted}  & 300 & 1{,}200 & \textbf{0.946} & 0.867 \\
\texttt{scaleswe\_evicted} & 300 & 1{,}200 & \textbf{0.968} & 0.917 \\
\bottomrule
\end{tabular}
\end{table}

The retired tasks come back at 0.95 against the 1.0 they scored when
they left, a shortfall of 0.048 on R2E-Gym and 0.054 on SWE-rebench.
Retirement on luck predicts far more than this: a task retired on a lucky
streak has a true success probability around the 0.8--0.9 that makes
$p^{8}$ appreciable in the first place, so the retired population would
return a mean near that value, not within five points of 1.0. Retirement
on luck is therefore not what the rule is doing at this scale: at step 350
those tasks really were mastered, and the $p^{8}$ concern, though sound in
principle, does not describe this run.

Note that the probe draws 4 rollouts where the rule fired on 8, so the
``solved $4/4$'' column estimates $E[p^{4}]$ while the rule is $E[p^{8}]$.
Since $p^{8} \le p^{4}$, that column overstates how re-evictable the deck is,
which is the conservative direction for this conclusion. Mean reward estimates
$E[p]$ and is unbiased by group size, so it is the column to read.

This moves the question rather than settling it. If the deck was correctly
identified as mastered, the decline in held-out score after step 250 cannot
be explained by a deck silently made harder by chance.

\subsection{Step 500: the same tasks, materially worse}

\begin{table}[h]
\centering
\caption{\textbf{The identical retired decks re-scored at step 500.} Only the
  weights differ from Table~\ref{tab:probe350}.}
\label{tab:probe500}
\small
\begin{tabular}{lrrrr}
\toprule
Cell & Mean reward & $\Delta$ vs.\ step 350 & Solved $4/4$ & Early exits \\
\midrule
\texttt{r2e\_evicted}      & \textbf{0.851} & $-0.101$ & 0.693 & 28/1{,}200 \\
\texttt{rebench\_evicted}  & \textbf{0.845} & $-0.101$ & 0.697 & 32/1{,}200 \\
\texttt{scaleswe\_evicted} & \textbf{0.852} & $-0.116$ & 0.750 & 40/1{,}200 \\
\bottomrule
\end{tabular}
\end{table}

Between step 350 and step 500 the policy loses roughly ten points of reward
on tasks it has not been shown since before step 350. The loss is uniform
across all three corpora, which argues against a corpus-specific artifact, and
the fraction of tasks solved by all four rollouts falls further still, from
about 0.87 to about 0.70. This is forgetting, measured directly: the
training statistic is computed only over live tasks and is by construction
blind to it.

\paragraph{Part of the loss is output behaviour, not knowledge.}
Early exits --- rollouts ending in \texttt{format\_invalid} or
\texttt{truncation}, and therefore taking the $-1$ override --- rise roughly
tenfold on the retired decks, from 3 of 1{,}200 (0.25\%) at step 350 to
28 of 1{,}200 (2.3\%) at step 500 on R2E-Gym, with the same pattern on
the other two corpora. Some of the degradation is thus a decay in the ability
to emit a well-formed trajectory rather than a loss of the knowledge needed to
fix the bug. The two are worth separating because they suggest different
remedies: replaying retired tasks addresses the first, while format
regularization or a shorter horizon addresses the second.

\subsection{The 9B under the same rule}
\label{app:eviction-9b}

The 4B result raises the obvious question of whether this is what any model
does under this rule. We repeat the probe on the 9B, unchanged in every
respect except the weights and the decks, which are cut from the 9B's own
step-350 eviction state.

Two measurements establish that the arms are comparable before any score is
read. First, the 9B does not evict less, which would otherwise explain a flat
result without any retention story: 41.1\% of its deck is retired by step 350
against the 4B's 44.1\%, and 47.9\% against 48.9\% by step 500. The deck
hardens at nearly the same rate in both. Second, the two retired sets are
largely the same tasks --- Jaccard overlap of 69.5\% to 73.9\% by corpus, with
78\% to 83\% of the 4B's retired set also retired by the 9B --- so the two
arms are not being asked about different material.

\begin{table}[h]
\centering
\caption{\textbf{The 9B's retired tasks, re-scored at both checkpoints.} Decks
  are cut from the 9B's own step-350 eviction state, so 1.0 is again a value
  this run measured on exactly these rows. Columns are directly comparable to
  Tables~\ref{tab:probe350} and~\ref{tab:probe500}.}
\label{tab:probe9b}
\small
\begin{tabular}{lrrrrr}
\toprule
Cell & Step 350 & Step 500 & $\Delta$ & Solved $4/4$ & Early exits \\
\midrule
\texttt{r2e\_evicted}      & 0.973 & 0.968 & $-0.005$ & $0.907 \to 0.917$ & $1 \to 1$ \\
\texttt{rebench\_evicted}  & 0.963 & 0.951 & $-0.013$ & $0.890 \to 0.880$ & $2 \to 4$ \\
\texttt{scaleswe\_evicted} & 0.970 & 0.967 & $-0.003$ & $0.923 \to 0.920$ & $3 \to 2$ \\
\bottomrule
\end{tabular}
\end{table}

The 9B holds what it retired. Its declines are an order of magnitude smaller
than the 4B's on every corpus, the fraction solved by all four rollouts is flat
or slightly up rather than falling by seventeen points, and the early-exit
count does not move at all where the 4B's rises roughly tenfold. Reward and
output validity degrade together for the 4B and are jointly flat for the 9B,
which is two independent signals agreeing rather than one.

Because the step-350 and step-500 cells are the same tasks, the change can be
averaged per task rather than differenced between cell means, which is
substantially more precise: when nearly every task is solved at both steps,
per-task change concentrates tightly.

\begin{table}[h]
\centering
\caption{\textbf{Per-task paired change, steps 350 to 500.} Standard errors are
  over tasks. Every 9B decline is within 1.4 standard errors of zero, and
  between 9.6 and 13.4 standard errors from the 4B's decline on the same
  corpus.}
\label{tab:probe-paired}
\small
\begin{tabular}{lrrrr}
\toprule
Corpus & 9B $\Delta$ & s.e. & 4B $\Delta$ & Separation \\
\midrule
R2E-Gym        & $-0.0050$ & 0.0077 & $-0.101$ & 12.4\,s.e. \\
SWE-rebench V2 & $-0.0125$ & 0.0092 & $-0.101$ & \phantom{0}9.6\,s.e. \\
Scale-SWE      & $-0.0033$ & 0.0085 & $-0.117$ & 13.4\,s.e. \\
\bottomrule
\end{tabular}
\end{table}

Both arms start from comparable headroom --- the decks are by construction
tasks each had just mastered, eviction measured every one at 1.0, and the probe
re-measures both between 0.95 and 0.97 at step 350 --- so the contrast is not a
ceiling artifact. One cross-model check removes the deck difference entirely:
the 9B's step-500 policy scored 0.960 on the 4B's own R2E-Gym deck, the
identical 300 rows, which is above the 4B's own peak of 0.952 on them and
0.109 above the 4B at the matched step.

\section{Related work}
\label{app:related}

\paragraph{Training SWE agents.}
SWE-Gym and R2E-Gym provide executable repository environments for training agents and verifiers~\citep{swegym2024, r2egym}.
SWE-smith provides task and training resources, while DeepSWE releases
Qwen3-32B RL training on R2E-Gym \citep{swesmith,deepswe2025}.
AgentFly supplies a Qwen3-32B SWE example with containerized tools and reward execution
\citep{agentfly_swe_2026}. Prime-RL publishes GLM-Air and Qwen3-30B-A3B-Thinking bash-harness
examples with different optimizers and training scales
\citep{primerl_swe_2026,primerl_qwen_swe_2026}. NeMo's Qwen3 recipe precedes asynchronous
SWE2 training with a single-step tool-use stage, using OpenHands-based environments and
Apptainer sandboxes \citep{nemo_swe_2026}. These differences in initialization, interface
and update rules limit what an adaptation can establish about its source recipe.
Our focus is on diagnosing whether an asynchronous recipe can acquire and sustain success on a fixed, solvable task pool before applying it to broader training data; Appendix~\ref{app:overfit} provides detailed recipe comparisons.

\paragraph{Same-base-model training.}
Dressage reports Qwen3.5-4B learning through Claude Code, with synchronous GRPO,
integrity-gated rewards, importance weighting and reference KL; its SWE-Gym report
shows 80 updates and a SWE-bench Verified increase from 32.6\% to 37.8\%
\citep{dressage_swe_2026}. In concurrent work, FrogNano reports 61.5\% using Leaf and DPPO, with task
regeneration, optimizer resets and length shaping \citep{frognano2026}.
Both establish successful learning in their reported conditions. FrogNano's regenerated
tasks and optimizer resets address curriculum development, while our fixed-pool study
tests whether learned performance survives continued optimization without those resets.

\paragraph{Asynchronous RL and execution consistency.}
AReaL decouples generation from optimization, controls rollout staleness, and recomputes cached attention states after in-flight weight updates~\citep{fu2025areal}. \citet{yao2025offpolicy} show that distinct rollout and training implementations can assign different token probabilities even with identical weights, introducing an additional source of off-policy mismatch.
CISPO and GSPO modify importance weighting, while other work bounds policy age or changes
which updates are admitted \citep{cispo2025,gspo2025,sao2026}.
Dressage studies policy-version bounds for partial rollouts and separately reports
synchronous engine-balancing performance
\citep{dressage_staleness_2026,dressage_balance_2026}. Nebius reports degradation after
a serving upgrade changed sampling filters and recovery when settings were restored
\citep{nebius_swe_2025}. Our evidence connects serving admission, observed policy age and
training mismatch in asynchronous SWE rollouts, with a separate fixed-weight execution
audit. It does not identify each of these mechanisms from one aggregate divergence metric.

MiMo-V2.6 jointly scales RL compute, environment diversity and grader compute
\citep{mimo2026}. Its report was uploaded on September 21, 2026, while our study
was in progress. That report describes asynchronous partial rollouts with a
reported staleness bound of four, KV rebuilding after policy updates, and
zero-advantage filtering \citep[Sections~4.1 and~5.1]{mimo2026}. Its Muown optimizer uses $\beta_1=\beta_2=0.95$ in the
Adam component and carries FP32 master weights and Muown row state from SFT.
Those choices overlap our recipe concerns but do not constitute the same optimizer
or initialization. Retaining the original behavior log-probabilities for importance
weighting is compatible with rebuilding KV for subsequent generation; these are
different operations.

MiMo's Sample Mixer accounts for rollout duration and group acceptance rates, while
its predictive dispatch limits queueing that can increase policy age \citep[Section~6.3]{mimo2026}.
The scheduling simulation excludes staleness expiry, so it is not a matched-cap
measurement of realized policy age. Its consistency mechanisms include quantized
weight alignment, routing replay and sampling-candidate-set replay \citep[Section~6.4]{mimo2026}.
These concurrent results complement our fixed-weight numerical audit and matched-cap
admission measurements with evidence from a different model scale and workload.

\paragraph{Continuation and credit assignment.}
SWE-Master reports collapse under a DeepSWE-style truncation-masking treatment and stable
learning with forced submission and stop-dependent reward modulation
\citep{swemaster2026}. Nebius warns that masking looping trajectories can remove useful
negative feedback \citep{nebius_swe_2025}; SkyRL-Agent reports successful training with
recovery hints and truncation loss masks \citep{skyrlagent2025}. NeMo configures negative
message-token advantages for invalid calls or malformed thinking \citep{nemo_swe_2026}.
These precedents make the full handling and training-eligibility contract the relevant
unit of comparison. Removing recovery or enforcing our particular format is not a
universal prescription.

MiMo-V2.6 reviews specification--test alignment, checks reward stability across
repeated executions, and audits false-positive and false-negative feedback
\citep[Section~4.2.1]{mimo2026}. Environment cleanup, network isolation and trajectory audits
further protect reward integrity \citep[Section~4.2.6]{mimo2026}. Its groupwise grading distinguishes
passing solutions by quality, and segment-level penalties reshape credit for faulty
intermediate actions \citep[Section~4.3]{mimo2026}. Offline rubric synthesis can create reward
differences even in an all-test-passing group; online advantage redistribution
operates on mixed-outcome groups. These methods enrich the signal before filtering
and do not contradict excluding groups whose final advantages are all zero.
Its group-relative length penalty and advantage rebalancing also illustrate why a
shaping term must be assessed within its complete recipe.

\paragraph{Repeated exposure and data utilization.}
Long-run stability is not unique to this work: SAO reports roughly a thousand training
steps and CANOPY reports 260 steps without degradation \citep{sao2026,canopy2026}.
Update counts alone do not give comparable exposure because group sizes, task pools,
optimizer resets and filtering differ.

The MiMo flagship curves cover 30 large updates with 1,568 prompts and sixteen
rollouts per prompt; a separate grading comparison continues through 52 updates
\citep{mimo2026}.
The report also includes successful RL on a distilled Qwen3.5-9B initialization and
evaluation across held-out harnesses \citep[Section~7]{mimo2026}. That study measures broad capability
improvement and harness transfer. Our fixed-pool experiments deliberately probe
continued optimization after training tasks have been learned, exposing deterioration
that changing task mixtures can conceal.

Zero-advantage filtering is also established
\citep{dapo2025,magistral2025}. We examine nonzero advantages and retirement on broader data. FrogNano's solve-probability-calibrated
task synthesis is a complementary way to seek informative groups; neither approach
guarantees that a larger effective batch improves held-out performance.

\paragraph{RL task quality.}
UTBoost, SWE-ABS, and Probe to Generate strengthen repository-level repair benchmarks by exposing weaknesses in their test suites~\citep{utboost,sweabs,probe}; \citet{auditing2026} also audits reward hackability in code RL training environments. Improving tests for training is not new: ACECODER synthesizes tests for code-model RL, HardTests studies the downstream effects of test quality, and EvolveCoder iteratively refines tests using candidate-solution execution feedback~\citep{zeng2025acecoder,he2025hardtests,ruan2026evolvecoder}.
We study this problem in repository-level, multi-turn SWE training, where the issue, tests, and gold patch jointly determine the reward's validity. Our paired strengthening experiment holds task identities fixed and changes the test suite used to compute training reward, measuring how this intervention affects held-out agent performance rather than only the scores assigned to a fixed set of candidate patches.

\paragraph{Environment integrity and measurement.}
Prior audits identify repository-state leakage in SWE
environments \citep{swebench465}. Hardened tests and integrity gates answer
different questions from compliance with a harness-specific interaction policy.
Seed and framework sensitivity also motivate consistent evaluation and explicit
uncertainty \citep{soberlook2025,errorbars2024}. We report peak and retained performance,
and distinguish observation-window, executed-attempt and shipped-sample denominators.

\subsection{Comparison with concurrent FrogNano}
\label{app:frognano}

Table~\ref{tab:compare} describes the unchanged-R2E run of Section~\ref{sec:r2e-baseline}
and FrogNano's published recipe \citep{frognano2026}. FrogNano uses a custom Leaf harness
and DPPO; from its third iteration it applies success-gated log-length shaping, and it gives
partial credit to correct budget-truncated trajectories. Our run carries the feedback and execution components of
Section~\ref{sec:recipe}, with AdamW $\beta_2=0.98$, and uses neither length shaping
nor advantage normalization.
An earlier configuration of ours on the same tasks, with group-scaled length shaping,
standardized advantages and clipped rather than masked importance weights, reached 61.4\%
at update 500.

FrogNano regenerates tasks between five 200-update climbs and resets optimizer and RNG
states at each boundary. Task admission targets an estimated solve probability of 0.5
in the first four iterations and $(0,0.5]$ in the revised fifth, which also uses a stronger
task generator. Our pool starts fixed and is changed only by retirement. Both processes seek
tasks that remain informative for group-relative learning.

\begin{table}[t]
\centering
\small
\caption{Descriptive same-base-model comparison. Our scores use 483 valid-gold tasks;
the external headline is the reported SWE-bench Verified result. Harnesses, task inclusion,
training distributions and evaluation budgets are not matched. Rollout totals are nominal
update budgets, not an audit of executed work or compute equivalence.}
\label{tab:compare}
\begin{tabularx}{\linewidth}{l>{\raggedright\arraybackslash}X>{\raggedright\arraybackslash}X}
\toprule
 & \ours{} on raw R2E-Gym & FrogNano \\
\midrule
Base model & Qwen3.5-4B & Qwen3.5-4B \\
Harness / objective & Mini-SWE-Agent-Plus / CISPO & Leaf / DPPO \\
Verified evaluation & 483 tasks, 4 rollouts, 100 max turns, temperature 1.0, top-$p$ 0.95, top-$k$ 20 & 500 tasks, 3 seeds, 150 max turns, temperature 0.6 top-$p$ 0.95, top-$k$ 20 \\
Best checkpoint (\%) & 61.96 (update 340) & 61.5 \\
Mean, updates 300--500 (\%) & 58.44 & Not reported \\
\midrule
Trainer / inference GPUs & 2 / 6 & 2 / 6 \\
Concurrency cap & 252 & Up to 256 \\
Learning rate & $10^{-6}$ & $10^{-6}$ \\
$\beta_2$ & 0.98 & 0.98 \\
Configured policy lag & Up to 8 updates & 1 or 3 updates \\
Rollouts per update & 128 & 256 \\
Nominal rollout slots & 43{,}520 to the best checkpoint & 256{,}000 \\
Reference KL / entropy bonus & None & None \\
Advantage normalization & None & Group standard deviation \\
Length shaping & None & Success-gated log penalty in climbs 3--5 \\
Optimizer updates & 800, persistent optimizer & Five 200-update climbs, optimizer reset \\
Task pool & Initially fixed 3{,}606; easy and hard retirement & Regenerated and calibrated between climbs \\
Checkpoint selection & SWE-bench Verified subset & SWE-bench Verified \\
\bottomrule
\end{tabularx}
\end{table}

Our held-out accuracy reaches 60.1--62.0\% at five checkpoints between updates 220 and
380 (56.5--59.4\% at the checkpoints in between), then eases to the mid-to-high 50s by
update 500; Section~\ref{sec:main-run}
examines the same pattern in the curated-data runs.

\section{Sandbox resource sharing and admission}
\label{app:sandboximplementation}

\paragraph{Requests and limits serve different purposes.}
Each sandbox requests 0.25 vCPU and 0.5\,GiB for scheduling, with limits of four vCPUs
and 16\,GiB. The memory request does not cap RSS or allocate private physical pages.
Each host shares its own headroom among resident sandboxes; unused memory on another
host cannot satisfy a running pod's allocation. Ephemeral disk requests and limits are
2\,GiB and 16\,GiB. Pooling is intended for staggered bursts, not simultaneous operation
of all sandboxes at their limits.

The standalone EC2 configuration has eight \texttt{m8idn.8xlarge} hosts, providing
256 vCPUs and 1{,}024\,GiB before reserves. At 252 concurrent sandboxes, requests total
63 vCPUs and 126\,GiB, while limits total 1{,}008 vCPUs and 4{,}032\,GiB.
The physical pool therefore covers roughly a quarter of simultaneous peak demand.
Our actual reference run uses node-local k3s on the GPU node; repricing it on this
separate pool is a budget comparison, not a record of its rental expenditure.

\paragraph{Pressure controls supplement scheduling.}
When enabled, the pressure monitor polls every fifteen seconds by default. It constrains
admission at memory usage of 0.85 relative to allocatable memory and releases the constraint
below 0.75; CPU thresholds are 0.95 and 0.85. These are implementation defaults, not
verified settings for every run. Restricting new admissions cannot reclaim resources from
running pods. Host reserves and kubelet eviction provide additional responses, but fast
memory growth can still reach the kernel OOM killer first.

Admission capacity also depends on system pods, pod-count limits and actual CNI/IPAM
address availability. Reading the node's advertised pod capacity alone does not establish
the number of usable sandbox slots. Appendix~\ref{app:sandboxobservations} separates
retried failures from terminal rollout errors; neither counter identifies the host-level
cause without telemetry.

\begin{figure}[t]
    \centering
    \includegraphics[width=0.8\linewidth]{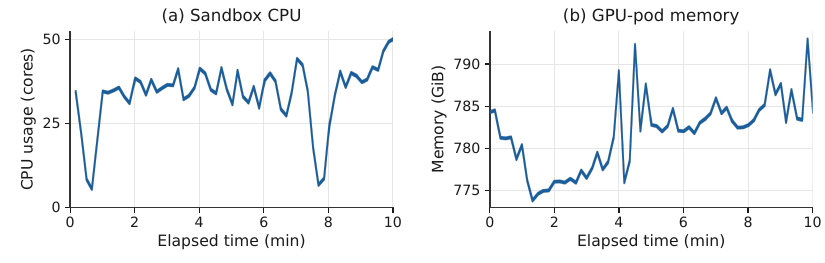}
    \caption{
        Resource usage during a ten-minute local k3s run, sampled every
        ten seconds.
        Left: aggregate CPU usage of the k3s cgroup subtree, expressed in
        core equivalents.
        Right: memory charged to the enclosing GPU-pod cgroup, including
        sandboxes, training and inference processes, and page cache.
    }
    \label{fig:sandbox-resource-usage}
\end{figure}

\paragraph{Observed resource usage.}
Figure~\ref{fig:sandbox-resource-usage} shows a ten-minute utilization trace
from an additional node-local k3s experiment. Our shared-resource sandbox
deployment supported more than 200 concurrent rollouts on a single GPU node, with
aggregate k3s CPU usage peaking at approximately 50 cores, well within the
node's 176 allocatable vCPUs. No new OOM kills were observed during this
window. For the tested workload, these results demonstrate the feasibility
of supporting substantial sandbox concurrency within the available
node-local CPU resources.

\section{Sandbox cost basis}
\label{app:sandboxcost}

\begin{table}[t]
    \centering
    \small
    \caption{Standalone sandbox rental budgets for 252 concurrent capacity slots over 69.93 hours
    (251{,}753 seconds), using the reference run's cap and reported runtime. Our row charges
    the entire EC2 pool and lists per-sandbox burst limits; commercial rows reserve the listed
    resources throughout. These are pricing scenarios, with performance unmeasured across
    providers. Sources and assumptions are in Appendix~\ref{app:sandboxcost}.}
    \label{tab:sandboxcost}
    \begin{tabular}{@{}llrr@{}}
    \toprule
    Setup & Per-sandbox resources & \$/slot-h & Total (\$) \\
    \midrule
    \textbf{Ours, EC2 pool (on-demand)} & 4 vCPU / 16 GiB limits & \textbf{0.0970} & \textbf{1,709} \\
    \midrule
    Northflank & 4 vCPU / 16 GB & 0.2000 & 3,524 \\
    Morph Devboxes & 4 vCPU / 16 GiB & 0.2000 & 3,525 \\
    Prime Intellect & 4 vCPU / 16 GB & 0.3600 & 6,344 \\
    Daytona$^\dagger$ & 4 vCPU / 16 GiB & 0.4608 & 8,121 \\
    E2B$^\dagger$ & 4 vCPU / 16 GiB & 0.4608 & 8,121 \\
    Modal & 2 physical cores / 16 GiB & 0.6680 & 11,772 \\
    Runloop & 4 CPUs / 16 GB & 0.8352 & 14,718 \\
    \bottomrule
    \end{tabular}
    \par\smallskip
    \begin{minipage}{\linewidth}
    \footnotesize
    In our actual 4B setup, sandboxes use spare CPU and host memory on the already-rented
    GPU node, with essentially zero additional infrastructure rental cost. The EC2 row is
    a standalone comparison budget. Its pool is 8$\times$\texttt{m8idn.8xlarge}:
    256 vCPUs and 1{,}024 GiB of physical
    capacity before host reserves. Each pod requests only 0.25 vCPU and 0.5 GiB; these
    requests govern placement, while bursts share spare capacity on its host.
    Sufficient account quotas are assumed. $^\dagger$The 16 GiB allocations require higher
    memory limits than the standard documented configurations; prices extrapolate unit rates.
    Modal equates one physical core to two vCPUs and also offers usage-based bursting.
    Morph's MCU rate bundles disk; up to 64 GiB does not increase this row's rate.
    \end{minipage}
\end{table}

\paragraph{Workload and accounting.} Table~\ref{tab:sandboxcost} uses $N=252$ capacity slots and
$T=251{,}753/3600\approx69.93139$ hours, giving $NT=17{,}622.71$ capacity-slot hours.
The source is a 500-step reference training run: its reported runtime is 251{,}753 seconds,
and its configured global concurrency cap is 252 (42 per client). Its sandbox configuration records
0.25/4 CPU requests/limits, 0.5/16\,GiB memory requests/limits, and node-local k3s.
The reported runtime supplies a rental-budget proxy; it is not a measured sandbox billing interval.
Our actual 4B setup colocates sandboxes on spare CPU and host memory of the already-rented GPU
node, with essentially zero additional infrastructure rental cost. The table charges a standalone
EC2 pool to make the rental comparison independent of this existing capacity. Continuous occupancy
of all slots is not assumed to have been observed. Let $r_{\mathrm{EC2}}$ be the hourly instance price and
$r_{\mathrm{CPU}},r_{\mathrm{mem}}$ a provider's hourly unit rates. Then
\begin{align*}
    C_{\mathrm{pool}} &= 8T r_{\mathrm{EC2}}, &
    p_{\mathrm{pool}} &= \frac{C_{\mathrm{pool}}}{NT}
                       = \frac{8r_{\mathrm{EC2}}}{N}, \\
    C_{\mathrm{provider}} &= NT(c r_{\mathrm{CPU}} + m r_{\mathrm{mem}}),
\end{align*}
where $c$ and $m$ are the CPU and memory quantities shown in the table, in that provider's
units. The additive provider formula applies to all rows except Morph, whose bundled rate
is given below. Totals use the exact duration and unrounded rates and are displayed to the
nearest dollar; hourly table entries are rounded to four decimal places.

\paragraph{EC2 rate.} The \href{https://aws.amazon.com/ec2/pricing/on-demand/}{AWS public
Linux on-demand price} for \texttt{m8idn.8xlarge} in US East (N.\ Virginia) is
\$3.05504 per instance-hour, retrieved September 16, 2026. The AWS price catalog lists
32 vCPUs, 128\,GiB of memory, and one 1{,}900\,GB NVMe instance-store device per host.
Eight hosts cost \$24.44032 per hour, \$1{,}709.15 for the reference duration, and
\$0.0969854 per capacity-slot hour at the assumed concurrency. The entire fleet is charged,
including capacity above the sum of requests. No spot or reserved-instance discount is used.

\paragraph{Commercial unit rates.} The following public rates were retrieved September 16,
2026. All rows apply CPU and memory rates to the listed allocation for the full duration.
\begin{itemize}
    \item \href{https://northflank.com/pricing}{Northflank}: \$0.01667 per vCPU-hour
          and \$0.00833 per GB-hour, giving $4(0.01667)+16(0.00833)=\$0.19996$ per hour.
    \item \href{https://cloud.morph.so/web/subscribe}{Morph Devboxes}: the standard rate is
          \$0.05 per Morph Compute Unit (MCU). Hourly usage is
          \[
            \max(c,\lceil m/4\rceil,\lceil d/16\rceil)
          \]
          for $c$ vCPUs, $m$\,GiB of memory, and $d$\,GiB of disk.
          Four vCPUs and 16\,GiB cost four MCU/hour, or \$0.20/hour,
          provided disk is at most 64\,GiB. Subscription credit discounts are excluded.
    \item \href{https://docs.primeintellect.ai/sandboxes/overview}{Prime Intellect}:
          \$0.05 per core-hour and \$0.01 per GB-hour. The documentation specifies whole
          vCPUs for VM sandboxes; $4(0.05)+16(0.01)=\$0.36$ per hour.
    \item \href{https://www.daytona.io/pricing}{Daytona}: \$0.0504 per vCPU-hour
          and \$0.0162 per GiB-hour, giving $4(0.0504)+16(0.0162)=\$0.4608$ per hour.
          This extrapolates to 16\,GiB: the standard Linux/container resource table in its
          \href{https://www.daytona.io/docs/en/sandboxes}{sandbox documentation} specifies
          a maximum of four vCPUs and 8\,GiB. A higher memory limit must be arranged;
          availability at the extrapolated price is not established here.
    \item \href{https://e2b.dev/pricing}{E2B}: \$0.0504 per vCPU-hour and \$0.0162 per
          GiB-hour, also giving \$0.4608/hour for four vCPUs and 16\,GiB. The published
          Pro plan lists memory above 8\,GiB as requiring contact with support. This row
          is likewise conditional on a higher memory limit at the stated unit rates.
    \item \href{https://modal.com/pricing}{Modal Sandboxes}: \$0.00003942 per physical
          core-second and \$0.00000667 per GiB-second. Using its stated equivalence of
          one physical core to two vCPUs, the row uses two physical cores and 16\,GiB:
          $3600[2(0.00003942)+16(0.00000667)]=\$0.668016$ per hour.
    \item \href{https://www.runloop.ai/pricing}{Runloop}: \$0.108 per CPU-hour
          and \$0.0252 per GB-hour, giving $4(0.108)+16(0.0252)=\$0.8352$ per hour.
\end{itemize}
CPU labels and GB/GiB units follow the cited rate or resource definitions; they do not
establish identical hardware performance or byte-for-byte memory equivalence. Runloop's
pricing page labels memory in GB, while its custom-resource documentation uses GiB.
Modal's
\href{https://modal.com/docs/guide/sandbox-resources}{sandbox resource documentation} also
permits requests below limits and bills the larger of requested resources and actual use.
Its table row prices a reservation of the listed resources. Northflank supports pause and
scale-to-zero, Morph supports sleep, and Daytona and E2B stop charging CPU and memory in
stopped or paused states. These options preclude a general claim that commercial services
must charge for peak resources throughout model-generation waits. Evaluating usage-based or
suspend/resume billing requires resource and activity traces. All vendor rows are calculated
pricing scenarios; workload performance on those services remains unmeasured.

\paragraph{Scope and quotas.} The comparison covers CPU and memory rental plus storage bundled
in the quoted rate. Separately billed storage,
image builds, network transfer, fixed subscription fees, taxes, and operational engineering
are excluded. No promotional credits or negotiated discounts are applied. Sufficient account
quotas are assumed, and increases may be necessary: for example, Prime Intellect's documented
default total CPU quota is 512 vCPUs, below the 1{,}008 needed for these reservations;
Morph's listed Devbox tiers reach 128 concurrent instances; E2B Pro lists 100 by default
with a paid concurrency add-on. Subscription and quota-related charges are outside the table.
Training and inference GPU costs are also outside this sandbox line item.
For the actual colocated deployment, the GPU-node rental remains in the training/inference
budget; essentially zero incremental sandbox rental does not mean zero allocated resource cost.

\paragraph{Sandbox observations.}
\label{app:sandboxobservations}
We examine sandbox errors in three training runs. In
the 500-step run,
the per-tick dispatcher counters record 149 failed attempts from sandbox errors
and 229 from sandbox transport errors. Its terminal-error fraction is positive in
67 of 500 training arrival windows. Summing the per-window rollout and error counts gives
174 terminal-error observations out of 138{,}624 (0.126\%); 166 exhausted their reschedule attempts
and eight were cancelled.
Over its first 467 steps, a second main run has 200 terminal-error observations out of 141{,}160 (0.142\%),
whereas a prompt/tool ablation has 4{,}123 out of 53{,}624 (7.69\%) over its first 188 steps.
These selected runs are not a representative reliability benchmark.

We removed identical duplicate logged rows and
summed the dispatcher's per-tick counters, which reset after logging and include retry attempts.
The separate per-window error-rate metric describes final outcomes in each
training arrival window. We weight it by that window's rollout count and
cross-check against the logged error-type counts; this denominator excludes failed retry
attempts and is not the shipped training cohort. All three runs record the 252-rollout cap and
the same sandbox resource requests and limits, with node-local k3s configurations. Their metrics neither establish
which host-level pressure controls were active nor diagnose OOM or other root causes.
Per-node telemetry and a deployment-matched comparison are needed to attribute errors and
quantify pooling's effect on throughput and task success.

\end{document}